\documentclass[twoside,11pt]{article}

\usepackage{jair}
\usepackage{theapa}
\usepackage{rawfonts}

\usepackage{amsmath,amssymb,amsthm}
\usepackage{enumerate}
\usepackage{booktabs}
\usepackage[dvipsnames]{xcolor}
\usepackage{xspace}
\usepackage{ifthen}

\usepackage{pifont} % for \cmark and \xmark (for table in serializations)
\newcommand{\cmark}{\ding{51}}
\newcommand{\xmark}{\ding{55}}

\usepackage{tikz}
\usetikzlibrary{arrows.meta, calc, positioning, backgrounds}
\tikzset{%
  diagonal fill/.style 2 args={fill=#2, path picture={
    \fill[#1, sharp corners] (path picture bounding box.south west) -|
                             (path picture bounding box.north east) -- cycle;}},
  reversed diagonal fill/.style 2 args={fill=#2, path picture={
    \fill[#1, sharp corners] (path picture bounding box.north west) |-
                             (path picture bounding box.south east) -- cycle;}}
}

\usepackage{algorithmicx}
\newboolean{includeProofs}
\setboolean{includeProofs}{true} % flip to include/exclude proofs
\newcommand{\Proof}[1]{\ifthenelse{\boolean{includeProofs}}{\vspace{-1eM}\begin{proof}\color{black} #1 \end{proof}}{}}
\def\eob{{ % end-of-block symbol: complex stuff, taken from QED.sty, do not modify except for symbol below
  \parfillskip=0pt % so \par doesnt push \square to left
  \widowpenalty=10000 % so we don't break the page before \square
  \displaywidowpenalty=10000 % ditto
  \finalhyphendemerits=0 % TeXbook exercise 14.32
  \leavevmode % \nobreak means lines not pages
  \unskip % remove previous space or glue
  \nobreak % don't break lines
  \hfil % ragged right if we spill over
  \penalty50 % discouragement to do so
  \hskip.2em % ensure some space
  \null % anchor following \hfill
  \hfill % push \square to right
  \raisebox{0pt}{\tikz{\draw[fill=black] (0,0) rectangle +(.5ex,1.5ex);}} % actual symbol
  \par}
} % build paragraph

\newboolean{includeExamples}
\setboolean{includeExamples}{true} % flip to include/exclude examples
\newcommand{\Example}[1]{\ifthenelse{\boolean{includeExamples}}{#1}{\stepcounter{example}}}
\newcounter{example}
\newenvironment{boxed-example}[1][\unskip]{%
  \small\refstepcounter{example}
  \smallskip
  \begin{tcolorbox}[colframe=ForestGreen!60!black,colback=ForestGreen!10!white,breakable,enhanced,title=\textbf{Example \theexample: #1}]}{%
  \end{tcolorbox}}

\usepackage{caption}
\usepackage{tcolorbox}
\tcbuselibrary{breakable, skins}
\tcbset{colframe=black!60!white,colback=black!5!white,left=0mm,top=0.5mm,bottom=0.5mm,right=1.5mm}
\newcounter{myalgorithm}

\newcommand{\Omit}[1]{}
\newcommand{\NoOmit}[1]{#1}
\newcommand{\tup}[1]{\langle #1 \rangle}
\renewcommand{\O}{\mathcal{O}} % asymptotics
\colorlet{alert}{red!80!black}

\newcommand{\hector}[1]{\textcolor{red}{#1}}
\newcommand{\alert}[1]{\textcolor{alert}{#1}}

\newcommand{\citeay}[1]{\citeauthor{#1} \citeyear{#1}}
\newtheoremstyle{colored}% name of the style to be used
  {\topsep}% measure of space to leave above the theorem. E.g.: 3pt
  {\topsep}% measure of space to leave below the theorem. E.g.: 3pt
  {\itshape\color{black}}% name of font to use in the body of the theorem
  {}% measure of space to indent
  {\bfseries}% name of head font
  {}% punctuation between head and body: NONE <-- only diff. from default
  { }% space after theorem head; " " = normal interword space
  {\thmname{#1}\thmnumber{ #2}\thmnote{ \textcolor{black}{\normalfont(#3)}}.}
\newtheoremstyle{grayed}% name of the style to be used
  {\topsep}% measure of space to leave above the theorem. E.g.: 3pt
  {\topsep}% measure of space to leave below the theorem. E.g.: 3pt
  {\itshape\color{mygray}}% name of font to use in the body of the theorem
  {}% measure of space to indent
  {\bfseries}% name of head font
  {}% punctuation between head and body: NONE <-- only diff. from default
  { }% space after theorem head; " " = normal interword space
  {\textcolor{mygray}{\thmname{#1} G\thmnumber{#2}\thmnote{ (#3)}}}
\newtheoremstyle{unnumbered}% name of the style to be used
  {\topsep}% measure of space to leave above the theorem. E.g.: 3pt
  {\topsep}% measure of space to leave below the theorem. E.g.: 3pt
  {\itshape\color{black}}% name of font to use in the body of the theorem
  {}% measure of space to indent
  {\bfseries}% name of head font
  {}% punctuation between head and body: NONE <-- only diff. from default
  { }% space after theorem head; " " = normal interword space
  {\thmname{#1}\thmnote{ \textcolor{black}{\normalfont(#3)}}.}

\theoremstyle{colored}
\newtheorem{definition}{Definition}
\newtheorem{theorem}[definition]{Theorem}
\newtheorem{lemma}[definition]{Lemma}
\theoremstyle{unnumbered}
\newtheorem{assumption}{Assumption}

\newcommand{\sieve}{\textsc{Sieve}\xspace}
\newcommand{\states}{\text{states}}
\newcommand{\Q}{\mathcal{Q}}
\newcommand{\OPT}{OPT}

\newcommand{\pplus}{\hspace{-.05em}\raisebox{.15ex}{\footnotesize$\uparrow$}}
\newcommand{\mminus}{\hspace{-.05em}\raisebox{.15ex}{\footnotesize$\downarrow$}}

\newcommand{\EQ}[1]{#1{\,{=}\,}0}
\newcommand{\GT}[1]{#1{\,{>}\,}0}
\newcommand{\DEC}[1]{#1\mminus}
\newcommand{\INC}[1]{#1\pplus}
\newcommand{\UNK}[1]{#1?}

\newcommand{\prule}[2]{\{ #1 \} \mapsto \{ #2 \}}

\newcommand{\iw}[1]{\ensuremath{\text{IW}(#1)}\xspace}
\newcommand{\iwf}[1]{\ensuremath{\text{IW}_{#1}}\xspace}
\newcommand{\siwp}{\ensuremath{\text{SIW}_{\prec}}\xspace}
\newcommand{\siwR}{\ensuremath{\text{SIW}_{\text R}}\xspace}
\newcommand{\nstate}[1]{\ensuremath{#1[\mathsf{state}]}} % state for node n
\newcommand{\ncost}[1]{\ensuremath{#1[\mathsf{cost}]}}   % cost of node n
\newcommand{\cost}{\textup{cost}}

\newcommand{\AlgCOMMENT}[1]{\textcolor{NavyBlue}{#1}}
\newcommand{\AlgINPUT}{\textbf{Input:}\xspace}
\newcommand{\AlgOUTPUT}{\textbf{Output:}\xspace}
\newcommand{\AlgFAILURE}{\textbf{FAILURE}\xspace}
\newcommand{\AlgACCEPT}{\textbf{ACCEPT}\xspace}
\newcommand{\AlgREJECT}{\textbf{REJECT}\xspace}

\newcommand{\atoms}{\text{atoms}}
\newcommand{\size}{\text{size}}
\newcommand{\ocost}{\textup{cost}^{*}}
\newcommand{\pcost}{\prec_{cost}}
\newcommand{\reachable}{\textup{Reachable}}

\newcommand{\QBlocks}{\ensuremath{\Q_{Blocks}}\xspace}
\newcommand{\QClear}{\ensuremath{\Q_{Clear}}\xspace}
\newcommand{\QOn}{\ensuremath{\Q_{On}}\xspace}
\newcommand{\QGrid}{\ensuremath{\Q_{Grid}}\xspace}
\newcommand{\QGridXY}{\ensuremath{\Q_{Grid_2}\!}\xspace}
\newcommand{\QDelivery}{\ensuremath{\Q_{D}}\xspace}
\newcommand{\QDeliveryS}{\ensuremath{\Q_{D_1}\!}\xspace}
\newcommand{\QMarbles}{\ensuremath{\Q_{M}}\xspace}
\newcommand{\QMarblesS}{\ensuremath{\Q_{M_1}\!}\xspace}
\newcommand{\QHanoi}{\ensuremath{\Q_{Hanoi}\!}\xspace}
\newcommand{\QHanoiOdd}{\ensuremath{\Q_{HanoiOdd}\!}\xspace}

\plainheading{1}{2023}{1-40}{10/23}{?/??}
\ShortHeadings{General Policies, Subgoal Structure, and Planning Width}{Bonet \& Geffner}
\firstpageno{1}

\begin{document}
\allowdisplaybreaks

\title{General Policies, Subgoal Structure, and Planning Width}
\author{\name Blai Bonet \email bonetblai@gmail.com \\
       \addr Universitat Pompeu Fabra, Spain
       \AND
       \name Hector Geffner \email hector.geffner@ml.rwth-aachen.de \\
       \addr RWTH Aachen University, Germany \\
       \addr Link\"{o}ping University, Sweden
}

\maketitle

\begin{abstract}
  It has been observed that many classical planning domains with atomic goals can be solved
  by means of a simple polynomial exploration procedure, called IW, that runs in time exponential in
  the problem width,  which in these cases is bounded and small.  Yet, while the notion of width
  has become part of state-of-the-art planning algorithms such as BFWS, there is   no good explanation
  for why so many  benchmark domains have bounded width when atomic goals are considered. In this work, we address
  this question by  relating bounded width  with the existence of  general optimal  policies
  that in each planning instance are  represented by tuples  of atoms of bounded size. We also define
  the notions  of  (explicit) serializations  and serialized width that have  a broader scope
  as many domains have  a bounded serialized width but no bounded width.
  Such problems   are solved non-optimally in polynomial time by a suitable variant of the Serialized IW algorithm.
  Finally, the language of general policies and the semantics of serializations are combined to yield
  a simple, meaningful, and expressive language for specifying  serializations in compact form in the form of sketches,
  which can be used for encoding domain control knowledge by hand or for learning it from small examples.
  Sketches express general problem decompositions in terms of subgoals,  and  sketches of bounded width express problem decompositions
  that can be solved in polynomial time.\footnote{% \textbf{Note for reviewers:}
    This paper is a  heavily revised version of \citeay{bonet:aaai2021}. While the key notions of problem width, general policies,
    and sketches are from that paper, the relations among these notions have been reworked to make them more meaningful
    and transparent. The paper also contains many new results. A summary of them and a discussion of  their meaning can be found in Sect.~7.
  }
\end{abstract}

\section{Introduction}

Width-based search methods exploit the structure of states to enumerate  the state space in ways
that are different than ``blind'' search methods such as  breadth-first and  depth-first \cite{nir:ecai2012}.
This is achieved by associating a non-negative integer  to each state generated in the search,
a so-called  \emph{novelty measure}, which is defined by  the size of the smallest factor in the state
that has not been seen in previously  generated states.  States are deemed more novel and
hence preferred in the exploration search when this novelty measure is smaller.
Other types of  novelty measures have been used in reinforcement learning for dealing
with sparse rewards and in genetic algorithms for dealing with local minima
\cite{rl-exploration1,rl-exploration2,rl-exploration3}, but the results  are mostly empirical.
In classical planning,  where novelty measures are  part of state-of-the-art search algorithms
\cite{nir:icaps2017,nir:aaai2017},
% and  the resulting algorithms have  been applied
% in settings where  there is no declarative model of the actions or  goals
%  \cite{miquel:ijcai2015,guillem:ijcai2017,tomas:,bandres:aaai2018},
there is  a solid  body of theory  that relates  a specific type of
novelty measures  with  a  notion of problem \emph{width}
that  bounds the complexity of planning problems \cite{nir:ecai2012}. 

The basic width-based planning algorithms are simple and  assume a fixed number of \emph{Boolean state features} $F$ that
in classical planning are  given by the atoms in the problem. The procedure \iw{1} is a breadth-first search that starts in the given
initial state and prunes all the states that do not make a feature from $F$ true for the first time in the search.
\iw{k} is like \iw{1} but using conjunctions of up to $k$ features from $F$ instead.
Alternatively, \iw{k} can be regarded as a breadth-first search that prunes states $s$ with novelty measures
that are greater than $k$, where the novelty measure of $s$ is the size of the minimum conjunction (set) of  features that
is   true in $s$ but false in all the states generated before $s$. 

For many benchmark domains, it has been shown that \iw{k} for a small  value of $k$
suffices to compute plans, and indeed optimal plans, for any atomic goal \cite{nir:ecai2012}.
State-of-the-art planning  algorithms like BFWS \cite{nir:icaps2017,nir:aaai2017}  make use  of this property
for \emph{serializing}  conjunctive goals into atomic ones, an idea that is also  present
in algorithms that pre-compute atomic landmarks \cite{hoffmann:landmarks}
and  use them as  counting heuristics \cite{lama}.

An important  open question in the area is why these width-based methods are  effective, and in particular, why
so many domains have a small width when atomic goals are considered. Is this a property of the domains?
Is it an accident of the domain encodings  used? In this work, we address these and related questions.
For this, we bring the notion of \emph{general policies}; policies that solve multiple instances of a planning
domain all at once \cite{srivastava08learning,bonet:ijcai2015,hu:generalized,BelleL16,anders:generalized},
while using the formulation of general policies  expressed in terms of  finite sets of rules
over a  fixed set of Boolean and numerical features  \cite{bonet:ijcai2018}.

% that we restrict to be {linear}.
% i.e., the value of a feature $\phi(s)$ in a state must be computed in time that is
% linear in the number of atoms $N$ in the problem and the number of such values must be linear in $N$
% as well. Some of the formal results to be developed

In this paper,   a class of instances is shown to have \emph{bounded width} when  there is a \emph{general optimal policy}
for the class which can be ``followed'' in each instance by  just considering atom tuples of bounded size,
without assuming knowledge of the policies or the features involved. The existing planning domains
that have been shown to have bounded width can be all characterized in this way.
In addition, the notion of general policies is extended to comprise serializations that split problems into subproblems.
A serialization has bounded width when the resulting subproblems have bounded width and can be 
solved greedily for reaching the problem goal. A general policy turns out to be a  serialization of width zero;
namely, one in which the subproblems can be solved in a single step. 
Finally, the syntax of general policies is combined with the semantics of serializations
to yield a simple, meaningful, and expressive language for specifying  serializations succinctly in the form of
\emph{sketches},  which can be used to encode domain control knowledge by hand, or for learning it from examples
\cite{drexler:icaps2021,drexler:icaps2022}.

% While exploring these potential uses of sketches is beyond the scope
% of this paper, the use of sketches and the meaning of the theoretical
% results are all illustrated through a number of examples.

The paper is organized as follows. We review first the notions of planning, width, and general policies,
and relate width with the size of the tuples of atoms that are needed to apply such policies. 
We then introduce serializations, the more general notion of serialized width, the relation between general policies
and serialized width, and policy sketches. We finally summarize the main contributions, discuss extensions and limitations,
related work, and conclusions.

\section{Planning}

A classical planning problem is a pair $P=\tup{D,I}$ where $D$ is a first-order
\emph{domain}, such as a STRIPS domain, and $I$ contains information about the instance \cite{geffner:book,ghallab:book,pddl:book}.
The domain $D$ has a set of predicate symbols $p$ and a set of action schemas with
preconditions and effects given by atoms $p(x_1, \ldots, x_k)$, where $p$ is a predicate
symbol of arity $k$, and each $x_i$ is an argument of the schema.
The instance information is a tuple $I=\tup{O, \textit{Init},G}$ where $O$ is a
set of object names $c_i$, and $\textit{Init}$ and $G$ are sets of
\emph{ground atoms} $p(c_1, \ldots, c_k)$ denoting the initial and
goal situations.

A classical problem $P=\tup{D,I}$ encodes a state model $S(P)=\tup{S,s_0,S_G,\textit{Act},A,f}$
in compact form where the states $s \in S$ are \emph{sets of ground atoms} from $P$ (assumed to be the true
atoms in $s$), $s_0$ is the initial state $I$, $S_G$ is the set of goal states $s$ such that $G \subseteq s$,
$\textit{Act}$ is the set of ground actions in $P$, $A(s)$ is the set of ground actions
whose preconditions are (true) in $s$, and $f(a,s)$, for $a \in A(s)$, represents
the state $s'$ that follows action $a$ in the state $s$. 
An action sequence $\sigma=a_0,a_1,\ldots,a_{n}$ is applicable in $P$ if
$a_i \in A(s_i)$ and $s_{i+1}=f(a_i,s_i)$, for $i=0,\ldots,n$.
The states $s_i$ in the sequence are said to be \emph{reachable} in $P$,
and $\states(P)$ denotes the set of reachable states in $P$.
The applicable action sequence $\sigma=a_0,a_1,\ldots,a_n$ is a \emph{plan} if $s_{n+1} \in S_G$.
The \emph{cost} of a plan is given by its length, and a plan is \emph{optimal}
if there is no shorter plan. If there is a plan for $P$, the goal $G$ of $P$
is said to be reachable, and $P$ is said to be \emph{solvable}.
The \emph{cost} of $P$ is the cost of an optimal plan for $P$, if such a plan
exists, and infinite otherwise.
A reachable state $s$ is a \emph{dead-end}  if the goal is not  reachable
in the problem $P[s]$ that is  like $P$ but with $s$ as the initial state.

We refer to subsets of ground atoms in a planning problem $P$ as
tuples of atoms, or atom tuples. An atom tuple $t$ is reachable in $P$ if 
there is a reachable state $s$ in $P$ that makes true all the atoms in $t$.
%Likewise, a \emph{tuple of (ground) atoms} $t$ is reachable in $P$
%if  a state $s$  where $t$ is true is reachable in $P$.
% if the goal is reachable in the problem $P[G=t]$ that is like $P$ but with goal $G=t$. 
The cost of a reachable state $s$ is the minimum cost of a plan that reaches $s$,
the cost of a reachable tuple $t$ is the minimum cost of a state that makes $t$ true,
and the cost of a state or tuple that is unreachable is infinite.

\Omit{
  \hector{Define (not displayed) reachable states and dead-ends. E.g.,} 
  A {state} $s$ over problem $P=(D,O,I,G)$ is a collection of ground atoms.  The state $s$ is a {goal} if $G \subseteq s$, and a {dead end}
  if a goal state cannot be reached from $s$ in $P$.   A state $s$ denotes a unique valuation for the ground atoms in $P$: $s\vDash r(c_1,\ldots,c_k)$
  iff $r(c_1,\ldots,c_k)$ belongs to $s$.

  \hector{See this ...}
  
  \begin{definition}[Domains, problems, and states]
    \label{def:domains-problems-states}
    A \textbf{domain} is a pair $D=(R,F)$ where
    $R$ is a set of \textbf{primitive predicate symbols} with their corresponding arities, and
    $F$ is a set of \textbf{features} defined in terms of the primitive predicates with their corresponding
    range of feature values.
    A \textbf{problem} $P$ over domain $D=(R,F)$ is a tuple $P=(D,O,I,G)$ where $O$ is a set of unique object
    names $c$ (objects), and $I$ and $G$ are sets of ground atoms that denote the \textbf{initial} and \textbf{goal} states of $P$.
    A ground atom $r(c_1, \ldots, c_{a(r)})$ is made of a predicate $r \in R$ and an object tuple in $O^{a(r)}$ for the
    arity $a(r)$ of $r$.
    A \textbf{state} $s$ over problem $P=(D,O,I,G)$ is a collection of ground atoms.
    The state $s$ is a \textbf{goal} if $G \subseteq s$, and a \textbf{dead end} if a goal state cannot be reached from $s$ in $P$.
    A state $s$ denotes a unique valuation for the ground atoms in $P$: $s\vDash r(c_1,\ldots,c_k)$
    iff $r(c_1,\ldots,c_k)$ belongs to $s$.
  \end{definition}
}

\section{Width}

While the standard definition of width is based on the consideration
of sequences of atom tuples \cite{nir:ecai2012}, the formulation
below is slightly more general and more convenient for our purposes.

% In the following, a ``tuple'' of atoms for a problem $P$ refers to a subset
% of atoms for $P$; i.e., the order or multiplicity of atoms in a tuple does
% not matter.

%\hector{General, revised Definition of width that doesn't require partitioning
%  set of tuples $T$ into sequence of sets $T_0$, \ldots, $T_n$. It should be
%  equivalent.
%}

\begin{definition}[Admissible tuple set]%[Admissible $T$]
  \label{def:admissible}
  A set $T$ of \textbf{reachable} atom tuples in a planning
  problem $P$ is \textbf{admissible} for $P$ if
  \begin{enumerate}[1.]
  \item some tuple in $T$ is true in the initial state of $P$, and
    %\item $\OPT(t)$ is non-empty for each tuple $t$ in $T_i$, $0\leq i\leq n$,
  \item any optimal plan for a tuple $t$ in $T$, that is \textbf{not} an optimal plan for $P$,
     can be extended into an optimal plan for another tuple $t'$ in $T$ by adding a single action.
  \end{enumerate}
\end{definition}

\citeay{nir:ecai2012} say that a \textbf{sequence} $(t_0,t_1,\ldots,t_n)$ of
atom tuples is admissible if $t_0$ is true at the initial state, any optimal
plan for $t_i$ can be extended with one action into an optimal plan
for $t_{i+1}$, $0\leq i<n$, and any optimal plan for %the tuple
$t_n$ is an optimal plan for $P$. It is easy to show that in such a case, the set $T=\{t_0,t_1,\ldots,t_n\}$
is admissible according to the new definition of admissibility. 
There cases, however,  where all  optimal plans  for a fixed tuple $t$ in $T$
can be extended with a single action into   plans that are optimal for \textbf{either}
$t'$ or $t''$ but not for a unique tuple of the same size. 
% or more, in $T$ (i.e., there is no a single $t'$ such
% that \emph{all} optimal plans for $t$ can be extended into an optimal plan
% for $t'$).
In such cases, the width of a problem can be smaller according to the new definition. 

For a tuple $t$ and a set of tuples $T$, $|t|$ denotes the number of
atoms in $t$, $|T|$ denotes the number of tuples in $T$, and $\size(T)$
denotes the maximum $|t|$ for a tuple $t$ in $T$.
The \textbf{width} of a problem $P$ is the size of a minimum-size set
of tuples $T$ that is admissible:

\begin{definition}[Width]
  \label{def:width}
  The \textbf{width} of a STRIPS planning problem
  \Omit{ %%% No me parece complicar las cosas aquí, preventivamente;
    %%% Mejor hacerlo donde se necesite HG
     %%% Tambien puse ``not solvable P'' en lugar de no admissible set for P
    \footnote{The condition for
    $P$ to be a STRIPS problem is important. In particular, it is
    important for the goal to be expressed as conjunction of atoms
    (i.e., positive literals).
    On the Marbles domain, Example~\ref{ex:pi:marbles} below, the problems
    $P$ are not STRIPS problems as the goals can only be expressed with conjunctions
    of negative literals, and no such problem $P$ admits an admissible
    set of tuples.}
  }
  $P$ is $w(P) \doteq \min_T \,\size(T)$
  where $T$ ranges over the sets of tuples that are \textbf{admissible} in $P$.
  If $P$ is solvable in zero or one step, $w(P)$ is set to 0, and
  if $P$ is not solvable at all, $w(P)$ is set to  $\infty$.
\end{definition}

\Omit{
  \begin{definition}[Width of $P$]
    \label{def:width}
    The \textbf{width} $w(P)$ of a solvable problem $P$ is \textbf{bounded} by non-negative integer $k$,
    $w(P)\leq k$, if there is a \textbf{sequence $T=(T_0,T_1,\ldots,T_n)$  of non-empty sets of atom
    tuples of size at most $k$ such that:}
    \begin{enumerate}[1.]
      \item each tuple $t$ in $\cup_{i=0}^n T_i$ is reachable,
      \item each tuple in $T_0$ is true in the initial state of $P$,
      %\item $\OPT(t)$ is non-empty for each tuple $t$ in $T_i$, $0\leq i\leq n$,
      \item any optimal plan for a tuple $t$ in $T_i$ can be extended into an optimal plan
        for some tuple $t'$ in $T_{i+1}$ by adding a single action, $0\leq i<n$, and
      \item any optimal plan for a tuple in $T_n$ is optimal for $P$.
    \end{enumerate}
    The \textbf{width} of $P$, $w(P)$ is $0$ if $P$ is solvable in at most one
    step, it is $N$ if $P$ is unsolvable and $N$ is the number of atoms in $P$,
    and otherwise it is $\argmin{k} w(P) \leq k$.
    Sequences $T$ that comply with conditions 1--4 are called \textbf{admissible sequences} for $P$.
  \end{definition}

  The definition of width above is a slight generalization of the original
  definition by \citeauthor{nir:ecai2012} (\citeyear{nir:ecai2012}) where
  the sets of tuples $T$ cannot contain two tuples $t_i$ and $t'_i$ with
  the same cost $i$ (namely both reachable in $i$ steps).
  This revised definition increases the scope of problems that can
  be provably (optimally) solved by the IW algorithm (see below).
}

\noindent
The definition of width for classes of problems $\Q$ is then:

\begin{definition}[Width of $\Q$]
  \label{def:width:Q}
  The \textbf{width} $w(\Q)$ of a collection $\Q$ of STRIPS problems is
  the minimum integer $k$ such that $w(P)\leq k$ for each problem $P$ in
  $\Q$, or $\infty$ if no such $k$ exists.
\end{definition}

The reason for defining the width of $P$ as $0$ or $\infty$ according to
whether $P$ is solvable in one step or  not solvable at all, respectively,
is convenience. Basically, problems are solvable in time exponential in their width,
and hence, $w(P)=0$ implies that $P$ can be solved in constant time, if a
constant branching factor is assumed. At the same time, a bounded width for a class of problems implies that
all the problems in the class are solvable, something which is not
ensured by setting $w(P)$ to $N+1$, where $N$ is the number of problem
atoms \cite{nir:ecai2012}. 

\Omit{ %%% Addressed above by $P$ being solved ..
  \alert{What happens when there is no admissible sequence; i.e., when $P$ is not solvable?
  Saying that the width is $N+1$ in such a case may generate confusion and problems,
  specially with serializations.
  For example, suppose that $\Q$ is a class of solvable Spanner problems. These can
  be solved in polynomial time with the right serialization. Now, adjoin to $\Q$ a
  small instance with $N=2$ atoms that is unsolvable. The extended collection then
  will have bounded width, yet the serialization won't solve all problems in $\Q$.}
}

% The first condition in the definition forbids the presence of spurious tuples
% in any of the sets of atom tuples.

\Example{
  \bigskip
  \begin{boxed-example}[The Blocksworld domain]
    \label{ex:blocks}
    \begin{enumerate}[$\bullet$]
    \item \QBlocks is the class of all Blocksworld problems over the standard
      domain specification with 4 operator schemas: stack/unstack and pick/putdown
        operators. This class of problems has \textbf{unbounded width} as shown below.
      %%%
      \item \QClear is the subclass of \QBlocks made of the problems whose goal is
        the single atom $clear(x)$, for some block $x$, and where the gripper is initially empty.
%         For determining, the width of  \QClear, let us assume
%         for simplicity, we assume that the gripper is empty in the initial state.

        Let $B_1,\ldots,B_\ell$ be the blocks above $x$, from top top bottom in the initial
        state, and let us consider the set of tuples $T=\{t_0,t_1,\ldots,t_{2\ell-1}\}$ where
        \begin{enumerate}[--]
          \item $t_0=\{clear(B_1)\}$, and
          \item $t_{2i-1}=\{hold(B_i)\}$ and $t_{2i}=\{ontable(B_i)\}$ for $1\leq i\leq\ell$.
        \end{enumerate}
        It is easy to check that the two conditions in Definition~\ref{def:admissible} hold for $T$.
        Hence, $w(P)\leq 1$,  and since $w(P)>0$,  as the goal cannot be reached in zero or one step in general,
        $w(P)=1$ and $w(\QClear)=1$.
\Omit{Simplified this -- Hector
        If $hold(x)$ is true in the initial state of $P$, $w(P)=0$ as with a single action
        the atom $clear(x)$ is reached. On the other hand, if $hold(y)$ is true for some $y\neq x$,
        the set $T$ must be enlarged with the singleton $\{ontable(y)\}$.
        Therefore, $w(\QClear)=1$ as we have shown that $w(P)\leq 1$ for any $P$ in \QClear,
        and there are problems $P$ in $\QClear$ such that $w(P)=1$.
        }
      %%%
      \item \QOn is the subclass of \QBlocks made of the problems whose goal is the single
        atom $on(x,y)$ for two blocks $x$ and $y$.

        Let us calculate the width of an arbitrary instance $P$ with the assumption that
        the blocks $x$ and $y$ are initially at \emph{different} towers.
        Let $B_1,\ldots,B_\ell$ (resp.\ $D_1,\ldots,D_m$) be the blocks above $x$ (resp.\ $y$),
        in order from top to bottom, in the initial state.
        Let us consider the set $T\,{=}\,\{t_0,\ldots,t_{2\ell},t'_0,\ldots,t'_{2m},t''_0,t''_1\}$
        of tuples where
        \begin{enumerate}[--]
          \item $t_i$, $0\leq i<2\ell$, is as in the previous example,
          \item $t_{2\ell}\,{=}\,\{ontable(B_\ell)\}$,
          \item $t'_{2i}\,{=}\,\{hold(D_i),clear(x)\}$ for $0\leq i\leq m$,
          \item $t'_{2i-1}\,{=}\,\{ontable(D_i),$ $clear(x)\}$ for $0\leq i\leq m$,
          \item $t''_0\,{=}\,\{hold(x),clear(y)\}$, and
          \item $t''_1=\{on(x,y)\}$.
        \end{enumerate}
        It is not difficult to check that $T$ is {admissible}.
        Later, we will show that $w(P)>1$ and thus $w(P)=2$.
        If the two blocks $x$ and $y$ are in the same tower in the initial state,
        a different set of tuples must be considered but the width is still bounded
        by 2. Hence, $w(\QOn)=2$.
    \end{enumerate}
  \end{boxed-example}
}

\subsection{Algorithms \iw{T}, \iw{k}, and IW}

If $T$ is an admissible set of tuples for $P$, there is a very simple algorithm
\iw{T} that solves $P$ optimally by expanding no more than $|T|$ states.
The algorithm,  shown in Fig.~\ref{fig:iw(T)},  carries out a forward,
breadth-first search where every newly generated node that does not make
a tuple in $T$ true for the first time in the search is pruned.

\begin{figure}
  \begin{tcolorbox}[title=\textbf{Algorithm~\ref{alg:iw(T)}: \iw{T} Search}]
    \refstepcounter{myalgorithm}
    \label{alg:iw(T)}
    \begin{algorithmic}[1]\small
      \State \AlgINPUT Planning problem $P$ with $N$ ground atoms
      \State \AlgINPUT Set $T$ of atom tuples from $P$
      \smallskip
      \State  Initialize \textbf{perfect hash table} $H$ for storing the tuples in $T$ on which the operations of
      \Statex insertion and look up take constant time
      \State  Initialize FIFO queue $Q$ on which the enqueue and dequeue operations take constant time
      \smallskip
      \State Enqueue node for the initial state $s_0$ of $P$
      \State While $Q$ is not empty:
      \State\qquad Dequeue node $n$ for state $s$
      \State\qquad If $s$ is a goal state, return the path to node $n$                \hfill\AlgCOMMENT{(Solution found)}
      \State\qquad If $s$ makes true some tuple $t$ from $T$ that is not in $H$:
      %\State\qquad\qquad If $s$ is a goal state, return the path to node $n$                \hfill\AlgCOMMENT{(Solution found)}
      \State\qquad\qquad Insert all tuples from $T$ made true by $s$ in $H$
      \State\qquad\qquad Enqueue a node $n'$ for each successor $s'$ of $s$
      \smallskip
      \State Return \AlgFAILURE \hfill\AlgCOMMENT{($T$ is not admissible for $P$)}
    \end{algorithmic}
  \end{tcolorbox}
  \caption{\iw{T} is a breadth-first  search that prunes nodes that do not satisfy a tuple in $T$ for the first time
    in the search. Algorithm \iw{k} is \iw{T} where $T$ is the set of conjunctions
    of up to $k$ atoms in $T$.
    Conditions for the completeness and optimality of \iw{T} are given in Theorem~\ref{thm:iw(T)}.
    %\hector{***Shouldn't lines 8 and 9 be swapped (IW(0) property).****}
  }
  \label{fig:iw(T)}
\end{figure}

\begin{theorem}[Completeness of \iw{T}]
  \label{thm:iw(T)}
  If $T$ is an \textbf{admissible} set of atom tuples in problem $P$,
  \iw{T} finds an \textbf{optimal plan} for $P$.
  Moreover, \iw{T'} finds an optimal plan for $P$ for any set $T'$
  that contains $T$.
\end{theorem}
\Proof{%
  The first claim is a special case of the second.
  Assuming that $T'$ contains an admissible set $T$, it is easy
  to show by  induction that at the beginning of each iteration
  of the loop, the queue contains at least one node
  that represents an optimal path for some tuple in $T$ not yet visited.
  Hence, \iw{T} cannot return \AlgFAILURE and must find a plan for $P$.
  Since the nodes are ordered in the queue by their costs,
  such a plan must be optimal. % \hector{*** Not using admissibility; proof not right, must be refined a bit ***}.
}

% "In some cases, these bounds .. "
% -> La parte del parrafo que se inicia ahi, es ambigua y no muy gramatical
% Which cases? Clever choice of .., preprocessing, or simply ..
% los 3 casos no tienen la misma estructura gramatical
% -> quitaría todo el parrafo desde In some cases, ..
% Y pondría en su lugar que estas suposiciones se cumplen
% in the STRIPS setting.
To obtain simple expressions that bound the time and space used by \iw{T},
and other algorithms, we make the following assumptions on the time/space
required by the basic operations on states and tuples.
For some domains, these bounds can be attained by suitable modifications
of the algorithms, like clever choice of data structures, and preprocessing
that can be amortized.
%or simply special characteristics of the problems (e.g., the number of atoms
%made true by a state is bounded by a constant).
% Hector: no need this extra clarification
% \emph{In any case, the time/space complexities of algorithms in this paper may
% need to be adjusted for deviations from these assumptions.}

\Omit{
  We use the following assumption in order to simplify the time and space analyses
  of the algorithms in this paper. Actual time and space requirements can
  be obtained by adjusting the given complexities accordingly, depending on the
  intrinsic details of the chosen data structures used to store and manipulate
  atoms, tuples, actions, and states.
}

\begin{assumption}[Simple time and space analyses]
  \label{assumption}
  For any planning problem $P$ and reachable state $s$ in $P$,
  the generation of the set $Succ(s)$ of its successor states takes \textbf{time proportional to $|Succ(s)|$,}
  and checking whether $s$ is a goal state or makes true a given atom tuple $t$ takes \textbf{constant time.}
  %\hector{** Reasonable? ***}
  Likewise, each such state can be stored in a \textbf{constant amount of memory.}
  These complexities are \textbf{independent} of the numbers of ground
  atoms and actions in $P$, and the size of the tuple $t$.
\end{assumption}

\Omit{
  The third property in Definition~\ref{def:width} is crucial for obtaining
  a polynomial time algorithm, \iw{k}, that is able to find a solution
  (optimality aside) for a problem with bounded width, as it allows the search
  to safely prune states (cf.\ line 8 in \iw{k}) that do not satisfy a tuple
  for the first time, no matter the order in which the states are generated.
  It is an open question whether there is an alternative and effective
  parametrization of planning problems that guarantee polynomial time solvability,
  not necessarily optimal.
}

\noindent
Under this  assumption, the  time and space requirements of \iw{T} can be expressed as:

\begin{theorem}[Complexity of \iw{T}]
  \label{thm:iw(T):bounds}
  Let $P$ be a planning problem with branching factor bounded by $b$,
  and let $T$ be a set of atom tuples.
  Then,
  \begin{enumerate}[1.]
    \item \iw{T} expands and generates at most $|T|$ and $b|T|$ nodes, respectively,
      thus running in $\O(bT^2)$ time and $\O(bT)$ space (where the $T$ inside the
      $\O$-notation refers to $|T|$).
    %\item If $T$ is \textbf{admissible,} \iw{T} finds an \textbf{optimal plan} for $P$.
    \item If $P$ has $N$ ground atoms, the number of atoms that ``flip'' value across a
      transition in $P$ is \textbf{known and bounded by a constant} (as in STRIPS domains),
      and $\size(T)\leq k$,    then a running time of $\O(bTN^{k-1})$ can be obtained.
%       which improves $\O(bT^2)$ when $|T|=\Omega(N^k)$.
  \end{enumerate}
\end{theorem}
\Proof{%
  Recall that a node is generated if it is inserted into the queue (cf.\ lines 5 and 11),
  and it is expanded if its successor nodes are generated (cf.\ line 11).
  A node is expanded if it makes true some tuple in $T$ for the first time in the
  search (cf.\ line 9).
  Thus, \iw{T} expands and generates up to $|T|$ and $b|T|$ nodes, respectively.

  The construction of the hash table in line 3 requires time and space
  linear in the number $|T|$ of keys to be stored.
  %Thus, the total space required is $\O(bT)$.

  The test in line 8 requires checking whether some tuple $t$ in $T$ that belongs
  to the state $s$ is not in the hash table. This can be done in $\O(T)$ time by
  iterating over each tuple in $T$ and checking whether it belongs to $s$
  and the hash table.
  Hence, assuming constant time and space for the generation of successor states and
  the check of tuple satisfiability by states, \iw{T} runs in $\O(bT^2)$ time and
  $\O(bT)$ space.

  \medskip\noindent
  For obtaining the bound described in 2.  % Second claim. %\textbf{(2)}
  an  implementation that  keeps track of the set of atoms $\Delta$  that change
  value when a successor state $s'$ of $s$ is generated is needed. Then, a tuple $t$ true in 
  $s'$ is  novel iff it contains some atom in $\Delta$ as the tuples  true in  $s'$
  that do not contain such atoms are also  true in  $s$.
  If all the tuples in $T$ are of size at most $k$, and the size of $\Delta$ is bound
  by a constant (omitted in the  $\O$-notation), the number of tuples that need to be checked
  in line 8 is at most $\O(N^{k-1})$.
  % Thus, the running time changes from $\O(bT^2)$  to $\O(bTN^{k-1})$.
}

The algorithm \iw{k} \cite{nir:ecai2012} is a special case of \iw{T}
where $T$ is the set $T^k$  of all conjunctions of up to $k$
atoms.\footnote{There is a minor difference between the algorithm \iw{k}
  of \citeay{nir:ecai2012} and the version that results from \iw{T} when
  $T$ is set to $T^k$.
  In the first case, non-novel nodes are pruned once they are generated
  and thus not enqueued during the search. In the second case, a node is
  pruned when it is selected for expansion if it is not novel and it is
  not a goal.
  This difference does not affect the formal properties of the algorithm
  (optimality and time/memory complexity) except in a ``border case'',
  ensuring that \iw{0} is complete and optimal for problems of width zero,
  solvable in one step, according to the new definition.
}
% For $k=0$, $T^k$ is the singleton that only contains the empty tuple,
% and thus \iw{0} is simply a search of depth $1$.
%The algorithm \iw{0} is \iw{T^0} when except for $\iw{0}$ that is set to be a search of depth $1$.
Versions of \iw{T} have been used before for planning in a class of
video-games \cite{tomas:gvg}, where \iw{1} explored too few nodes
and \iw{2} too many. The set $T$ was then defined to comprise a
selected class of atoms pairs. The properties of \iw{k} are:

\begin{theorem}[\citeauthor{nir:ecai2012}, 2012]
  \label{thm:width:nir}
  Let $P$ be a planning problem with $N$ atoms, branching factor bounded
  by $b$, and where the number of atoms that ``flip'' value across a
  transition is known and bounded. For each non-negative integer $k$, \iw{k} expands up to $N^k$ nodes,
  generates up to $bN^k$ nodes, and runs in $\O(bN^{2k-1})$ time and $\O(bN^k)$ space.
  \iw{k} is guaranteed to solve $P$ optimally (shortest path) if $w(P) \leq k$.
\end{theorem}
%The time bound $\O(bN^{2k-1})$ for \iw{k} relies on the assumption that
%each transition in $P$ affects at most a constant number $M$ of atoms.
%Without this assumption, as when \iw{k} is run on simulators \cite{refs-simulators},
%the time bound becomes $\O(bN^{2k})$.
%% This is specially relevant on problems of width $k=1$ as then \iw{1} runs
%% in linear time rather than quadratic time.
\Proof{%
  Direct from Theorems~\ref{thm:iw(T)} and \ref{thm:iw(T):bounds}.
}

Finally, the algorithm IW, shown in Fig.~\ref{fig:iw}, runs \iw{k} for increasing
values $k=0,1, \ldots,N$ stopping when a plan is found, or when no plan is found
after $k=N$, where $N$ is the number of atoms in $P$.

\Omit{
  Since a state in $P$ makes true at most the $N$ atoms in $P$,
  if \iw{k} does not find a plan for $k=N$, then $P$ has no solution.
  For convenience, aligned with the definition of $w(P)=0$,
  \iw{0} is set to the algorithm that checks whether the initial
  state $s_0$ or some of its successors is a goal state.
}

\begin{figure}
  \begin{tcolorbox}[title=\textbf{Algorithm~\ref{alg:iw}: IW Search}]
    \refstepcounter{myalgorithm}
    \label{alg:iw}
    \begin{algorithmic}[1]\small
      \State \AlgINPUT Planning problem $P$ with $N$ ground atoms
      \smallskip
      \State For $k=0,1,\ldots,N$ do:
      \State\qquad Run \iw{k} on $P$
      \State\qquad If \iw{k} finds a plan for $P$, return the plan
      \smallskip
      \State Return ``no plan exists for $P$'' \hfill\AlgCOMMENT{($P$ has no solution)}
    \end{algorithmic}
  \end{tcolorbox}
  \caption{IW performs multiple \iw{k} searches for increasing values
    of $k=0,1,\ldots,N$, where $N$ is the number of ground atoms in $N$.
    The completeness of IW is given in Theorem~\ref{thm:iw}.
  }
  \label{fig:iw}
\end{figure}

\begin{theorem}[Completeness of IW]
  \label{thm:iw}
  %IW solves any planning problem $P$; i.e., it either finds a plan for $P$
  %or reports that no such plan exists.
  %Furthermore, if $P$ satisfies the conditions in Theorem~\ref{thm:width:nir}
  If $w(P)\leq k$, IW finds a plan for $P$, \textbf{not necessarily optimal,}
  in time and space bounded by $\O(bN^{2k-1})$ and $\O(bN^k)$ respectively.
  %% This "converse" is wrong. IW doesn't find a plan for any Marbles problem
  %%Conversely, if IW does not find a plan for $P$, the problem $P$ has no solution.
\end{theorem}
\Proof{%
  %IW calls \iw{i} for $i=0,2,\ldots$ until a plan for $P$ is found or \iw{i}
  %fails for $i=N$. Since \iw{N} explores the whole state space, if no plan
  %is found, no plan for $P$ exist.
  If $w(P)\leq k$, \iw{k} finds an optimal plan for $P$, but a sub-optimal
  plan may be found by \iw{i} for $i<k$.
  In either  case, by Theorem~\ref{thm:width:nir}, \iw{i} runs in  time and space
  $\O(bN^{2i-1})$ and $\O(bN^i)$, respectively, for $i=1, \ldots, k$, 
  which are  dominated  by the expressions with  $i=k$.
  %%Notice that \iw{N} only prunes nodes whose associated states have already
  %%been expanded. Thus, \iw{N} is a standard breadth-first search over the
  %%reachable states in $P$. Hence, if \iw{N} finds no plan, $P$ has no solution.
}

It is not always necessary to execute all the iterations in the loop in IW.
Indeed, if the call to \iw{k}, for $k=i$, ends up pruning only \emph{duplicate states},
the search is already complete and further calls of \iw{k} for $k=i+1,i+2,\ldots,N$,
will expand and prune exactly the same  sets of nodes expanded and pruned by \iw{i}.

The procedure IW solves problems $P$ of  width bounded by $k$  in polynomial
time but not necessarily optimally. \iw{k}  solves such problems optimally:

\begin{theorem}
  \label{thm:iw:Q}
  Let $\Q$ be a collection of problems of \textbf{bounded width}.
  Then, any problem $P$ in $\Q$ is solved in polynomial time by
  the IW algorithm.
  If $w(\Q)\leq k$, \iw{k} optimally solves any instance in $\Q$
  in polynomial time.
\end{theorem}
\Proof{%
  Let $w(\Q)=k$.
  For any planning problem $P$ in $\Q$, IW runs \iw{i} until $i=k$
  when \iw{k} is guaranteed to find a plan for $P$. Each run of \iw{i}
  takes polynomial time as $k$ is fixed for any problem $P$ in $\Q$.
  If the bound $k$ is \textbf{known}, \iw{k} can be run instead,
  solving each problem $P$ in $\Q$  optimally.
  \Omit{
    If integer $k$ bounds the width of $\Q$, $w(P)\leq k$ for any problem
    $P$ in $\Q$. Therefore, run $\iw{i}$ for $i=1,2,\ldots,k$, and keep the
    minimum-length plan found across these runs, or record that no plan exist
    if all runs fail to find a plan.
    The bound on the width implies that an optimal plan is found, if $P$
    is solvable, or $P$ is shown to be unsolvable otherwise.
    Additionally, since $k$ is fixed, all these runs take polynomial time
    for any problem $P$ in the class $\Q$.
  }
}

\Example{
  \begin{boxed-example}[\iw{k} for Blocksworld instances]
    \label{ex:iwk}
    % Pick(x): ontable(x), clear(x), empty -> -ontable(x), -clear(x), -empty, hold(x)
    % Putdown(x): hold(x) -> ontable(x), clear(x), empty, -hold(x)
    % Stack(x,y): hold(x), clear(y) -> on(x,y), clear(x), empty, -clear(y), -hold(x)
    % Unstack(x,y): on(x,y), clear(x), empty -> hold(x), clear(y), -clear(x), -empty, -on(x,y)
    \begin{enumerate}[$\bullet$]
      \item If $w(\QBlocks)\leq k$, every problem $P$ in \QBlocks would be optimally
        solvable in polynomial time by \iw{k}, Theorem~\ref{thm:iw:Q}.
        %(time exponential in $k$, but $k$ is fixed for any $P$ in \QBlocks).
        Since computing optimal plans for arbitrary instances of Blocksworld is NP-hard
        \cite{chenoweth:blocks:np-hard,gupta-nau:blocks:np-hard}, the width of
        \QBlocks must be \textbf{unbounded,} unless $\text{P}=\text{NP}$.
      \item Any transition in a Blocksworld instance flips at most 5 atoms.
        Hence, since $w(\QClear)=1$ (resp.\ $w(\QOn)=2$), \iw{1} (resp.\ \iw{2})
        is guaranteed to find an optimal plan for any problem $P$ in $\QClear$
        (resp.\ $\QOn$) with $N$ ground atoms in $\O(N)$ time and space
        (resp.\ $\O(N^3)$ time and $\O(N^2)$ space).
      \Omit{
      \item Transitions for problems in $\QGrid$ and $\QGridXY$ also flips a constant
        number of atoms.
        Hence, \iw{1} (resp.\ \iw{2}) is guaranteed to find an optimal plan for
        any problem in the class in linear time and space (resp.\ $\O(N^3)$ time
        and $\O(N^2)$ space), where $N$ is the number of objects in the problem.
      }
    \end{enumerate}
  \end{boxed-example}
}

%% On the other hand, a problem $P$ may be solvable and yet IW not being
%% able to find a plan for $P$. This happens when $w(P)=\infty$ because there is
%% no admissible set $T$ of atom tuples for $P$; see the Marbles example below.

From now on, we assume that the actions in the instances for a class
$\Q$ flip at most a constant number of atoms, which is actually
the case when the instances come from a STRIPS domain. %defined by action schemas of bounded arity.

\section{General Policies}

The (general) policies over a class $\Q$ of problems $P$ are
first defined \textbf{semantically,} then \textbf{syntactically.}
Semantically, a policy $\pi$ is regarded as a relation on state
pairs $(s,s')$ that is only true for state transitions.
A state transition is a pair of states $(s,s')$ such that there is
an action $a$ that is applicable in $s$ and which  maps $s$ into $s'$.
The reason for policies to be defined as relations on state pairs and
not on state-action pairs  (e.g., as in RL and MDPs)
is that the set of actions changes across the instance in $\Q$. 
Policies $\pi$ expressed as relations on state transitions $(s,s')$
specify the possible actions to do in $s$ indirectly and
non-deterministically \cite{bonet:ijcai2018}: if $(s,s')$ is a
state transition in the relation $\pi$, \emph{any} action $a$ that
maps $s$ into $s'$ is allowed by $\pi$ and can thus be applied at the
state $s$.

\begin{definition}[Policies]
  \label{def:policies}
  A \textbf{policy} $\pi$ for a class of problems $\Q$ is
  a \textbf{binary relation} on $\cup_{P\in\Q}\states(P)$; i.e., on the
  reachable states of the problems in $\Q$, such that
  the state pair $(s,s')$ in $P$  is in $\pi$ only if $(s,s')$
  is a \emph{state transition} in $P$.
  Furthermore,
  \begin{enumerate}[1.]
    \item A state transition $(s,s')$ in a problem $P$ is a \textbf{$\pi$-transition}
      if $(s,s')\in\pi$, and $s$ is \textbf{not} a goal state in $P$.
    %\item A pair of states $(s,s')$ over a problem $P$ is a \textbf{$\pi$-transition} if $(s',s)\in\pi$, and $s$ is not a goal in $P$.
    %\item A \textbf{$\pi$-sequence} on $P$ is a sequence $\tau=s_0,s_1,\ldots,s_n$ of
    %  states in $P$ such that $(s_i,s_{i+1})$ is a $\pi$-transition for $0\leq i<n$.
    %\item A \textbf{$\pi$-trajectory} on $P$ is a $\pi$-sequence that is also a state trajectory in $P$.
    \item A \textbf{$\pi$-trajectory} in $P$ is a sequence $s_0,s_1,\ldots,s_n$ of
      states in $P$ such that $(s_i,s_{i+1})$ is a $\pi$-transition, $0\leq i<n$,
      and $s_0$ is the initial state of $P$.
    \item A $\pi$-trajectory $s_0,s_1,\ldots,s_n$ in $P$ is \textbf{maximal} if there
      are no $\pi$-transitions $(s_n,s)$ in $P$, or $s_n=s_i$ for some $0\leq i<n$.
      In the latter case, the trajectory is \textbf{cyclic.}
    \item The policy $\pi$ \textbf{solves} $P$ if \textbf{every maximal} $\pi$-trajectory
      $s_0,s_1,\ldots,s_n$ ends in a goal state (i.e., $s_n$ is a goal state).
    \item The policy $\pi$ solves $P$ \textbf{optimally} if \textbf{every maximal}
      $\pi$-trajectory reaches a goal state in $n$ steps, where $n$ is the cost of $P$.
    \item The policy $\pi$ \textbf{solves} $\Q$ if $\pi$ \textbf{solves} each problem
      $P$ in $\Q$, and it is \textbf{optimal} for $\Q$ if it solves each problem in
      $\Q$ \textbf{optimally.}
    %\item Relation $\pi$ \textbf{solves} $\Q$ iff $\pi$ solves every problem in $\Q$.
    %  The following better moved to where it is used...
    %\item The collection $\Q$ is \textbf{$\pi$-closed} iff for any \textbf{$\pi$-sequence} $\tau=s_0,s_1,\ldots,s_n$
    %  on a problem $P$ in $\Q$, there is a problem $P'$ in $\Q$ such that $\tau$ is a \textbf{$\pi$-trajectory}
    %  seeded at the initial state of $P'$.
    %\item Relation $\pi$ is a \textbf{policy} for $P$ (resp.\ $\Q$) iff it solves $P$ (resp.\ $\Q$).
    %  \alert{(Notice that policy by definition already solves $P$ (or $\Q$)... so don't need to qualify this later)}
  \end{enumerate}
  %Notice that $\pi$-trajectories cannot go past goal states by definition.
\end{definition}

Sufficient and necessary conditions for a general policy $\pi$ to solve a
class of problems $\Q$ can be expressed with suitable notions that apply to
each of the instances in $\Q$:
%\hector{*** I removed dead-ends and safe here; not needed. Please see and fix  occurrences of safe in Examples, proofs, etc ***}

\begin{definition}[Policy concepts]
  \label{def:policy:concepts}
  Let $\Q$ be a class of problems, and let $\pi$ be a policy for $\Q$.
  Then, %\hector{modified these three items a bit to use $\pi$ to make iff  below true}
  \begin{enumerate}[1.]
    \item $\pi$ is \textbf{closed} in $\Q$ if there is a $\pi$-transition $(s,s')$
      from every $\pi$-reachable state $s$ in $P\in\Q$ that is not a \textbf{goal} state.
      %or a \textbf{dead-end} state.
    %\item $\pi$ is \textbf{safe} in $\Q$ if there is no $\pi$-transition $(s,s')$
      %from a $\pi$-reachable, \textbf{non-dead} state $s$ in $P\in\Q$ to a
      %\textbf{dead-end} state.
    \item $\pi$ is \textbf{acyclic} in $\Q$ if there is no cyclic  $\pi$-trajectory
      starting in an initial state for $P\in\Q$.
  \end{enumerate}
\end{definition}

\begin{theorem}[Requirements for solvability]
  \label{def:policy:solves}
    %A policy $\pi$ solves a class of problems $\Q$ iff $\pi$ is closed, safe, and acyclic in $\Q$.
    A policy $\pi$ solves a class of problems $\Q$ iff $\pi$ is closed  and acyclic in $\Q$.
\end{theorem}
\Proof{%
  Direct.
  If $\pi$ is closed and acyclic, $\pi$ solves every problem in $\Q$.
  Conversely, if $\pi$ is either not closed or acyclic, % in $\Q$,
  there is at least one problem $P$ in $\Q$ that $\pi$ does not solve.
}

\Example{
  \begin{boxed-example}[General (semantic) policy for \QClear]
    \label{ex:pi:clear:semantic}
    \begin{enumerate}[$\bullet$]
      \item Let $\pi$ be the {general policy} for \QClear defined as the set
        of transitions $(s,s')$ such that:
        \begin{enumerate}[$a$)]
          \item for a block $y$ \textbf{above} $x$, $s\vDash \text{clear}(y) \land \text{hand-empty}$
            and $s'\vDash\text{hold}(y)$, or
             % (notice that as $y$ is above $x$, $\text{clear}(x)$ must    be false in $s$), or
          \item $s\vDash\text{hold}(y)$ and $s'\vDash\text{ontable}(y)$.
        \end{enumerate}
        %a transition $(s,s')$ is a $\pi$-transition {iff}
        %A)~ in $s$, $clear(x)$ is false,  $clear(y)$ is true for $y$  above $x$, and  \emph{hand-empty}  is true, a and in $s'$, $clear(x)$ is false and $hold(y)$ is  true,
        %B)~in $s$, $clear(x)$ is false  and $hold(y)$ is true for some $y$, and in $s'$, $ontable(y)$ s true.

        To show that $\pi$ solves all the instances in \QClear one needs to show that $\pi$ is closed and acyclic.
        For closedness, it is easy to see that in every reachable state $s$ in an instance $P$ in $\Q$ that is not a goal
        state, there are successors $s'$ for which condition ($a$) or ($b$) holds.
        %Indeed, A or B will apply according to whether \emph{hand-empty} is true or false in $s$.
        %In the first case, a state $s'$, where $hold(y)$ is true can be achieved by a pick-up action, and in the second case, by a putdown action.
        %Safeness is direct as there are no dead-end states in Blocksworld.
        Finally, $\pi$ is acyclic because in any transition of a $\pi$-trajectory, one of the blocks that is initially
        above $x$ is unstacked or  placed on the table, and no block is ever stacked above $x$, so every $\pi$-trajectory
        must be finite.
    \end{enumerate}
  \end{boxed-example}
}

\subsection{Rule-based Policies}

General policies are relations over the state pairs that are state transitions,
and following \citeay{bonet:ijcai2018}, these relations can be defined in a compact way by means of rules of
the form $C \mapsto E$ over a fixed set $\Phi$ of Boolean and numerical features over $\Q$.

\begin{definition}[Features]
  \label{def:features}
  A Boolean (resp.\ numerical) feature for a problem $P$ is a function
  that maps reachable states into Boolean values (resp.\ non-negative integers).
  A feature for a class $\Q$ of problems is a feature that is defined
  on each problem $P$ in $\Q$.
\end{definition}

\noindent
When providing time bounds, we will be interested in \textbf{linear features}
defined as follows:

\begin{definition}[Linear features]
  \label{def:features:linear}
  A feature $\phi$ for problem $P$ with $N$ ground atoms is \textbf{linear}
  if for any reachable state $s$ in $P$, the value $\phi(s)$ can be computed
  in $\O(N)$ time (i.e., linear time), and if $\phi$ is numerical, the size
  of $\{\phi(s):s\in\states(P)\}$ is $\O(N)$.
  A set $\Phi$ of features is a \textbf{set of linear features} for $P$ if
  each $\phi$ in $\Phi$ is linear for $P$, and it is a set of linear features
  for a class $\Q$, if $\Phi$ is a set of linear features for each problem
  $P$ in $\Q$.
\end{definition}

\noindent
The form of the rules that make use of the features is as follows:

\begin{definition}[Rules]
  \label{def:rules}
  Let $\Phi$ be a set of features. A $\Phi$-rule is a rule of the form $C\mapsto E$
  where the \textbf{condition} $C$ is a set (conjunction) of Boolean feature conditions
  and the \textbf{effect} $E$ is a set (conjunction) of feature value changes.
  A Boolean feature condition is of the form $p$, $\neg p$, $n=0$, and $n>0$
  for Boolean and numerical features $p$ and $n$ in $\Phi$.
  Feature value changes are of the form $p$, $\neg p$, $p?$ for Boolean $p$,
  and $\DEC{n}$, $\INC{n}$, and $\UNK{n}$ for numerical $n$.
%   Additionally, for numerical features $n$, $E$ may contain  $\EQ{n}$ or $\GT{n}$, if $\DEC{n}$ is in $E$.
  %\alert{** Following NeurIPS paper, we can add effects $\EQ{n}$ and $\GT{n}$
  %for numericals $n$ that are decreased. This allows to distinguish decrements
  %that reach zero from those that don't **}
  %%The condition $\GT{n}$ is assumed in rules with effects $\DEC{n}$ or $\UNK{n}$.
  %\alert{(Removed requirement $\GT{n}\in C$ when $E$ contain $\DEC{n}$ or $\UNK{n}$,
  %as a decrement can only across a transition if the feature is greather than zero initially)}
\end{definition}

A collection of rules defines a binary relation over state transitions in $\Q$.
While we consider \textbf{rule-based policies} first that define policies,
we will also use rules to define another  type of binary relation, \textbf{serializations}.
While policies select  state transitions $(s,s')$ from states $s$, serializations
select state pairs $(s,s')$ where $s'$ is not necessarily reachable from $s$
through a single action:

\begin{definition}
  \label{def:pi:transition}
  A state pair $(s,s')$ \textbf{satisfies} the condition $C$ of a $\Phi$-rule
  $C \mapsto E$ if the feature conditions in $C$ are all true in $s$.
  The transition $(s,s')$ \textbf{satisfies} the effect $E$ if the values for
  the features in $\Phi$ change from $s$ to $s'$ according to $E$; i.e.,
  \begin{enumerate}[1.]
    \item if $p$ (resp.\ $\neg p$) is in $E$, then $p(s')=1$ (resp.\ $p(s')=0$),
    \item if $\DEC{n}$ (resp.\ $\INC{n}$) is in $E$, $n(s)>n(s')$ (resp.\ $n(s)<n(s'))$,
    \item if $p$ (resp.\ $n$) is not mentioned at all in $E$, $p(s)=p(s')$ (resp.\ $n(s)=n(s')$), and
    %\item if $\EQ{n}$ (resp.\ $\GT{n}$) is in $E$, $n(s')=0$ (resp.\ $n(s')>0$).
  \end{enumerate}
  The state pair $(s,s')$ is \textbf{compatible} with a rule $C\mapsto E$
  if $s$ satisfies $C$, and the pair satisfies $E$. The pair is then also said to
  satisfy the rule itself. The state pair is compatible with a \textbf{set} $R$ of rules
  if it is compatible with at least one rule in the set.
  The pair $(s,s')$ is then said to be an $R$-pair, or to be \textbf{in} $R$.
\end{definition}

\noindent
A set of rules $R$ defines a policy in the following way:

\begin{definition}[Rule-based policies]
  \label{def:policies:rules}
  Let $\Q$ be a collection of problems and let $\Phi$ be a set of features for $\Q$.
  A set of rules $R$ over $\Phi$ defines the policy $\pi_R$ for $\Q$ in which a
  state pair $(s,s')$ over an instance $P \in \Q$ is in $\pi_R$ iff $(s,s')$
  is a state transition in $P$ that satisfies a rule in $R$.
  The set of rules $R$ solves $\Q$ iff the policy $\pi_R$ solves $\Q$.
  %Thus, $\prec_R$ is a policy for $P$ (resp
  %and thus a \textbf{policy} for $\Q$, given by: $s'\prec_R s$ iff the pair $(s,s')$ is compatible with $R$.
  %We often denote the policy $\prec_R$ simply with $R$.
\end{definition}

%\alert{** In the examples 4-6: follow similar structure or at least provide similar
%info; e.g., features are linear; width of class is X; here a rule based policy,
%policy $\pi_R$ solves class. **}

\Example{
  \begin{boxed-example}[General policy for \QClear]
    \label{ex:pi:clear}
    \begin{enumerate}[$\bullet$]
      \item A general policy for \QClear can be expressed using rules over
        the set $\Phi=\{H,n\}$ of two features, where $H$ is the Boolean feature that
        is true when the gripper holds a block, and $n$ is the numerical feature that
        counts the number of blocks above $x$. The policy has the two rules:
        % DON'T NEED TO INCLUDE \GT{n} IN SECOND RULE
        \begin{alignat*}{1}
          &\prule{\neg H, \GT{n}}{H, \DEC{n}} \,, \\
          &\prule{H}{\neg H} \,.
          %&\prule{H, \GT{n}}{\neg H} \,.
        \end{alignat*}
        The first rule says that when the gripper is empty and there are blocks above $x$,
        an action that decreases $n$ and makes $H$ true must be chosen, while the second
        rule says that when the gripper holds a block, an % and there are blocks above $x$, an
        action that makes $H$ false and does not affect $n$ must be selected.
        This policy is  slightly more general than the one in the previous example
        as a block being held can be put ``away'' in any position except above $x$.
      \item As we saw before, $w(\QClear)=1$ and this policy solves any instance $\QClear$.
    \end{enumerate}
  \end{boxed-example}
  \begin{boxed-example}[The Grid domain]
    \label{ex:pi:grid}
    \begin{enumerate}[$\bullet$]
      \item The Grid domain involves an agent that moves in a rectangular grid of
        arbitrary but finite size, and its goal is to reach a specific cell in the
        grid. Any instance in Grid is encoded with objects for the cells in the
        grid, and a unary predicate $pos(c)$ that is true when the position of
        the agent is cell $c$. The class of Grid problems is denoted by \QGrid.

        The $\text{Grid}_2$ domain is like Grid except that positions are encoded
        with horizontal and vertical coordinates. For a $m\times n$ grid,
        there are $m$ (resp.\ $n$) objects to encode the vertical (resp.\ horizontal)
        coordinates, and two unary predicates $hpos(h)$ and $vpos(v)$ to encode
        that the agent is at column $h$ and row $v$ in the grid.
        The class of $\text{Grid}_2$ problems is denoted by \QGridXY.
      \item Observe that each reachable state in a problem $P$ in \QGrid (resp.\ \QGridXY)
        makes true exactly 1 (resp.\ 2) atoms. Since all problems are solvable, $w(\QGrid)\leq 1$ and $w(\QGridXY)\leq 2$.
        On the other hand, problems in \QGrid cannot be solved in one step, in general,    and thus $Q(\QGrid)=1$.
        We show below that $w(\QGridXY)>1$ which  implies $w(\QGridXY)=2$.
      \item For either encoding, an optimal general policy can be expressed with a
        single rule over the singleton $\Phi=\{d\}$ of features where $d$
        measures the distance to the goal cell:
        \begin{alignat*}{1}
          &\prule{\GT{d}}{\DEC{d}}\,.
        \end{alignat*}
        Clearly, any transition $(s,s')$ compatible with the rule moves the agent one
        step closer to the goal, and the policy solves  optimally any problem in
        \QGrid or \QGridXY. The feature $d$ is linear in \QGrid but quadratic in \QGridXY.
    \end{enumerate}
  \end{boxed-example}
  \begin{boxed-example}[The Delivery domain]
    \label{ex:pi:delivery}
    \begin{enumerate}[$\bullet$]
      \item The Delivery domain $D$ involves an agent that moves in a rectangular
        grid, and packages  spread over the grid that can be picked up or
        dropped  by the agent, which can hold  one package at a time.
        A state for the domain thus specifies the position of the agent (i.e.,
        a cell on the grid), and positions for the different packages,
        where the position of a package can be either a cell in the grid, or the
        agent's gripper.
        The task of the agent is to ``deliver'' all  packages, one at a time, to a
        designated ``target'' cell.

        The position of the agent can be encoded using a single unary predicate
        as for the $Grid$ domain, or two unary predicates as for the $Grid_2$ domain.
        %where the predicates encode the coordinates of the agent along the rectangular axes.
        For simplicity, we assume an encoding of positions using a single predicate.

        An instance for Delivery consists of objects for the different cells in
        the grid, and objects for the different packages.
        The atoms are as follows: $pos(c)$ (resp.\ $ppos(p,c)$) that is true when
        the agent (resp.\ the package $p$) is in cell $c$, $holding(p)$ that is
        true when the agent is holding package $p$, and $empty$ that is true when
        the agent holds no package.
        Under this encoding, the goal can be expressed as the conjunction $\bigwedge_i ppos(p_i,target)$
        where $p_i$ is the $i$th package, $target$ is the target cell,
        and the conjunction is over all the packages.
      \item The class of all Delivery problems in denoted by \QDelivery, while \QDeliveryS
        denotes the class of Delivery problems with exactly one package.
        Below we show that the width of \QDelivery is \textbf{unbounded,}
        and $w(\QDelivery)>1$.
        On the other hand, if $s$ is a state in a problem $P$ in $\QDeliveryS$,
        $s$ either makes true exactly two atoms of form $pos(c)$ and $holding(p)$,
        for the unique package $p$, or makes true exactly three atoms of form
        $pos(c)$, $ppos(p,c)$, and $empty$.
        However, the atom $empty$ is determined by the atoms $ppos(p,c)$ and
        $holding(p)$: $holding(p)\Rightarrow\neg empty$, and $ppos(p,c)\Rightarrow empty$
        for any cell $c$.
        This fact allows the construction of an admissible set $T$ of size 2 for
        any problem $P$ in \QDeliveryS. Hence, $w(\QDeliveryS)=2$.
      \item A set of meaningful features for Delivery is $\Phi=\{H,p,t,u\}$ that
        capture whether the agent is holding a package, the
        distance to a nearest \emph{undelivered package} (zero if the agent is
        holding a package, or no package to be delivered remains), the distance
        to the target cell, and the number of undelivered packages,  respectively.
        The following set $R$ of rules define a policy $\pi_R$:
        %Let us consider the rule-based policy $\pi_R$ given by the  set of rules $R$:
        \begin{alignat*}{1}
          &\prule{\neg H,\GT{p}}{\DEC{p},\UNK{t}}\,, \\
          &\prule{\neg H, \EQ{p}}{H}\,, \\
          &\prule{H,\GT{t}}{\DEC{t}}\,, \\
          &\prule{H, \EQ{t}, \GT{u}}{\neg H, \DEC{u}, \UNK{p}}\,.
        \end{alignat*}
        The first rule says that when holding no package and undelivered packages
        remain in the grid, the agent must move closer to a nearest undelivered package.
        The second that when the agent holds no package and is at the same cell
        of an undelivered package, the agent must pick the package.
        The third that when holding a package and the agent is not at the target cell,
        it must move closer to the target cell.
        The last rule says that when the agent holds a package and is at the
        target cell, it must drop the package at the cell (i.e., deliver it).

        It is not difficult to see that the policy $R$ solves any problem $P$ in
        both  \QDelivery and  \QDeliveryS. % , even though the width of the former is unbounded.
        \Omit{Variants of the Delivery package that can be solved essentially with the
        rules over a slightly changed semantics for the features include problems
        that contain ``irrelevant packages'', where different packages may need
        to be delivered at different target cells.}
    \end{enumerate}
  \end{boxed-example}
}

\section{Envelopes}

We address next the  relation between general policies and problem width
by introducing  the notion of \emph{envelopes.}
An envelope for a binary relation $\mu$ over the states in a problem $P$
is a set of reachable states $E$ that obeys certain closure properties:

\begin{definition}[Envelopes]
  \label{def:envelope}
  Let $P$ be a problem, and let $\mu$ be a binary relation on $\states(P)$.
  A subset $E\subseteq\states(P)$ is an \textbf{envelope of $\mu$ for $P$},
  or simply a \textbf{$\mu$-envelope} iff
  \begin{enumerate}[1.]
    \item the initial state $s_0$ of $P$ belongs to $E$, and
    \item if $s$  is a non-goal state in $E$, there is a state $s'$ in $E$
      such that $(s,s')$ is a $\mu$-transition; i.e., a state transition
      in $P$ that is also in $\mu$.
  \end{enumerate}
  The envelope is \textbf{backward-closed}, abbreviated as \textbf{closed},
  iff for each state $s'$ in $E$ that is not the initial state, there is a
  state $s$ in $E$ such that $(s,s')$ is a $\mu$-transition.
  $E$  is an \textbf{optimal envelope} iff each maximal $\mu$-trajectory
  contained in $E$ and that starts at a non-goal state $s$ in $E$ is a
  suffix of an \textbf{optimal trajectory} for $P$.
\end{definition}

Notice that if $\pi$ is a policy that solves $P$, the set of states
reached by $\pi$ in the way of the goal form a $\mu$-envelope for
$\mu=\pi$, and similarly, the set of states in \emph{any}
non-empty set of $\pi$-trajectories also form a $\mu$-envelope.
These envelopes are actually closed, and indeed, for any state $s$
in these envelopes there is a $\mu$-trajectory that reaches the goal
and passes through $s$. If the policy $\pi$ is optimal for $P$, the
envelopes based on the relation $\mu=\pi$ are optimal too.

Envelopes can be defined in ways that do not involve policies.
In particular, \emph{cost-envelopes} are defined in terms of a binary
relation $\mu=\pcost$ that appeals to cost considerations.
For this, optimal state costs and the binary  $\pcost$ relation
are defined as follows:

\begin{definition}[Optimal state costs and cost relation]
  \label{def:pcost}
  The cost of a state $s$ in $P$, $\cost(s)$, is the cost (length) of
  a min-cost plan for reaching $s$ from the initial state $s_0$,
  and $\infty$ if there is no such plan.
  The \textbf{optimal cost} of a state $s$ in $P$, $\ocost(s)$,
  is $\cost(s)$ if $s$ is a \textbf{non-goal state}, and the cost
  of $P$ (i.e., the cost of an optimal plan for $P$) if $s$ is a
  \textbf{goal state}.
  %if such a plan exists, and $\infty$ if there are no plans for $P$.
  %%The \textbf{optimal cost} of a \textbf{non-goal state} $s$ in $P$,
  %%$\ocost(s)$, is $\cost(s)$.
  %%%initial state $s_0$ of $P$, and $\infty$ is there is no such plan.
  %%The optimal cost of a \textbf{goal state} $s$ is the cost (length) of
  %%a min-cost plan for reaching a goal state (not necessarily $s$), and
  %%$\infty$ is there is no such plan.
  The \emph{cost relation} $\pcost$ is the set of state pairs $(s,s')$
  from $P$ such that $\ocost(s)<\ocost(s')<\infty$.
  %\hector{*** Isn't it the other way around: $\ocost(s')<\ocost(s)<\infty$ ? ***}
  %In particular, no $\pcost$-trajectory can go through or end in a
  %state $s$ with $\ocost(s)=\infty$.
  If $(s,s')$ is in $\pcost$, we write $s \pcost s'$.
\end{definition}

The reason that the optimal cost of goal states is defined as the cost of
the problem $P$ is to have a correspondence between goal reaching trajectories
that are optimal and $\pcost$-trajectories: 

\Omit{
  \begin{theorem}[Envelopes for acyclic relations]
    \label{thm:envelope:acyclic}
    Let $P$ be a problem, let $\mu$ be a binary relation on
    $\states(P)$, and let $E$ be a $\mu$-envelope for $P$.
    If $\mu$ is acyclic in $P$, then for each state $s_1$ in $E$,
    1)~there is a $\mu$-trajectory $s_1, \ldots, s_n$ contained in $E$ where $s_n$ is a goal state, and
    2)~all maximal $\mu$-trajectories from $s_1$ contained in $E$ reach a goal state.
  \end{theorem}
  \Proof{%
    Let $E$ be a $\mu$-envelope, and let $s_1$ be a state in $E$.
    If $s_1$ is not a goal state, by definition of envelope,
    there is a state $s_2$ in $E$ such that $(s_1,s_2)$ is a
    $\mu$-transition (i.e., the pair $(s_1,s_2)$ is a transition
    in $P$ and it is also in $\mu$). Iterating we find a $\mu$-trajectory
    $s_1,s_2,\ldots$ that either ends in a goal state $s_n$ in $P$,
    or it is infinite. Since $\mu$ is acyclic in $P$, and the number
    of states in $P$ is finite, the trajectory cannot be infinite.
    Therefore, the trajectory is finite and ends in a goal state.
    Likewise, if $s_1,s_2,\ldots$ is a maximal $\mu$-trajectory
    contained in $E$, it must be finite and must end in a goal
    state since otherwise it can be further extended with a
    $\mu$-transition.
  }
}

\begin{lemma}[$\pcost$-trajectories]
  \label{lemma:pcost}
  Let $P$ be a planning problem with initial state $s_0$, and
  let $\tau=s_0,s_1,\ldots,s_n$ be a \textbf{goal-reaching}
  state trajectory in $P$.
  Then, $\tau$ is a $\pcost$-trajectory iff $\tau$ is an optimal trajectory for $P$.
\end{lemma}
\Proof{%
  \textbf{($\Rightarrow$)}
  Let us assume that $\tau$ is a $\pcost$-trajectory. We use induction
  on $i$ to show that the prefix $\tau_i\doteq s_0,s_1,\ldots,s_i$ is
  an optimal trajectory for $s_i$, $0\leq i\leq n$.
  The claim is true for $i=0$.
  Let us assume that it holds for $i=k$, and let us consider the
  trajectory $\tau_{k+1}$ leading to state $s_{k+1}$.
  By assumption, $\ocost(s_k)<\ocost(s_{k+1})$.
  On the other hand, the existence of the trajectory implies
  $\cost(s_{k+1}) \leq 1+\cost(s_k)$.
  Therefore, $\cost(s_{k+1})=1+\cost(s_k)$ and the trajectory
  $\tau_{k+1}$ is optimal for $s_{k+1}$.
  Finally, $\tau$ must be an optimal trajectory for $P$ since otherwise
  $\ocost(s_n)<\cost(s_n)=1+\cost(s_{n-1})=1+\ocost(s_{n-1})$
  which implies $\ocost(s_n)\leq \ocost(s_{n-1})$, and thus
  $s_{n-1}\not\pcost s_n$.

  \medskip\noindent
  \textbf{($\Leftarrow$)}
  Let us assume that $\tau$ is an optimal trajectory for $P$.
  By the principle of optimality, the trajectory $\tau_i$ must
  be optimal for $s_i$, $0\leq i<n$.
  Therefore, $\cost(s_0)=0$ and $\cost(s_{i+1})=1+\cost(s_i)$, $0\leq i<n$.
  Observe that $s_n$ must be a \textbf{closest} goal state to $s_0$
  and thus $\ocost(s_n)=\cost(s_n)$.
  Hence, $\tau$ is a $\pcost$-trajectory.
}

A property of cost-envelopes, i.e., $\pcost$-envelopes, is that they are
\emph{optimal}, and that  optimal $\mu$-envelopes are cost-envelopes
independently of the relation $\mu$:

\begin{theorem}[Optimality of cost-envelopes]
  \label{thm:cost-envelope:optimality}
  Let $P$ be a planning problem, and let $E$ be a subset of reachable states in $P$.
  If $E$ is an optimal envelope for some binary relation $\mu$ on $\states(P)$,
  then it is a cost-envelope.
  If $E$ is a cost-envelope, then it is an optimal envelope.
  %some binary relation $\mu$ on $\states(P)$.
  %Then, $E$ is a closed and optimal envelope iff $E$ is a cost-envelope.
  %$E$ is a cost-envelope for $P$ iff $E$ is an optimal envelope.
\end{theorem}
\Proof{%
  \textbf{($\Rightarrow$)}
  Let us assume that $E$ is an optimal envelope for some relation $\mu$.
  We need to show that $E$ satisfies the two conditions in Definition~\ref{def:envelope}
  for the relation $\pcost$.
  The first condition is direct since the initial state $s_0$ of $P$ belongs
  to $E$ as $E$ is a $\mu$-envelope.
  For the second condition, let $s$ be a non-goal state in $E$.
  By the optimality of $E$, there is an optimal trajectory for $P$ of form
  $\tau=s_0,s_1,\ldots,s_\ell,\ldots,s_n$ such that $s_\ell=s$ and
  $\{s_\ell,s_{\ell+1},\ldots,s_n\}\subseteq E$.
  Since $\tau$ is optimal, $\ocost(s_i)=i$ for $i=0,1,\ldots,n$, and thus
  the transition $(s,s_{\ell+1})$ is a $\pcost$-transition with $s_{\ell+1}\in E$.

  \medskip\noindent
  \textbf{($\Leftarrow$)}
  Let us assume that $E$ is a cost-envelope, and let $\tau_1=s_\ell,s_{\ell+1},\ldots,s_n$
  be a maximal $\pcost$-trajectory contained in $E$ that starts at state $s_\ell$.
  In particular,
  \[ \ocost(s_\ell)\ <\ \ocost(s_{\ell+1})\ <\ \cdots\ <\ \ocost(s_n)\ <\ \infty \,. \]
  Since $s_\ell$ is a reachable state in $P$, there is an optimal trajectory
  $\tau_2=s_0,s_1,\ldots,s_\ell$ from $s_0$ to the state $s_\ell$ which
  satisfies, by its optimality,
  \[ \ocost(s_0)\ <\ \ocost(s_1)\ <\ \cdots\ <\ \ocost(s_\ell) \,. \]
  Therefore, the combined trajectory $\tau=\tau_2,\tau_1$ is a goal-reaching
  $\pcost$-trajectory. By Lemma~\ref{lemma:pcost}, $\tau$ is an optimal
  trajectory for $P$ that contains $\tau_1$ as a suffix.
  Since $\tau_1$ is arbitrary, the envelope $E$ is optimal.
}

We will see that a relation between the width of a problem and optimal
policies can be established by considering cost-envelopes defined  by tuples
of atoms.

\subsection{Cost Envelopes and Problem Width}

For a set $T$ of \textbf{reachable atom tuples} over $P$, the set $OPT(T)$
is the set of min-cost states that reach the tuples in $T$:

\begin{definition}[Optimal states for $T$]
  \label{def:optimal-states}
  Let $P$ be a problem,
  let $t$ be a reachable atom tuple in $P$, and
  let $T$ be a set of such tuples.
  $\OPT(t)$ is the set of \textbf{optimal states} for $t$ in $P$,
  i.e., the set of min-cost states where $t$ is true, and
  $\OPT(T) \doteq \cup\{\OPT(t):\text{tuple $t$ is in $T$}\}$.
\end{definition}

\noindent
It turns out that $OPT(T)$ is a cost-envelope iff $T$ is admissible:
%\hector{Why you say ``contains''? Looks that it is not needed; this comes out somewhere else too.}
%% The optimal set of states for $T$, $\OPT(T)$, is a cost-envelope iff $T$ contains an admissible set of tuples:

\begin{theorem}[Admissibility and cost-envelopes]
  \label{thm:cost-envelope:admissible}
  Let $P$ be a planning problem, and let $T$ be a set of reachable
  atom tuples in $P$.
  Then, $T$ is admissible for $P$ iff $OPT(T)$ is a cost-envelope.
\end{theorem}
\Proof{
  \textbf{($\Rightarrow$)}
  Let us assume that $T$ is admissible for $P$, and let $s_0$ be the
  initial state of $P$.
  We need to show that $\OPT(T)$ is a cost-envelope:
  %namely, that the states
  %in $\OPT(T)$ are reachable from $s_0$, and that the two conditions in
  %Def.~\ref{def:envelope} hold for $\OPT(T)$ for the relation $\pcost$.
  %The reachability condition is fullfilled by definition of $\OPT(T)$.
  %For the other two conditions:
  \begin{enumerate}[1.]
    % Condition: s_0 belongs to \OPT(T')
    \item Since $T$ is admissible, there is tuple $t$ in $T$ with $s_0\vDash t$.
      Hence, $s_0$ belongs to $\OPT(T)$.
    % Condition: If s is non-goal state in \OPT(T'), there is \pcost-transition (s,s') with s' in \OPT(T')
    \item Let $s$ be a non-goal state in $\OPT(T)$; i.e., there is tuple $t$
      in $T$ with $s\vDash t$, and there is an optimal trajectory $\tau$ for $t$
      that ends in $s$.
      Using that $T$ is admissible, it is easy to show that $\tau$ can be
      extended into an optimal trajectory $\tau,s_1,\ldots,s_n$ for $P$,
      where the trajectories $\tau,s_1,\ldots,s_i$ are optimal for tuples $t_i$ in $T$;
      i.e., the states $s_1,s_2,\ldots,s_n$ all belong to $\OPT(T)$.
      On the other hand, by Lemma~\ref{lemma:pcost}, $\tau,s_1,\ldots,s_n$
      is a $\pcost$-trajectory.
      Hence, $(s,s_1)$ is a $\pcost$-transition.
      %%By Theorem~\ref{thm:width:optimal-plans}, $\tau$ can be extended into an
      %%optimal plan $\tau,s_1,s_2,\ldots,s_n$ for $P$ where the trajectory
      %%$\tau,s_1$ is optimal for some tuple $t'$ in $T$.
      %%Hence, the transition $(s,s_1)$ is a $\pcost$-transition by Theorem~\ref{thm:pcost}.
      %%Clearly, $s_1$ belongs to $\OPT(t')\subseteq\OPT(T)$.
  \end{enumerate}

  \noindent
  \textbf{($\Leftarrow$)}
  Let us assume that $\OPT(T)$ is a cost-envelope.
  We need to show that $T$ is admissible; namely, that the two conditions
  in Definition~\ref{def:admissible} hold for $T$.
  \begin{enumerate}[1.]
    \item Since $\OPT(T)$ is an envelope, it contains the initial state $s_0$
      of $P$. Hence, there is a tuple $t\in T$ such that $s_0\vDash t$.
    \item Let $\tau$ be an optimal trajectory for a tuple $t\in T$ that ends in a
      \textbf{non-goal} state $s$. We need to show that $\tau$ can be extended
      with a single step into an optimal trajectory for a tuple $t'\in T$.
      Since $\OPT(T)$ is a cost-envelope and $s\in\OPT(T)$, there is a
      $\pcost$-trajectory $s,s',\tau'$ \textbf{entirely} contained in $\OPT(T)$
      that starts at $s$, transitions to $s'$, and ends in a goal state.
      Hence, there is a tuple $t'\in T$ such that $s'\in\OPT(t')$, and
      the joined trajectory $\tau,s',\tau'$ is a goal-reaching $\pcost$-trajectory
      that starts at $s_0$.
      By Lemma~\ref{lemma:pcost}, the trajectory $\tau,s'$ is optimal for $s'$,
      and thus for the tuple $t'$ in $T$.
  \end{enumerate}
  \vskip -1.7em
}

Therefore, optimal plans for $P$ can be found by running the \iw{T} algorithm,
provided that \emph{there is} a subset of tuples $T' \subseteq T$ such that
$\OPT(T')$ is a cost-envelope:

\begin{theorem}
  \label{thm:cost-envelope:iw(T)}
  The algorithm \iw{T} finds an optimal plan for $P$ if $\OPT(T')$ is a
  cost-envelope for some $T' \subseteq T$. Likewise, \iw{k} finds an
  optimal plan for $P$ if $\OPT(T)$ is a cost-envelope for some set
  $T$ of conjunctions of up to $k$ atoms in $P$.
\end{theorem}
\Proof{%
  If $\OPT(T')$ is a cost-envelope, $T'$ is admissible by Theorem~\ref{thm:cost-envelope:admissible},
  and \iw{T} finds an optimal plan by Theorem~\ref{thm:iw(T)}.
  The second claim follows from the first and Theorem~\ref{thm:width:nir}.
}

The  \textbf{width} of a problem $P$ can thus be related to the min size of a
tuple set $T$ for which $\OPT(T)$ is a cost-envelope for $P$:

\begin{theorem}
  \label{thm:cost-envelope:width}
  Let $P$ be a planning problem, and let $T$ be a set of atom tuples in $P$.
  If $\OPT(T)$ is a cost-envelope, $w(P)\leq size(T)$.
  % Can't use equality because of $width(P)=0$ if $P$ solvable in 1 step.
\end{theorem}
\Proof{%
  By Theorem~\ref{thm:cost-envelope:admissible}, $T$ is admissible,
  and hence, by Definition~\ref{def:width}, $w(P)\leq\size(T)$.
}

\subsection{Optimal Policies and Problem Width}

A sufficient condition for $\OPT(T)$ to be a cost-envelope is for $\OPT(T)$
to be a $\pi$-envelope for an \textbf{optimal policy} $\pi$ for a class $\Q$ of problems
that includes the problem $P$:
% Closure property now commented isn't used.
%and the problems $P[s]$ for every state $s$ in $\OPT(T)$.
%Recall that $P[s]$ is a problem identical to $P$ but with initial state $s$:

\begin{theorem}[Envelopes for optimal policies]
  \label{thm:cost-envelope:optimal-pi}
  Let $P$ be a planning problem, %let $T$ be a set of atom tuples in $P$,
  let $\pi$ be an \textbf{optimal policy} for a class $\Q$ that includes $P$,
  and let $E$ be a subset of reachable states in $P$.
  %and the problems $P[s]$ for $s$ in $\OPT(T)$. %the states $s$ in $\OPT(T)$.
  %Then, $\OPT(T)$ is a cost-envelope in $P$ if $\OPT(T)$ is a \textbf{closed}
  %$\pi$-envelope in $P$.
  Then, $E$ is a cost-envelope in $P$ if $E$ is a \textbf{closed} $\pi$-envelope in $P$.
  %$\OPT(T)$ is a cost-envelope in $P$ if $\OPT(T)$ is a $\pi$-envelope in $P$ for an optimal policy $\pi$ for a class $\Q$
  %that includes $P$ and the problems $P[s]$ for each state $s$ in $\OPT(T)$.
\end{theorem}
\Proof{%
  By Theorem~\ref{thm:cost-envelope:optimality}, it is sufficient to show
  that $E$ is an optimal envelope. However, this is direct since for any
  maximal $\pi$-trajectory $\tau_1$ that is contained in $E$ and that starts
  at some state $s$, by closedness, there is a $\pi$-trajectory $\tau_2$
  from the initial state of $P$ to the state $s$.
  Hence, the trajectory $\tau=\tau_2,\tau_1$ is a goal-reaching $\pi$-trajectory
  from the initial state that must be optimal as $\pi$ is an optimal
  policy.
}

A closed $\pi$-envelope is an envelope formed by the states in a (non-empty)
set of $\pi$-trajectories, starting in the initial state of the problem and
ending at a goal state.
The theorem implies that if $\OPT(T)$ is a closed $\pi$-envelope, then $\OPT(T)$
is a cost-envelope, and then from Theorem~\ref{thm:cost-envelope:width}, that
the width of $P$ is bounded by the size of $T$:
%The condition in the above theorem implies two things: that $\pi$ must reach
%all optimal states $s \in OPT(t)$ for tuples $t$ in $T$, and that $\pi$ can
%reach the goal from all such states without leaving the envelope.
%The condition does not imply, on the other hand, that all $\pi$-trajectories
%must be inside the envelope $OPT(T)$.

%The relation between the width of problem $P$ and the existence of general
%policies for a class of problems $\Q$ that includes $P$ can then be expressed
%as follows.
%%A class of problems $\Q$ is said to be \textbf{closed} if it includes all problems
%%$P[s]$, when $P$ is in $\Q$ and $s$ is a reachable state in $P$ that is not a dead-dead.
%%All the classes of problems that we consider in the paper are closed in this sense:

%% For this, let us say that $\Q$ is a \textbf{closed set} of problems, if it includes $P[s]$
%% when it includes problem $P$, and $s$ is \textbf{alive} in $P$ (i.e., $s$ is a reachable state
%% in $P$ that is not a dead end).

\begin{theorem}[Optimal policies and width]
  \label{thm:main:width}
  %Let $P$ be a planning problem, let $T$ be a set of atom tuples in $P$,
  %and let $\Q$ be a closed set of problems that includes $P$.
  %The width of $P$ is bounded by $size(T)$, i.e.\ $w(P) \leq \size(T)$,
  %if $\OPT(T)$ is a closed $\pi$-envelope for an optimal policy $\pi$ for $\Q$.
  Let $\Q$ be a class of planning problems, and let $\pi$ be an optimal
  policy for $\Q$. Then,
  \begin{enumerate}[1.]
    \item If $P$ is a planning problem in $\Q$, and $T$ is a set of atom
      tuples in $P$ such that $\OPT(T)$ is a closed $\pi$-envelope, then
      $w(P)\leq \size(T)$.
    \item If $k$ is a non-negative integer such that for any problem $P$
      in $\Q$, there is a set $T$ of atom tuples in $P$ such that $\size(T)\leq k$
      and $\OPT(T)$ is a closed $\pi$-envelope, then $w(\Q)\leq k$.
  \end{enumerate}
\end{theorem}
\Proof{%
  The first claim follows by Theorems~\ref{thm:cost-envelope:width} and
  \ref{thm:cost-envelope:optimal-pi} as  the former establishes  that $w(P)\leq\size(T)$
  if $\OPT(T)$ is a cost envelope, and  the latter that $\OPT(T)$ is a cost envelope
  if it a $\pi$-envelope of an optimal policy $\pi$. %\hector{Check this line please}
  The second claim follows from the first since for any problem
  $P$ in $\Q$, $w(P)\leq k$ by the first claim and the assumed $T$.
}

This theorem is important as it sheds light on the notion of width and why
many standard domains have bounded width when atomic goals are considered.
Indeed, in such cases, the classes of instances $\Q$ admit optimal policies
$\pi$ that in each instance $P$ in $\Q$ can be
``followed'' by considering a set of tuples $T$ over $P$.
If $OPT(T)$ is a closed-envelope of $\pi$, it is possible to reach the goal
of $P$ optimally through a $\pi$-trajectory without knowing $\pi$ at all:
one can then pay attention to the set of tuples in $T$ only and just run
the \iw{T} algorithm. Indeed:

\begin{theorem}
  \label{thm:iwt:pi}
  Under the assumptions of Theorem~\ref{thm:main:width}, \iw{T} reaches the
  goal of $P$ optimally, through a $\pi$-trajectory; i.e., there is a
  goal-reaching $\pi$-trajectory $\tau=s_0,s_1,\ldots,s_n$ seeded at the initial
  state $s_0$ of $P$ such that all the states $s_i$ are expanded by \iw{T},
  except the goal state $s_n$ that is selected for expansion but is  not
  expanded.
\end{theorem}
\Proof{%
  As shown in the proof of Theorem~\ref{thm:main:width}, $\OPT(T)$ is a
  cost-envelope given the assumptions and $T$ is an admissible set for $P$.
  Therefore, by Theorem~\ref{thm:iw(T)}, \iw{T} finds an optimal path
  $\tau=s_0,s_1,\ldots,s_n$ for $P$; that is, \iw{T} expands nodes $n_i$
  that represent the prefixes $\tau_i=s_0,s_1,\ldots,s_i$ for $0\leq i<n$,
  and selects for expansion the node that represents the path $\tau$.
  Since $s_n$ is a goal state, \iw{T} returns the path $\tau$.
  %%If $OPT(T)$ is a closed-envelope, then it is a cost-envelope, and therefore
  %%\iw{T} reaches the goal of $P$ optimally \hector{invoke relevant theorems and polish this all}.
  %%It is easy to show that all tuples $t$ in $T$ are reached optimally by \iw{T} then.
  %%Let $s_t$ be the first state that reaches $t \in T$ in in the \iw{T} search
  %%and let us assume that $s_t$ is not a goal state.
  %%The state $s_t$ is optimal for $t$, i.e., $s_t \in OPT(t)$,  and therefore
  %%$\pi$ reaches $s_t$, and there must be a $\pi$-transition from $s_t$ to $s_{t'}$
  %%for some $t' \in T$ such that $s_{t'} \in OPT(t')$ as $OPT(T)$ is a $\pi$-envelope.
  %%This means that when \iw{T} select a state $s$ for expansion it has already
  %%expanded all the states in some  $\pi$-trajectory reaching $s$, and the same
  %%holds, when $s$ is a goal state, as state in the theorem.
}

When the conditions in Theorem~\ref{thm:main:width} hold, the \iw{T} algorithm
reaches the goal of the problem $P \in\Q$, optimally through a $\pi$-trajectory,
even if the policy $\pi$ or the features involved in the policy are not known.
We say in this case, that the set of tuples $T$ \textbf{represents} the policy
$\pi$ in the problem $P$, as the \iw{T} search delivers then a $\pi$-trajectory
to the goal, and it is thus ``following'' the policy.
Notice however that even in this case, it is not necessary for the set of tuples
$T$ to capture all the possible $\pi$-trajectories to the goal.
It suffices for $T$ to capture one such trajectory.
% at least, and indeed, all the
% trajectories that fall within the $OPT(T)$ envelope but no more.
The goal of $P$ then can be reached optimally through \iw{T} by expanding no more than
$|T|$ nodes (cf.\ Theorem~\ref{thm:iw(T):bounds}).

It is also not necessary for the set of tuples $T$ to be known for
solving the problem $P$ optimally. If the conditions in Theorem~\ref{thm:main:width}
hold for some set $T$ of tuples of size $k$, i.e., $k=size(T)$, the algorithm
\iw{k} will deliver an  optimal $\pi$-trajectory to the goal as well.
On the other hand, if there is an  upper bound $k$ on the size of the tuples,
but its value is not known, the algorithm IW would solve  the problem
in time and space exponential in $k$ but not necessarily through
an optimal  $\pi$-trajectory,  because non-optimal solutions
can potentially be found  by \iw{k'} when $k' < k$ (cf.\ Theorem~\ref{thm:iw}).

\subsection{Lower Bound on Width}

Finally, the following result provides a lower bound on  width:

\begin{theorem}[Necessary conditions for bounded width]
  \label{thm:width:necessity}
  Let $P$ be a planning problem, let $k$ be a non-negative integer, and let
  $T$ be the set of \textbf{all} atom tuples in $P$ of size at most $k$.
  If for every optimal trajectory $\tau$ for $P$, $\tau$ has a state that is
  \textbf{not} in $\OPT(T)$, then $w(P)>k$.
  %If $w(P)\leq k$, $\OPT(T^k)$ contains a closed $\pi$-envelope for some
  %policy $\pi$ that optimally solves $P$.
  %Consequently, if for every optimal plan $\tau$ for $P$, $\tau$ has a state
  %that is \textbf{not} in $\OPT(T)$, then $w(P)>k$.
\end{theorem}
\Proof{%
  We show the contrapositive of the claim. Namely, if $w(P)\leq k$, then there
  is an optimal trajectory $\tau$ for $P$ that is entirely contained in $\OPT(T)$.
  Thus, let $P$ be a problem such that $w(P)\leq k$, and let $T'$ be an
  admissible set for $P$ with $\size(T')\leq k$.
  We construct an optimal trajectory $\tau=s_0,s_1,\ldots,s_n$ inductively,
  using the admissible set $T'$.
  Indeed, $T'$ contains a tuple $t_0$ that is made true by the initial state $s_0$,
  and thus $s_0\in\OPT(T')$.
  For the inductive step, assume that we have already constructed the prefix
  $\tau_i=s_0,s_1,\ldots,s_i$ such that $\tau_i\subseteq\OPT(T')$ and $\tau_i$
  is an optimal trajectory for $s_i$; in particular. There is a tuple $t_i$ in $T'$
  such that $s_i\in\OPT(t_i)$.
  By admissibility, $\tau_i$ can be extended with one step into an optimal
  trajectory for a tuple $t'$ in $T'$; i.e., there is a state $s_{i+1}$ and tuple
  $t_{i+1}$ such that $s_{i+1}$ in $\OPT(t_{i+1})$ and $\tau_{i+1}\doteq\tau_i,s_{i+1}$
  is an optimal trajectory for $t_{i+1}$.
  By definition of admissible sets, this process can be continued until $\tau_i$
  becomes an optimal trajectory for $P$.
  At such moment, by construction, $\tau_i\subseteq \OPT(T')$.
  To finish the proof, observe that $T'\subseteq T$ since $\size(T')\leq k$.
}

\Example{
  \begin{boxed-example}[Width for the class \QOn]
    \label{ex:width:qon}
    \begin{enumerate}[$\bullet$]
      \item Let $P$ be a problem in \QOn where in the initial state the blocks
        $x$ and $y$ are not clear and in different towers.
      \item Let $T$ be the set of all atoms in $P$ (i.e., all atom tuples of size 1).
        Any optimal trajectory for $P$ contains a state $s$ in which both $x$ and $y$ are
        clear, just before moving $x$ on top of $y$.
        This state $s$ is not in $\OPT(T)$; e.g., it does not belong to $\OPT(clear(x))$
        because any optimal trajectory for $clear(x)$ does not move any block above $y$.
        By Theorem~\ref{thm:width:necessity}, $w(P)>1$ and thus $w(\QOn)>1$.
        As we saw before, $w(\QOn)\leq 2$. Hence, $w(\QOn)=2$.
    \end{enumerate}
  \end{boxed-example}
  \begin{boxed-example}[Width for Grid problems (classes \QGrid and \QGridXY)]
    \label{ex:width:grid}
    \begin{enumerate}[$\bullet$]
      \item The states in problems for \QGrid (resp.\ \QGridXY) make true
        exactly one (resp.\ two) atoms; i.e., the subset of atoms that identify the
        position of the agent.
      \item Let $\pi$ is an optimal policy for \QGrid (resp.\ \QGridXY), let $P$
        be a problem in \QGrid, and let $\tau$ be a goal-reaching $\pi$-trajectory
        seeded at the initial state of $P$.
        Let us consider the set of atom tuples $T\doteq\{\atoms(s):s\in\tau\}$ where
        $\atoms(s)$ refers to the subset of atoms made true at the state $s$.
        It is easy to show that $T$ is a closed $\pi$-envelope, and $\size(T)=1$
        (resp.\ $\size(T)=2$).
        Therefore, by Theorem~\ref{thm:main:width}, $w(\QGrid)\leq 1$ and $w(\QGridXY)\leq 2$.
      \item Likewise, $w(\QGrid)=1$ as there are problems that require more than one step.
      \item Let $P$ be a problem in \QGridXY in which the initial and target cells
        are neither in the same row nor same column. Any optimal trajectory for $P$
        contains some state $s$ where the agent is at a row and column different than
        those at the initial state. Such a state does not belong to $\OPT(T)$,
        where $T$ is the set of all atoms in $P$.
        By Theorem~\ref{thm:width:necessity}, $w(\QGridXY)>1$. Therefore, $w(\QGridXY)=2$.
    \end{enumerate}
  \end{boxed-example}
  \begin{boxed-example}[Width for Delivery problems (classes \QDeliveryS and \QDelivery)]
    \label{ex:width:delivery}
    \begin{enumerate}[$\bullet$]
      \item Problems in \QDeliveryS involve the agent and a single package.
        We showed before that $w(\QDeliveryS)=2$.
      \item The class \QDelivery has unbounded width, however. The intuition is that any admissible set
        of atom tuples must ``track'' the position of an arbitrary number of packages, those that have
        been already delivered. We make this intuition formal in the following.
      \item Let $P$ be a problem in \QDelivery with $k+2$ packages, let $\pi$ be an optimal policy for $P$,
        and let $T$ be the set of \textbf{all} atom tuples in $P$ of size at most $k$.
        Without loss of generality, we assume that the packages $p_1,p_2,\ldots,p_{k+2}$ must be delivered
        in such an order to guarantee optimality. %; i.e., any other order results in a suboptimal plan.
        %%We want to show that $\OPT(T)$ does not contain a closed $\pi$-envelope as then,
        %%by Theorem~\ref{thm:width:necessity}, $w(\QDelivery)>k$.
        %%So, let us suppose that $E$ is a closed $\pi$-envelope contained in $\OPT(T)$.
      \item Let $s$ be the state along an optimal trajectory $\tau$ for $P$ in which the packages
        $p_1,p_2,\ldots,p_{k+1}$ have been delivered, and the agent is at the target cell
        (just after delivering $p_{k+1}$).
        We now show that $s$ does not belong to $\OPT(T)$.
        Indeed, the tuples in $T$ can only track the position of at most $k$ objects in the set
        $\{p_1,p_2,\ldots,p_{k+1}\}$ of $k+1$ objects. Hence, no matter which tuple $t$ in $T$
        is considered, the state $s$ does not belong to $\OPT(t)$ as no state in $\OPT(t)$ has
        $k+1$ distinct packages at the target cell.
        Therefore, by Theorem~\ref{thm:width:necessity}, $w(P)>k$.
      \item Since \QDelivery contains problems with an arbitrary number of packages, $w(\QDelivery)=\infty$.
    \end{enumerate}
  \end{boxed-example}
}

\subsection{Algorithm \iwf{\Phi}}

The results above shed light on the power of the algorithms \iw{T}
and \iw{k} for problems that belong to classes $\Q$ for which there
is a general optimal policy $\pi$.
Indeed, if the set of optimal $T$-states, $OPT(T)$, forms  a closed $\pi$-envelope,
the algorithm \iw{T} is complete and optimal for $P$, and moreover,
reaches the goal of $P$ through $\pi$-trajectories, even if knowledge
of the policy or its features is not  known.

There is, however, a variant of \iw{T} that does not use tuples of atoms
or other particular details about the encoding for $P$, and uses instead
a set $\Phi$ of features.
The new algorithm, called \iwf{\Phi} and shown in Fig.~\ref{fig:iw(Phi)},
is like \iw{T} but works with feature valuations over $\Phi$ rather than
with atom tuples.
That is, \iwf{\Phi} does a breath-first search that prunes the states $s$
whose feature valuation $f(s)$ has been seen before during the search.

\begin{figure}
  \begin{tcolorbox}[title=\textbf{Algorithm~\ref{alg:iw(Phi)}: \iwf{\Phi} Search}] % for Linear Features}]
    \refstepcounter{myalgorithm}
    \label{alg:iw(Phi)}
    \begin{algorithmic}[1]\small
      \State \AlgINPUT Planning problem $P$ with $N$ atoms
      \State \AlgINPUT Set $\Phi$ of features, %with $k=|\Phi|$
                       and function $f$ to compute its valuation on states in $P$
      %\State \textbf{Input:} Function $f$ to compute feature valuations over $\Phi$
      \smallskip
      %\State Initialize a \textbf{perfect hash table} $T$ for storing feature valuations
      %  (i.e., $k$-tuples in $\{0,\ldots,N\}^k$) on which the operations of insertion and
      %  look up take constant time
      \State Initialize {hash table} $T$ for storing feature valuations
      \State Initialize FIFO queue $Q$ on which enqueue and dequeue operations take constant time
      \smallskip
      \State Enqueue node $n_0$ for the initial state $s_0$ of $P$
      \State While $Q$ is not empty:
      \State\qquad Dequeue node $n$ for state $s$
      \State\qquad If $s$ is a goal state, return the path to node $n$ \hfill\AlgCOMMENT{(Solution found)}
      \State\qquad If $f(s)$ is not in $T$:
      %\State\qquad\qquad If $s$ is a goal state, return the path to node $n$ \hfill\AlgCOMMENT{(Solution found)}
      \State\qquad\qquad Insert the feature valuation $f(s)$ in $T$
      \State\qquad\qquad Enqueue a node $n'$ for each successor $s'$ of $s$
      \smallskip
      \State Return \AlgFAILURE \hfill\AlgCOMMENT{(No suitable cost-envelope, cf.\ Theorem~\ref{thm:iw(Phi):optimal})}
    \end{algorithmic}
  \end{tcolorbox}
  \caption{An \iwf{\Phi} search is like an \iw{T} search but instead of tracking the
    tuples in $T$, it tracks feature valuations, and prune nodes whose valuation
    have been already seen. Guarantees for completeness and optimality are given
    in Theorem~\ref{thm:iw(Phi):optimal}.
  }
  \label{fig:iw(Phi)}
\end{figure}

% , on the other hand, it is no guaranteed, where completeness means
% that \iwf{\Phi} is guaranteed to find a solution when $P$ has a solution.
% The main obstacle to clear for proving completeness for \iwf{\Phi} is to show
% that the prunning based on feature valuations is safe (line 9).
% That is, that the queue $Q$ always contains a node for a state in a path to
% a goal state.

The question is the following.
Assuming that $\pi$ is an optimal rule-based policy that solves a class %of problems
$\Q$ that includes $P$, and that $\Phi$ is the set of features used by the policy:
does \iwf{\Phi} solve $P$ optimally?
%Moreover, does \iwf{\Phi} solve $P$ optimally if $\pi$ is an optimal policy for $\Q$?

It turns out that without extra conditions, the answer to this question is no.
One reason is that a policy $\pi$ may solve a problem by using a number of
feature valuations that is smaller (possibly, exponentially smaller) than the
number of states required to reach the goal.
In these cases, the \iwf{\Phi} search cannot get to the goal because any plan
must involve sequences where the same feature valuation repeats.
%will generate state sequences where same joint feature valuation will appear multiple times in the way to the goal.
For example, a policy for the Gripper domain where a number of balls have %$n$ have
to be carried from Room A to Room B, one by one, can be defined in terms of
 a set $\Phi$ of three Boolean features encoding whether the robot is in Room A, whether there are
balls still left in Room A, and whether a ball is being held by the robot.
The number of possible feature valuations is 8 but the length of the plans
grows  linearly with the number of balls.
%More generally, states in the path to goal may be pruned by \iwf{\Phi}
%by other, previously seen, states that do not lead to the goal.

We have the tools at our disposal however  to provide conditions that
ensure that the algorithm \iwf{\Phi} solves any problem $P \in \Q$ optimally,
if the policy $\pi$ does. %, and also conditions under which the optimality of the resulting  plan is guaranteed.
For a set of feature valuations $F$ over the features in $\Phi$, let $\OPT(F)$
stand for the set of min-cost states $s$ in $P$ with feature valuation $f(s)$  in $F$.
Sufficient conditions that ensure the completeness and optimality of \iwf{\Phi} can be expressed as follows:

% Formally, for a subset $F$ of (full) feature valuations for the features
% in $\Phi$, a state $s$ belongs to $\OPT(F)$ iff $s$ is a minimum-cost
% state that reaches the valuation $f(s)$.

\begin{theorem}[Completeness and optimality of \iwf{\Phi}]
  \label{thm:iw(Phi):optimal}
  Let $\Phi$ be a set of features for a planning problem $P$,
  and let $F$ be a set of feature valuations over $\Phi$.
  Then,
  \begin{enumerate}[1.]
    \item If $\OPT(F)$ is a cost-envelope, \iwf{\Phi} finds an optimal plan for $P$.
    %\item If $\pi$ is an optimal policy for a class $\Q$ that contains $P$, not
    %  necessarily defined over the features in $\Phi$, and $\OPT(F)$ is a closed
    %  $\pi$-envelope, \iwf{\Phi} finds an optimal plan for $P$.
    \item If $\pi$ is an optimal policy for a class $\Q$, and $\OPT(F)$ is a closed
      $\pi$-envelope for $P$ in $\Q$, \iwf{\Phi} finds an optimal plan for $P$.
  \end{enumerate}
  In either case, if the features in $\Phi$ are linear (cf.\ Definition~\ref{def:features:linear}),
  \iwf{\Phi} finds a plan of length $\O(N^{\ell})$ using $\O(bN^{\ell})$
  time, where $\ell$ is the number of \textbf{numerical features} in $\Phi$,
  $N$ is the number of atoms in $P$, and $b$ bounds the branching factor in $P$.
\end{theorem}
\Proof{%
  Essentially, the proof involves a similar but more complex invariant that the
  one used in the proof for the completeness of \iw{T} (cf.\ Theorem~\ref{thm:iw(T)}).
  We provide full details in what follows.
  The invariant that must be shown is:
  \emph{at the start of each iteration, the queue contains a node $n$ such
  that \nstate{n} is in $\OPT(F)$ and $\ncost{n}=\ocost(\nstate{n})$, where
  \nstate{n} denotes the state associated with the node $n$, \ncost{n} denotes
  the cost of $n$, and $\ocost(s)$ is the state function in Definition~\ref{def:pcost}.}
  We do an induction on the number of iterations:
  \begin{enumerate}[1.]
    \item The claim is true for the first iteration as $Q$ only contains
      the node $n_0$ for the initial state $s_0$ which is in $\OPT(F)$,
      and $\ncost{n_0}$ and $\ocost(s_0)$ are both equal to zero.
    \item Suppose that the claim holds at the start of iteration $i$.
      That is, $Q$ contains a node $n$ such that \nstate{n} is in $\OPT(F)$,
      and $\ncost{n}=\ocost(\nstate{n})$.
      We consider 4 cases:
      \begin{enumerate}[a)]
        \item The node $n$ is not dequeued. It then remains in $Q$ and the
          invariant holds for the next iteration.
        \item The node $n$ is dequeued and \nstate{n} is a goal state.
          Then, \iwf{\Phi} terminates with a goal-reaching path.
        \item The node $n$ is dequeued, \nstate{n} is not a goal state,
          and the node is not pruned (cf.\ line 9 in \iwf{\Phi}).
          Since \nstate{n} belongs to $\OPT(F)$, a node $n'$ is enqueued
          for a successor $s'$ of \nstate{n} such that $s'$ is in $\OPT(F)$.
          On the other hand, since $\OPT(F)$ is a cost-envelope, there is a
          goal-reaching $\pcost$-trajectory $s_0,\ldots,\nstate{n},s',\ldots$ which
          is an \textbf{optimal trajectory} for $P$ by Lemma~\ref{lemma:pcost}.
          Therefore,
          \begin{alignat*}{1}
            \ncost{n'}\ =\ 1 + \ncost{n}\ =\ 1 + \ocost(\nstate{n})\ =\ \ocost(s')\ =\ \ocost(\nstate{n'})
          \end{alignat*}
          where the third equality is by the principle of optimality.
          Hence, the invariant holds for the next iteration.
        \item The node $n$ is dequeued, \nstate{n} is not a goal state,
          and the node is pruned.
          Another node $n'$ with $f(\nstate{n'})=f(\nstate{n})$ was previously
          dequeued and expanded. Then,
          \begin{alignat*}{1}
            \ocost(\nstate{n'})\ \leq\ \ncost{n'}\ \leq\ \ncost{n}\ =\ \ocost(\nstate{n})\ \leq\ \ocost(\nstate{n'})
          \end{alignat*}
          where the second inequality is due to the nodes being ordered by costs in
          the queue, and the last inequality since $f(\nstate{n})=f(\nstate{n'})$ and
          $\nstate{n}\in\OPT(F)$.
          Therefore, equality holds throughout, $\ocost(\nstate{n'})=\ocost(\nstate{n})$,
          and thus $\nstate{n'}$ is in $\OPT(F)$.
          As in the previous case, at the iteration where the node $n'$ is dequeued,
          a node $n''$ for a successor state of $\nstate{n'}$ with $\nstate{n''}$ in
          $\OPT(F)$ and $\ncost{n''}=\ocost(\nstate{n''})$ is generated and enqueued.
          Thus, the invariant holds for the next iteration because $n''$
          is still in the queue since $\ncost{n}=\ncost{n'}<\ncost{n''}$.
      \end{enumerate}
  \end{enumerate}
  We now show that the path found by \iwf{\Phi} is optimal.
  Let $n^*$ be the last node dequeued by \iwf{\Phi}; i.e., \nstate{n^*} is a goal state.
  There are two complementary cases:
  \begin{enumerate}[$\bullet$]
    \item \nstate{n^*} is in $\OPT(F)$ and $\ncost{n^*}=\ocost(\nstate{n^*})$.
      Then, the path leading to $n^*$ is an optimal trajectory for $P$ by Lemma~\ref{lemma:pcost}.
    \item The negation of the first case.
      By the claim, at the time when $n^*$ is dequeued, the queue contains another node
      $n$ such that $\nstate{n}\in\OPT(F)$ and $\ncost{n}=\ocost(\nstate{n})$.
      On one hand, $\ncost{n}\leq C^*$ where $C^*$ is the cost of $P$.
      On the other hand, $\ncost{n^*}\leq\ncost{n}$ since $n^*$ is dequeued before $n$.
      Combining both inequalities, $\ncost{n^*}\leq C^*$ which means that the
      path found by \iwf{\Phi} of cost \ncost{n^*} is an optimal trajectory for $P$.
  \end{enumerate}

  \medskip\noindent
  The second claim is implied by the first since $\OPT(F)$ is a cost-envelope
  by Theorem~\ref{thm:cost-envelope:optimal-pi}.
  %\medskip\noindent
  Finally, for the complexity bounds, notice that the number of nodes expanded by
  \iwf{\Phi} is bounded by the number of different feature valuations,
  which is $\O(N^{\ell})$ if the features in $\Phi$ are linear and
  the number of numerical features in $\Phi$ is $\ell$.
  Thus, the plan length is $\O(N^\ell)$ while the number of generated
  nodes is $\O(bN^\ell)$, if $b$ bounds the branching factor.
  On the other hand, the operations on the hash table can be done
  in constant time on a perfect hash. %, while the operations on the queue also take constant time.
  Hence, \iwf{\Phi} runs in $\O(bN^\ell)$ time and space.
}

Notice that \iwf{\Phi} provides complexity bounds for the solvability of classes
of problems, \textbf{independently of the underlying structure of states.}
That is, \iwf{\Phi} does not know about atoms at all, only about feature valuations.
Two  encodings that are  different syntactically but equivalent semantically
will  yield the same behaviour in \iwf{\Phi}; e.g., two encodings of Blocksworld, 
one in which $clear$ is a primitive predicate and one where it is a derived predicate. 
%Thus, \iwf{\Phi} provides a further abstraction layer for classes of problems.

The set $\Phi$ of features in Theorem~\ref{thm:iw(Phi):optimal}.2 %and \ref{thm:iw(Phi)}
does not need to be the set of features on which the policy $\pi$ is defined.
If $F$ is the set of feature valuations reached by $\pi$, and $\reachable(\pi,P)$
is the set of states reachable by using $\pi$ in $P$, \iwf{\Phi} is guaranteed
to find an optimal plan when $\pi$ is an optimal policy, and $\OPT(F)=\reachable(\pi,P)$:

\begin{theorem}
  \label{thm:iw(Phi):simple}
  Let $\pi$ be a rule-based policy that \textbf{optimally solves} a problem $P$,
  and that is defined over the features in $\Phi$, and let $F$ be the set of
  feature valuations reached by $\pi$ in $P$.
  If $\OPT(F)=\reachable(\pi,P)$, then $\OPT(F)$ is a closed $\pi$-envelope,
  and thus \iwf{\Phi} finds an optimal plan for $P$.
\end{theorem}
\Proof{%
  By Theorem~\ref{thm:iw(Phi):optimal}, it is enough to show that $\OPT(F)$
  is a closed $\pi$-envelope.

  Clearly, the initial state $s_0$ belongs to $\OPT(F)$.
  If $s$ is a state in $\OPT(F)$, then 1)~there is a $\pi$-trajectory for $s$
  (as $s$ is $\pi$-reachable), 2)~there is a $\pi$-transition $(s,s')$ (as $\pi$
  solves $P$), and 3)~$s'$ belongs to $\OPT(F)$ (as $s'$ is reached by $\pi$).
  Closedness is direct by (1).
}

If different states have different feature valuations,
\iwf{\Phi} reduces to a breadth-first search, as only nodes for
duplicate states are pruned.
%In such a case, it is sufficient to pick the set $F$ of feature
%valuations as those resulting in a $\pi$-envelope.
The interesting uses of the theorem however are on settings where
the number of possible feature valuations is exponentially smaller
than the number of states.
We now present two examples involving the algorithm \iwf{\Phi}.

\Example{
  \vskip 1em
  \begin{boxed-example}[\iwf{\Phi} search on \QClear]
    \label{ex:iw(Phi):clear}
    \begin{enumerate}[$\bullet$]
      \item Let us consider the policy $\pi$ for \QClear previously defined
        in Example~\ref{ex:pi:clear} over the set of features $\Phi=\{H,n\}$.
        Fix a problem $P$ in \QClear, and let $F$ be the set of feature
        valuations reached by $\pi$ on $P$.
        The policy $\pi$ is optimal and $\OPT(F)=\reachable(\pi,P)$.
        By Theorem~\ref{thm:iw(Phi):simple}, \iwf{\Phi} finds an
        optimal plan for problem $P$.
        Notice that if $P$ has $N$ blocks, there are $2N$ different feature
        valuations, but an exponential number of configurations for the $N$
        blocks.
    \end{enumerate}
  \end{boxed-example}
  \begin{boxed-example}[The Marbles domain]
    \label{ex:pi:marbles}
    \begin{enumerate}[$\bullet$]
      \item The Marbles domain $M$ involves boxes and marbles.
        Boxes can be on the table and marbles inside boxes.
        Problems are specified with atoms $ontable(b)$ to tell
        that box $b$ is on the table,
        %$removed(b)$ to tell that box $b$ is not longer on the table,
        and $in(r,b)$ to tell that marble $r$ is in box $b$.
        The goal is to remove all boxes from the table, where a
        box can be removed only if it is empty.
        Marbles thus must be removed from boxes one at a time, in no specific order.
        The collection of all problems over the Marbles domain is denoted by \QMarbles,
        while $\QMarblesS\subseteq\QMarbles$ denotes the class of such problems
        with exactly one box.
      \item  Marbles is not a STRIPS domain as the goal must be specified
        with negative literals; i.e., $\bigwedge_b \neg ontable(b)$ where
        the conjunction is over all boxes $b$ in the problem.
        %First, each problem in \QMarbles has a plan, but the goal must
        %be specified with negative literals; i.e., $\bigwedge_b \neg ontable(b)$ where
        %the conjunction is over all boxes $b$.
\Omit{        Thus, no problem $P$ in \QMarbles has an admissible set $T$ of tuples as there
        are no such tuples to specify that a box has been removed, the goal has been reached, etc.
        Indeed, if $s_0$ is the initial state of a problem, any state $s$ that is
        reachable from $s_0$ is contained in $s_0$; i.e., the set of atoms made true by
        $s$ is a strict subset of the atoms made true by $s_0$.
        Hence, such a state $s$ is pruned by \iw{k}, for any value of $k$.
}
        %Therefore, $w(P)=\infty$ for any $P$ in \QMarbles, by definition.
      \item Yet there is a simple optimal policy $\pi$ for \QMarbles
        defined over the set $\Phi=\{n,m\}$ of numerical features where $n$ counts
        the number of boxes still on the table, and $m$ counts the number of marbles
        in the ``first box'' among those still on the table, where a static ordering
        of the boxes is assumed.
        A general optimal policy $\pi$ over $\Phi$ can be  expressed as follows:
        \begin{alignat*}{1}
          &\prule{\GT{m}}{\DEC{m}} \,, \\
          &\prule{\EQ{m}, \GT{n}}{\DEC{n}, \UNK{m}} \,.
        \end{alignat*}
        The first rule says to remove a marble from the first box on the table
        when such a box is not empty, while the second rule says to take an action
        that decreases the number of boxes on the table.
        % , yet observe that such a
        % transition does not necessarily removes the first box from the table as
        % it can remove any other empty box.
        % The role of the condition $\EQ{m}$ is to guarantee that a transition
        % compatible with the rule exists.
        The policy $\pi$ solves any problem in \QMarbles or \QMarblesS optimally,
        as any optimal plan must execute a number of actions that is equal to the total
        number of marbles in all boxes plus the number of boxes.
      \item For analyzing  the algorithm \iwf{\Phi} over instances in \QMarbles, 
        let $P$ be a problem in \QMarbles, and let $F$ be the set of feature valuations
        reached by $\pi$ on $P$.    The policy $\pi$ is optimal for \QMarbles and $\OPT(F)=\reachable(\pi,P)$.
        By Theorem~\ref{thm:iw(Phi):simple}, \iwf{\Phi} finds an optimal plan for $P$
        in $\O(N^3)$ time, where $N$ is the number of atoms in $P$, since the features
        in $\Phi$ are linear, and the branching factor in $P$ is bounded by $N$.
    \end{enumerate}
  \end{boxed-example}
}

\section{Serializations}

% We define the weakest possible notion of acyclicity that is required.
% This notion makes the statement/proofs of subsequent proofs simpler.

The problem of subgoal structure is critical in classical planning,
hierarchical planning, and reinforcement learning although in most cases
the problem has not been addressed formally. We draw on the language for
general policies to express decompositions into subproblems, and on the
notion of width for expressing and evaluating such decompositions and the
subgoal structures that result.

We start with the notion of \emph{serializations} which are defined
semantically as binary relations on states that are acyclic.
%%\textbf{Alive state} below refers to a reachable state $s$ in $P$
%%that is not a dead-end or a goal state:

\begin{definition}[Serializations]
  \label{def:serialization}
  %Let $\Q$ be a collection of problems, and let $r_\prec$ be a binary relation
  %on $\cup_{P\in\Q}\states(P)$.
  A \textbf{serialization} over a collection of problems $\Q$
  is a binary relation `$\prec$' over the states in $\cup_{P\in\Q}\states(P)$
  that is \emph{acyclic in $\Q$,} meaning that there is no set $\{s_0,s_1,\ldots,s_n\}$
  of \textbf{reachable states} in $P$, where $s_0$ is the initial state, such that
  $s_{i+1}\prec s_{i}$ for $i=0,\ldots,n-1$, and $s_j\prec s_n$ for some $0\leq j\leq n$.
  %\alert{** Change resolves Hanoi and makes sense **}
\end{definition}

%Recall that a binary relation $r$ is acyclic in $\Q$ if
%there is no sequence of states $s_0,s_1,\ldots,s_n$ in a problem $P$ in $\Q$
%such that each one of the \emph{state pairs} $s_i,s_{i+1}$ is in $r$
%and $s_n=s_k$ for $0 \leq k < n$.

For binary relations `$\prec$' that express serializations, the notation
$s'\prec s$ expresses in infix form that the state pair $(s,s')$ is in `$\prec$'.
There is no assumption that the state pair $(s,s')$ is a state transition;
namely, that $s'$ is at one step from $s$, and the meaning of $s'\prec s$
is that $s'$ is a possible subgoal state from state $s$.

A serialization `$\prec$' over $\Q$ splits a problem $P$ in $\Q$ into
\emph{subproblems.}
For a reachable state $s$ in $P$, the subproblem $P_\prec[s]$ is like $P$ but
with two changes: the initial state is $s$, and the goal states
are the states $s'$ such that $s'$ is a goal state of $P$, or $s' \prec s$.
% Then,

\begin{definition}[Subproblems]
  \label{def:subproblems}
  Let $\prec$ be a serialization over a class $\Q$, and let $P$ be a problem
  in $\Q$. The class of subproblems induced by $\prec$ on $P$, denoted by
  $P_\prec$, is the \emph{smallest class} that satisfies:
  \begin{enumerate}[1.]
    \item $P_\prec[s_0]$ is in $P_\prec$ for the initial state $s_0$ of $P$, and
    \item $P_\prec[s']$ is in $P_\prec$ if % there is %some state $s$ such that
      $P_\prec[s]$ in $P_\prec$, $s'\prec s$, $s'$ is not a goal state, and either
      \begin{enumerate}[a)]
        \item $s'$ is a successor state of $s$, or % that is not a goal state, or
        \item $s'$ is a reachable state from $s$, and there is no
          \emph{successor} $s''$ of $s$ that is either a goal state of $P$ or $s''\prec s$.
      \end{enumerate}
      We say that the subproblem $P_\prec[s]$ induces the subproblem $P_\prec[s']$.
      %and also that the latter subproblem is induced by the former.
  \end{enumerate}
\end{definition}

Cases $2a$ and $2b$ ensure that a non-successor state of $s$ is regarded as a possible subgoal from $s$
only when  no  successor state of $s$  is subgoal from $s$. In other words, while the possible subgoal states from $s$ do not have to be at a
minimum distance from $s$, they have to be at distance 1 when there is a subgoal state that is at such a distance.
%%The difference between conditions 2a and 2b is that when $s$ has successors
%%$s'$ that are either a goal state or $s'\prec s$, there are no subproblems
%%induced by states that are reachable from $s$ in two or more steps.
%%Observe that if the subproblem $P_\prec[s]$ belongs to $P_\prec$, there
%%is a state sequence $s_0,s_1,\ldots,s_n$ such that $s_n=s$, and
%%the $P_\prec[s_i]$ induces $P_\prec[s_{i+1}]$, $0\leq i<n$.

Intuitively, a serialization is ``good'' if it results in subproblems that
have small, bounded width that can be solved greedily in the way to the goal.

\begin{definition}[Serialized width]
  \label{def:serialization:width}
  Let `$\prec$' be a serialization over a class of problems $\Q$, and let $P \in \Q$.
  Then,
  \begin{enumerate}[1.]
    \item The \textbf{(serialized) width} of $P$, denoted as $w_\prec(P)$, is
      the minimum non-negative integer $k$ that bounds the width $w(P_\prec[s])$
      of all the subproblems $P_\prec[s]$ in $P_\prec$. %for the alive states $s$ in $P$.
    \item The \textbf{(serialized) width} of $\Q$, denoted as $w_\prec(\Q)$,
      is the minimum non-negative integer $k$ that bounds the serialized width
      $w_\prec(P)$ of the problems $P$ in $\Q$.
  \end{enumerate}
\end{definition}

Starting from a problem $P$ in $\Q$, a serialization may lead to state $s$
if the subproblem $P_\prec[s]$ belongs to the subclass $P_\prec$ of subproblems
induced by $\prec$. Hence, if for a dead-end state $s$, the subproblem
$P_\prec[s]$ belongs to $P_\prec$,  by definition, $w(P_\prec[s])=\infty$
and thus $w(\Q)=\infty$ as well.

\Omit{
    For serializations of bounded width to result in greedy, and eventually,
    polynomial algorithms, serializations '$\prec$' have to be \textbf{safe}
    in the following way: if $s$ is an alive state in $P$, and $s'\prec s$
    for another state $s'$ in $P$, then $s'$ is not a dead-end state;
    namely, dead-end states should not be possible subgoals for alive states.

    \Omit{
      Serializations and policies have been defined as binary relations
      on state pairs $(s,s')$. One difference is that serializations must be acyclic;
      the other difference is that policies only consider state transitions $(s,s')$;
      namely, state pairs separated by a single action. No such restriction is made
      for serializations that may include a pair $(s,s')$ where $s'$ is many steps
      away from $s$.
    }
}

Interestingly, serializations of zero width are policies, and vice versa,
policies are serializations of zero width.
%%However, the latter correspondence requires the assumption that either
%%the class of problems $\Q$ is \textbf{closed,} or that the binary relation
%%that defines the policy is acyclic in $\Q$.
%%A class $\Q$ is closed if $\Q$ contains the problem $P[s]$, that is like
%%$P$ but with initial state $s$, when it contains the problem $P$ and $s$
%%is an alive state in $P$:
%%%%in the following sense: if  $P$ is in $\Q$, the planning problem $P[s]$
%%%%that is like $P$ but with initial state $s$, is also in $\Q$,
%%%%for any alive state $s$ in $P$:

\Omit{
  The difference between serialization and policies vanishes for serializations of
  \emph{zero width}; i.e., serializations that result into subproblems that can be
  solved in a single step.
  More specifically, a zero-width serialization for a class $\Q$ is indeed a policy
  that solves $\Q$, but the converse is not necessarily true unless $\Q$ is closed,
  as defined above.
}

%%A class of problems $\Q$ is \textbf{closed} iff for any problem $P$ in $\Q$ and any
%%alive state $s$ in $P$, the planning problem $P[s]$ (that is like $P$ but with initial
%%state $s$) also belongs to $\Q$.

\begin{theorem}[Zero-width serializations and policies]
  \label{thm:serialization:policies}
  Let $\Q$ be a class of problems, and let $\prec$ be a binary relation
  on $\cup_{P\in\Q}\states(P)$.
  Then, $\prec$ is a serialization of zero width for $\Q$ iff $\prec$ is a policy that solves $\Q$.
  %\begin{enumerate}[1.]
  %  \item If $\prec$ is a serialization of zero width for $\Q$, $\prec$ is a policy that solves $\Q$.
  %  \item If $\Q$ is closed and $\prec$ is a policy that solves $\Q$, $\prec$ is a serialization of zero width for $\Q$.
  %  \item If $\prec$ is a policy that solves $\Q$ and $\prec$ is acyclic in $\Q$, $\prec$ is a serialization of zero width for $\Q$.
  %\end{enumerate}
\end{theorem}
\Proof{%
  In this proof, we write $s\prec s'$ to denote that the pair $(s,s')$ is
  in the relation $\prec$, either when $\prec$ denotes a serialization or
  a policy.

  \medskip\noindent
  \textbf{($\Rightarrow$)}
  Assume that $\prec$ is a serialization of zero width for $\Q$.
  Clearly, $\prec$ is a policy as it is a binary relation on $\cup_{P\in\Q}\states(P)$.
  It remains to show that for any problem $P$ in $\Q$, every maximal $\prec$-trajectory
  seeded at the initial state $s_0$ of $P$ is goal reaching.

  Let $\tau=s_0,s_1,\ldots,s_n,\ldots$ be one such trajectory; i.e., $s_i\prec s_{i+1}$ for $i\geq0$.
  Since $\states(P)$ is a finite set and $\prec$ is acyclic in $P$,
  $\tau$ must be of finite length. Let us assume that it ends at $s_n$.
  If $s_n$ is not a goal state, the subproblem $P_\prec[s_n]$ belongs
  to $P_\prec$ by Definition~\ref{def:subproblems}.
  Then, $w(P_\prec[s_n])=0$ since $\prec$ is of zero width.
  This means that there is a successor state $s'$ of $s_n$ such that $s_n\prec s'$.
  Hence, $\tau$ is not a maximal trajectory contradicting the assumption.
  Therefore, $s_n$ must be a goal state.

  \medskip\noindent
  \textbf{($\Leftarrow$)}
  Assume that $\prec$ is a policy that solves $\Q$.
  We first show that $\prec$ is acyclic in $\Q$, and then that its width is zero.

  Let $P$ be a problem in $\Q$, and let $s_0,s_1,\ldots,s_n$ be a set of
  reachable states in $P$ such that $s_0\prec s_1\prec\cdots\prec s_n$.
  We need to show that there is no index $0\leq j\leq n$ such that $s_n\prec s_j$.
  Let us suppose that there is such an index $j$.
  Then, the trajectory $\tau=s_0,s_1,\ldots,s_j,\ldots,s_n,s_j$
  would be a maximal $\prec$-trajectory in $P$.
  As $\tau$ is not goal reaching, $\prec$ does not solve $\Q$ as assumed.
  Therefore, $s_n\prec s_j$ does not hold.

  \smallskip\noindent
%   \textbf{Claim.}
  If $P_\prec[s]$ is a subproblem in the class $P_\prec$, the state $s$ is reachable
  through a $\prec$-trajectory from the initial state $s_0$ of $P$; i.e., there is
  a $\prec$-trajectory $s_0,s_1,\ldots,s_n$ with $s_n=s$.
  Indeed, if $P_\prec[s]$ is in $P_\prec$, there is a \emph{state sequence}
  $s_0,s_1,\ldots,s_n$ such that $s_n=s$, and the subproblem $P_\prec[s_i]$ induces
  $P_\prec[s_{i+1}]$ for $0\leq i<n$.
  However, $\prec$ is only defined on state transitions as it is a policy.
  Therefore, such a state sequence is a state trajectory.

  \smallskip
  Let us compute the width of a subproblem $P_\prec[s]$ in $P_\prec$.
  By the claim, there is a $\prec$-trajectory $s_0,s_1,\ldots,s_n$ such that $s_n=s$,
  and by definition of subproblem, $s_n$ is not a goal state.
  Then, since $\prec$ is a policy that solves $\Q$, there is a state
  transition $(s,s')$ such that $s\prec s'$. Hence, $w(P_\prec[s])=0$,
  which implies $w_\prec(P)=0$ and $w_\prec(\Q)=0$.
}

\Example{
  \begin{boxed-example}[Zero-width serializations for different classes]
    \label{ex:serializations:1}
    \begin{enumerate}[$\bullet$]
      \item
        The rule-based policies given for the classes \QClear, \QGrid, \QMarbles,
        and \QDelivery represent serializations of zero width, as determined by 
        Theorem~\ref{thm:serialization:policies}.
    \end{enumerate}
  \end{boxed-example}
}

% As expected, if all subproblems have solution, every problem in $\Q$ has solution,
% and a solution can be constructed greedily, backtrack-free, by solving subproblems
% in a sequential manner.

An important property of serializations of bounded width
is that they permit the decomposition of a problem into subproblems
which can be solved greedily in polynomial time without the need to backtrack.
% Moreover, each of the subproblems can be solved in polynomial time.
This  property is  exploited by the algorithm \siwp shown in Fig.~\ref{fig:siwp}.

\begin{figure}
  \begin{tcolorbox}[title=\textbf{Algorithm~\ref{alg:siwp}: \siwp Search}]
    \refstepcounter{myalgorithm}
    \label{alg:siwp}
    \begin{algorithmic}[1]\small
      \State \AlgINPUT Testable serialization `$\prec$' for class $\Q$
      \State \AlgINPUT Planning problem $P$ in class $\Q$
      \smallskip
      \State Initialize state $s$ to initial state $s_0$ for $P$
      \State While $s$ is not a goal state of $P$:
      \State\qquad Do an IW search from $s$ to find $s'$ that is either a goal state or $s'\prec s$
      \Statex\qquad\quad (i.e., the goal test in line 9 of \iw{k} is augmented with $s'\prec s$ where
      \Statex\qquad\quad $s'$ is the state for the dequeued node $n$ in line 7)
      \State\qquad If $s'$ is found, set $s\gets s'$. Else, return \AlgFAILURE \hfill\AlgCOMMENT{(Serialized width of $P$ is $\infty$)}
      \smallskip
      \State Return the path from $s_0$ to the goal state $s$ \hfill\AlgCOMMENT{(Solution found)\,\,}
    \end{algorithmic}
  \end{tcolorbox}
  \caption{\siwp solves a problem $P$ by using the serialization to decompose $P$
    into subproblems, each one that is solved with an IW search.
    Testable serialization means that there is an algorithm for testing $s'\prec s$ for any pair of states.
    The completeness of \siwp is given in Theorem~\ref{thm:siwp}.
  }
  \label{fig:siwp}
\end{figure}

\begin{theorem}[Completeness of \siwp]
  \label{thm:siwp}
  Let `$\prec$' be a serialization for a class $\Q$ of problems.
  If $w_\prec(\Q)\leq k$, then any problem $P$ in $\Q$ can be solved by \siwp.
  Moreover, the IW search in line~5 of \siwp takes $\O(bN^{2k-1})$ time and
  $\O(bN^k)$ space, where $N$ is the number of atoms, and $b$ bounds the
  branching factor in $P$.
\end{theorem}
\Proof{%
  Let $s_0$ be the initial state of a problem $P$ in $\Q$, and let
  $\tau=s_0,s_1,\ldots,s_n$ be a \textbf{state sequence} in $P$ where
  all states, except perhaps $s_n$, are non-goal states.
  We say that $\tau$ is a $\prec$-sequence iff for each index $0\leq i<n$,
  the state $s_{i+1}$ is reachable from the state $s_i$ and $s_{i+1}\prec s_i$,
  but if the state $s_{i+1}$ is not a successor of $s_i$, then $s_i$ has no
  successor $s'$ such that $s'\prec s_i$.
  Observe that if $\tau$ is a $\prec$-sequence, then a simple inductive argument
  shows that the subproblem $P_\prec[s_i]$ induces the subproblem $P_\prec[s_{i+1}]$,
  $0\leq i<n$.
  If $s_n$ is a goal state, the sequence is called a $\prec$-solution.

  \smallskip\noindent
%   \textbf{Claim.}
  If $\tau=s_0,s_1,\ldots,s_n$ is a $\prec$-sequence for $P$, there is a $\prec$-solution
  $\tau'$ for $P$ that extends $\tau$.
  Indeed, if $s_n$ is a goal state, $\tau$ is already a $\prec$-solution.
  Otherwise, the subproblem $P_\prec[s_n]$ belongs to $P_\prec$.
  Since $w_\prec(P)\leq k$, there is a state $s_{n+1}$ reachable from $s_n$
  that is either a goal state, or $s_{n+1}\prec s_n$; i.e., the sequence
  $\tau,s_{n+1}$ is a $\prec$-sequence.
  Iterate until finding an extension $\tau'$ of $\tau$ that ends in a goal state,
  which can be done because $\prec$ is acyclic, and the number of states in $P$ is finite.

  \smallskip
  % By the claim,
  It is easy to see that at the start of each iteration
  of the loop, \siwp has discovered a $\prec$-sequence $\tau=s_0,s_1,\ldots,s_n$
  that ends at the current state $s_n=s$. Hence, since $\prec$ is acyclic,
  the loops eventually ends.
  The time and space bounds for each call of IW in line~5 of \siwp follow
  directly from Theorem~\ref{thm:iw(T)}.
}

However, even if the subproblems are solved greedily and in polynomial time by IW,
the total number of calls to IW, and hence the total running time of \siwp, cannot be bounded
without extra assumptions on the structure of the serialization. Indeed, there are serializations
that split a problem into an exponential number of subproblems, like the Hanoi example below.
However, once we move to \emph{serializations} expressed by means of rules akin to those used to express
policies, we will be able to provide conditions and bound the running time of \siwp.

\Example{
  \medskip
  \begin{boxed-example}[The Hanoi domain]
    \label{ex:hanoi}
    \begin{enumerate}[$\bullet$]
      \item $\QHanoi$ is the class of Towers of Hanoi problems involving 3 pegs,
        numbered from 0 to 2, and any number of disks, where the initial and goal
        states correspond to single towers at different pegs, respectively.
        Recently, \citeay{hanoi:aaai2023} refer to a general strategy that solves
        problems of moving a single tower from peg 0 to peg 2:
        \begin{quote}
          \it Alternate actions between the smallest disk and a non-smallest disk.
          When moving the smallest disk, always move it to the left.
          If the smallest disk is on the first pillar, move it to the third one.
          When moving a non-smallest disk, take the only valid action.
        \end{quote}
      \item This strategy can be expressed as a rule-based policy
        using three Boolean features $p_{i,j}$, $1 \leq i < j \leq 3$,
        that are true if the top disk at peg $i$ is smaller than the top disk at peg $j$.
        % Each of the resulting subproblems can be solved in a single step, but there is
        % indeed, an exponential number of subproblems to solve.
        However, in order to account for the alternation of movements, it must be
        assumed that the planning encoding adds an extra atom $e$ that is true
        initially, and that flips with each movement. Provided then with a Boolean
        feature $q$ that tracks the value of $e$, a general policy for Hanoi can
        be expressed with the following set $R$ of rules over the features  $\Phi=\{q,p_{1,2},p_{1,3},p_{2,3}\}$:
        %with the following set $R$ of rules:
        \begin{alignat*}{2}
          &\text{\% Movements of the smallest disk}\notag \\
          &\prule{q, p_{1,2},p_{1,3}}{\neg q, \UNK{p_{1,2}}, \neg p_{1,3}, \neg p_{2,3}}       &\qquad &\text{(Move the smallest from peg 1 to peg 3)} \\
          &\prule{q, \neg p_{1,2},p_{2,3}}{\neg q, p_{1,2}, p_{1,3}, \UNK{p_{2,3}}}            &\qquad &\text{(Move the smallest from peg 2 to peg 1)} \\
          &\prule{q, \neg p_{1,3}, \neg p_{2,3}}{\neg q, \neg p_{1,2}, \UNK{p_{1,3}}, p_{2,3}} &\qquad &\text{(Move the smallest from peg 3 to peg 2)} \\[1em]
          &\text{\% Movements of the other disk}\notag \\
          &\prule{\neg q, p_{1,2},p_{1,3},p_{2,3}}{q, \neg p_{2,3}}                            &\qquad &\text{(Move the other from peg 2 to peg 3)} \\
          &\prule{\neg q, p_{1,2},p_{1,3},\neg p_{2,3}}{q, p_{2,3}}                            &\qquad &\text{(Move the other from peg 3 to peg 2)} \\
          &\prule{\neg q, \neg p_{1,2},p_{1,3},p_{2,3}}{q, \neg p_{1,3}}                       &\qquad &\text{(Move the other from peg 1 to peg 3)} \\
          &\prule{\neg q, \neg p_{1,2},\neg p_{1,3},p_{2,3}}{q, p_{1,3}}                       &\qquad &\text{(Move the other from peg 3 to peg 1)} \\
          &\prule{\neg q, p_{1,2},\neg p_{1,3},\neg p_{2,3}}{q, \neg p_{1,2}}                  &\qquad &\text{(Move the other from peg 1 to peg 2)} \\
          &\prule{\neg q, \neg p_{1,2},\neg p_{1,3},\neg p_{2,3}}{q, p_{1,2}}                  &\qquad &\text{(Move the other from peg 2 to peg 1)}
        \end{alignat*}
      \item In problems with an odd, respectively even, number of disks, the policy moves
        a single tower at peg $i$ to peg $j$, where $j=(i-1\!\mod 3)$, resp.\ $j=(i+1\!\mod 3)$.
        By moving the smallest disk to the right rather than to the left,
        the target peg changes to $j=(i+1\!\mod 3)$, resp.\ $j=(i-1\!\mod 3)$.
      \item The policy $\pi_R$ defined by rules in $R$ is thus general for the class
        \QHanoiOdd that contains the problems with an odd number of disks, initial
        situation with a single tower at peg 0, and goal situation with a single
        tower at peg 2. A general policy for \QHanoi can be obtained by considering additional
        Boolean features that tell the parity of the number of disks, and the pegs
        for the initial and final towers. % , and a slightly different set of rules).
      \item The policy $\pi_R$ defines a serialization of zero width by
        Theorem~\ref{thm:serialization:policies}.
        This serialization splits a problem $P$ in \QHanoiOdd into an exponential
        number of subproblems as $2^n-1$ steps are needed to solve a Hanoi problem
        with $n$ disks.  The algorithm \siwR solves any problem $P$ in \QHanoiOdd,
        but not in polynomial time.
        \Omit{
          \alert{** Policy transfers tower in peg 0 to peg 2, for odd number of disks, and to peg 1 for even number of disks **}
          \alert{CHECK} The resulting rule-based policy solves $\QHanoi$ and it also defines a serialization of width $0$ that splits
          the Hanoi problems into an exponential number of subproblems that can all be solved in one step.
          \Omit{
            On the other hand, it is not difficult to endow policies with an
            (internal) state, among a finite and fixed set of such states, so
            that rules are attached to such states and may change it.
            In such a case, the alternation is easily achieved with two internal
            states.  \alert{(** Termination test, i.e.\ \sieve, is not affected by such states; i.e., a slight modification of \sieve still works **)}
          }
          \alert{$\QHanoi$ is not closed and not sure if $\pi$ solves the closure of $\QHanoi$. In a sense, the policy assumes
            that only certain states will be reached in the way to the goal. PROBLEM is that *subproblems* for serialization defined
            in terms of alive states, not those reachable by policy. NOT SURE how to fix this right away. Of course, has simple solution,
            but not direct. Need to think a bit about this to be able to talk about this policy as a serialization. In a sense,
            if we can talk about *closest* states for defining subproblems, but then we cannot use \siwR algorithm but if width is $k$,
            \siwR($k$). ** Think **}
        }
    \end{enumerate}
  \end{boxed-example}
}

\subsection{Rule-Based Serializations: Sketches}

As with policies, the binary relations that encode serializations
can be compactly represented by means of rules.
The syntax of the rules is exactly the syntax of policy rules, and
the only difference is in the semantics of the rules where state
pairs $(s,s')$ are not limited to state transitions:

\begin{definition}[Sketches]
  \label{def:serialization:rules}
  \label{def:sketch}
  Let $\Q$ be a collection of problems, let $\Phi$ be a set of features
  for $\Q$, and let $R$ be a set of rules over $\Phi$.
  The rules in $R$ define the \textbf{binary relation} $\prec_R$ over the
  states in $\cup_{P\in\Q}\states(P)$ given by $s'\prec_R s$ iff the
  state pair $(s,s')$ is compatible with some rule in $R$.
  If $\prec_R$ is \textbf{acyclic in $\Q$,} % (cf.\ Definition~\ref{def:serialization}),
  $\prec_R$ is a serialization over $\Q$.
  % $(\Q,\prec_R)$ is a serialization.
  \end{definition}

The rules that define serializations are called \textbf{sketch rules,} and
sets of such rules are called \textbf{sketches}. The sketch width of $\Q$
given a sketch $R$ is the serialized width of $\Q$ under the serialization
$\prec_R$ defined by $R$.

\begin{definition}[Sketch width]
  \label{def:sketch:width}
  Let $R$ be a set of rules that define a serialization $\prec_R$ over a class
  $\Q$ of problems. The \emph{sketch width} of $R$ over $\Q$, denoted by $w_R(\Q)$,
  is $w_R(\Q)\doteq w_{\prec_R}(\Q)$.
\end{definition}

\Omit{
  Finally, a sketch is said to be \textbf{safe} if it results in a serialization
  that is safe; i.e., if there is no state pair $(s,s')$ over an instance $P$ in
  $\Q$ that satisfies a sketch rule with $s$ being an alive state, and $s'$ being
  a dead-end state reachable from $s$.
}

\noindent
A rule-based policy $\pi$ that solves $\Q$ is a sketch $R$ for $\Q$ of zero width:

\begin{theorem}[Rule-based policies and sketches]
  \label{thm:sketch:policies}
  Let $R$ be a set of rules defined in terms of a set of features $\Phi$ for
  a class $\Q$ of problems.
  Then, $R$ is a rule-based policy that solves $\Q$ iff $R$ is a sketch
  of zero width for $\Q$.
\end{theorem}
\Proof{%
  Straightforward using Theorem~\ref{thm:serialization:policies}.
}

\subsection{Algorithms}

If $R$ is a sketch of bounded width over a class $\Q$, the
problems in $\Q$ can be solved by the \siwR algorithm, shown
in Fig.~\ref{fig:siwR}, where $s' \prec_R s$ is tested by checking
if some rule in $R$ is compatible with the state pair $(s,s')$.
However, to bound the complexity of \siwR, a bound in the total
number of subproblems that need to be solved is needed.
A simple way to bound such a number is to require that the subgoal
states $s_0, \ldots, s_n$ in state sequences compatible with a
sketch $R$ have different feature valuations:
%%A simple way to bound such a number is to require that different
%%subproblems $P_\prec[s]$ and $P_\prec[s']$ correspond to different
%%feature valuations $f(s)$ and $f(s')$:

\begin{figure}
  \begin{tcolorbox}[title=\textbf{Algorithm~\ref{alg:siwR}: \siwR Search}]
    \refstepcounter{myalgorithm}
    \label{alg:siwR}
    \begin{algorithmic}[1]\small
      \State \AlgINPUT Sketch $R$ that defines relation $\prec_R$
      \State \AlgINPUT Planning problem $P$ in collection $\Q$
      \smallskip
      \State Initialize state $s$ to initial state $s_0$ for $P$
      \State While $s$ is not a goal state of $P$:
      \State\qquad Do an IW search from $s$ to find $s'$ that is either a goal state or $f(s')\prec_R f(s)$
      \Statex\qquad\quad (i.e., the goal test in line 9 of \iw{k} is augmented with $f(s)\prec_R f(s')$ where
      \Statex\qquad\quad $s'$ is the state for the dequeued node $n$ in line 7)
      \State\qquad If $s'$ is found, set $s\gets s'$. Else, return FAILURE \hfill\AlgCOMMENT{(Serialized width of $P$ is $\infty$)}
      \smallskip
      \State Return the path from $s_0$ to the goal state $s$ \hfill\AlgCOMMENT{(Solution found)\,\,}
    \end{algorithmic}
  \end{tcolorbox}
  \caption{\siwR is \siwp with the serialization $\prec_R$ induced by the sketch $R$, which is testable.
    The completeness and complexity of \siwR is given in Theorem~\ref{thm:siwR}.
  }
  \label{fig:siwR}
\end{figure}

\begin{definition}[Feature-acyclic sketches]
  \label{def:sketch:f-acyclic}
  Let $\Q$ be a class of problems, and let $R$ be a set of rules for $\Q$
  defined on a set $\Phi$ of features that define a binary relation $\prec_R$.
  The relation $\prec_R$, or simply  $R$, is said to \textbf{feature-acyclic}
  over $\Q$ if it is so for each problem $P$ in $\Q$, where the latter means
  that there is no set $\{s_1,s_2,\ldots,s_n\}$ of reachable states in $P$ such
  that $s_1\prec_R s_2\prec_R\cdots\prec_R s_n$, and $f(s_i)=f(s_j)$ for some
  $1\leq i<j\leq n$.
\end{definition}

\noindent Clearly, if $R$ is feature-acyclic over $\Q$, then $\prec_R$ is (state) acyclic
over $\Q$, and hence $\prec_R$ is a serialization, and $R$ is a sketch.
%a serialization (cf.\ Definition~\ref{def:serialization}), and $R$ is a sketch
%(cf.\ Definition~\ref{def:sketch}).
The complexity bound for algorithm \siwR follows:

\begin{theorem}[Completeness of \siwR]
  \label{thm:siwR}
  Let $R$ be a \textbf{feature-acyclic} sketch for a class $\Q$ of problems
  of width bounded by $k$.
  \siwR solves any problem $P$ in $\Q$ in polynomial time (exponential only $k$,
  not in the size of $P$).
  In particular, if the features are linear, $P$ is solved by \siwR
  in $\O(N^{\ell}(N^{k+1}+bN^{2k-1}))$ time and $\O(bN^k)$ space, producing
  a plan of length $\O(N^{\ell+k})$, where $N$ is the number of atoms in $P$,
  $b$ bounds the branching factor in $P$, and $\ell$ is the number of
  \textbf{numerical features} in $\Phi$.
  \Omit{
    In the general case, if $(\Q,R)$ has width bounded by $k$, \siwR solves any
    problem $P$ in $\Q$ in $\O(D^{|\Phi|}(N^k\Lambda+bN^{2k-1}))$ time and
    $\O(bN^k)$ space, producing a plan of length $\O(D^{|\Phi|}N^k)$, where
    $D$ bounds the domain size for the features in $\Phi$, and $\Lambda$ bounds
    the time to compute the feature valuations $f(s)$ for the states $s$ in $P$.
  }
\end{theorem}
\Proof{%
  The \siwR algorithm is the \siwp algorithm that uses the serialization
  `$\prec_R$' induced by the sketch $R$. Hence, by Theorem~\ref{thm:siwp},
  \siwR solves any problem $P$ in $\Q$, and each call to IW in line~5 of
  \siwR takes $\O(bN^{2k-1})$ time and $\O(bN^k)$ space.

  As $\prec_R$ is feature-acyclic, 
  the number of subproblems to solve is bounded by the maximum number of
  feature valuations that can appear when solving $P$.
  In the case of linear features, this number is $\O(N^{\ell})$.
  For each expanded state in each call to IW, the value of the features
  are computed in $\O(|\Phi|N)=\O(N)$ time.
  Thus, the total running time of \siwR is $\O(N^{\ell}(N^{k+1}+bN^{2k-1}))$.

  For the space required by \siwR, since the solutions to the subproblems
  produced by   IW do not need to be stored, the space complexity of \siwR is
  is the space complexity of the IW calls; namely, $\O(bN^k)$.
  The length of the overall plan, however, is bounded by the
  number of subproblems times their maximum possible lengths as 
  $\O(N^{\ell+k})$.
}

\Example{
  \begin{boxed-example}[Sketches for the Delivery domain]
    \label{ex:sketch}
    Table~\ref{table:sketches:delivery} contains different sets of rules over the
    set $\Phi=\{H,p,t,u\}$ of features for the Delivery domain.
    For each such set, the table indicates whether the set is feature-acyclic,
    and contains the sketch width for the classes \QDeliveryS and \QDelivery.
    The width is only specified for sets that are acyclic.
    We briefly explain the entries in the table without providing formal proofs,
    but all the details can be easily filled in with the results in the paper.
    \begin{enumerate}[$\bullet$]
      %\item Table~\ref{table:sketches:delivery} contains different sets of rules over the
      %  set $\Phi=\{H,p,t,u\}$ of features for Delivery; cf.\ Example~\ref{ex:pi:delivery}.
      %  For each such set, the table provides indicates whether the set is (structurally)
      %  terminating, and the sketch width for \QDeliveryS and \QDelivery.
      %  The width is only specified for sets that are terminating as rendered by \sieve.
      %\item We briefly explain the entries in the table without providing formal proofs,
      %  but all the details can be easily filled with the results in the paper.
      \item $R_0$ is the empty sketch whose width is the same as the plain width.
        %2 for \QDeliveryS and unbounded for \QDelivery.
      \item The rule $\prule{H}{\neg H,\UNK{p},\UNK{t}}$ in $R_1$ does not help in
        initial states that do not satisfy $H$, and hence the width remains 2 and
        $\infty$ for \QDeliveryS and \QDelivery, respectively.
      \item The rule $\prule{\neg H}{H,\UNK{p},\UNK{t}}$ in $R_2$ says that a state
        $s$ where $\neg H$ holds can be ``improved'' by finding a state $s'$ where $H$
        holds, while possibly affecting $p$, $t$, or both.
        This rule splits every problem $P$ in \QDeliveryS into two subproblems:
        achieve $H$ first and then the goal, reducing the sketch width of \QDeliveryS
        to 1. %  but not the sketch width of \QDelivery.
      \item The rule set $R_3$ is not acyclic and thus not a proper sketch.
      \item The sketch $R_4$ decomposes problems using the feature $u$ that counts
        the number of undelivered packages, reducing the width of \QDelivery to 2, but
        not affecting the width of \QDeliveryS. The reduction occurs because each problem
        $P$ in \QDelivery is split into subproblems, each one for delivering a single
        package, similar to the problems in \QDeliveryS.
      \item $R_5$ combines the rules in $R_2$ and $R_4$.
        Each problem in \QDelivery is decomposed into subproblems, each one like a problem in \QDeliveryS,
        and each problem in \QDeliveryS is further decomposed into two subproblems of width 1 each.
        The combined result is that the sketch width of \QDeliveryS and \QDelivery both get
        reduced to 1.
      \item The sketches $R_6$ and $R_7$ do not help to reduce the width for either class.
        The rule in $R_6$ generate subproblems of zero width until reaching a state where
        $\neg H$ and $\EQ{p}$ holds, for which the remaining problem has width 2 or $\infty$
        for either \QDeliveryS or \QDelivery, respectively.
        $R_5$, on the other hand, does not help as the initial states do not satisfy $H$.
      \item Finally, the sketch $R_8$ yields a serialization of zero width, and hence a full
        policy, where each subproblem is solved in a single step.
        % However, the policy defined      by this sketch is syntactically different from the one in Example~\ref{ex:pi:delivery}.
    \end{enumerate}
  \end{boxed-example}
}

\begin{table}
  \centering
  \begin{tabular}{lcccccc}
    \toprule
                                                               &&         &&\multicolumn{2}{c}{Sketch width} \\
    \cmidrule{5-6}
    Rule set                                                   && Acyclic && \QDeliveryS & \QDelivery \\
    \midrule
    $R_0=\{\,\}$                                               &&  \cmark &&           2 &   $\infty$ \\
    $R_1=\{\prule{H}{\neg H,\UNK{p},\UNK{t}}\}$                &&  \cmark &&           2 &   $\infty$ \\
    $R_2=\{\prule{\neg H}{H,\UNK{p},\UNK{t}}\}$                &&  \cmark &&           1 &   $\infty$ \\
    $R_3=R_1\cup R_2$                                          &&  \xmark &&         --- &        --- \\
    $R_4=\{\prule{\GT{u}}{\DEC{u},\UNK{H},\UNK{p},\UNK{t}}\}$  &&  \cmark &&           2 &          2 \\
    $R_5=R_2\cup R_4$                                          &&  \cmark &&           1 &          1 \\
    $R_6=\{\prule{\neg H,\GT{p}}{\DEC{p},\UNK{t}}\}$           &&  \cmark &&           2 &   $\infty$ \\
    $R_7=\{\prule{H,\GT{t}}{\DEC{t},\UNK{p}}\}$                &&  \cmark &&           2 &   $\infty$ \\
    $R_8=R_2\cup R_4\cup R_6\cup R_7$                          &&  \cmark &&           0 &          0 \\
    \bottomrule
  \end{tabular}
  \caption{Different sketches for the Delivery domain, one rule set per line. The table shows
    whether each  rule set is feature-acyclic and also upper bounds  the  width for sketch
    for the classes \QDeliveryS and \QDelivery of Delivery problems. 
    The rule set $R_3$ is not a proper sketch as it is not acyclic; hence, the entries marked as `---'.
    For feature-acyclic sketches of bounded width, \siwR solves any instance in the class in polynomial time.
%     Unbounded sketch width over a class $\Q$ does not imply that \siwR may fail due to reaching a
%     dead-end: if all the problems $P$ in $\Q$ have bounded serialized width, \siwR is still guaranteed to
%     find a plan, but not necessarily in polynomial time.
  }
  \label{table:sketches:delivery}
\end{table}

\subsection{Acyclicity and Termination}

The notion of acyclicity appears in three places in our study.
First, if a policy $\pi$ is closed and acyclic in a problem $P$,
then $\pi$ solves $P$. Second, serializations must be acyclic, as otherwise,
even if subproblems have small, bounded width, the \siwp procedure may get
stuck in a cycle. %  of subproblems without getting to the goal.
Third, feature acyclicity has been used above to provide runtime bounds.

\Omit{
  \alert{** This subsection needs to be simplified and shortened; keeping the essential only.
    Same with Theorem 40: If Sieve accepts, then 1) no infinite $R$-sequences (state acyclicity; termination),
    2) no $R$-sequence where the same feature valuation is encountered twice (feature acyclicity). Also, don't
    see reference in text to Fig 7? **
  }
}

% does not ensure that the problem is solved as a cyclic behaviour
% may appear where the same subproblem appears infinitely often.

\Omit{
  \alert{** Termination analysis can provide better bounds on the
    number of subproblems, policy executions, etc. E.g., if \textsc{sieve}
    takes $k$ iterations to render a acyclic graph, probably instead of
    the $N^{|\Phi|}$ factor, we should have a $N^k$ factor. In a sense,
    $k$ is the max loop nesting. We can add a comment if clear enough;
    but do not have to complicate ourselves ** I don't see that clear **
  }.

  Policies and serializations are binary relations that, as we have seen,
  can be defined by means of rules: policy rules in the first case, sketch
  rules, in the second, with exactly the same syntax but with a slightly
  different semantics.
}

Interestingly, there are \emph{structural} conditions on the set of rules $R$
that ensure that the resulting binary relation on pairs of states $(s,s')$
is \textbf{feature-acyclic} by virtue of the form of the rules and the features
involved, \emph{independently of the domain.}

This is the case, for example,
if $R$ only contains the rules $r_1=\prule{\neg H}{H,\DEC{n}}$ and $r_2=\prule{H}{\neg H}$.
A sequence of states $s_0,s_1,s_2,\ldots$ compatible with $R$ cannot contain
infinite state pairs $(s_i,s_{i+1})$ compatible with rule $r_1$, because
such a rule requires feature $n$ to decrease but $n$ cannot decrease below
zero and no rule allows $n$ to increase. Then, since rule $r_1$ cannot
be ``applied'' infinitely often, neither can rule $r_2$ which requires $r_1$ to restore the truth of the condition $H$.
This analysis is \emph{independent} of the underlying planning problem and the semantics of the features.

The notion of \textbf{termination} as captured by the \sieve algorithm for QNPs
\cite{sid:aaai2011,bonet:qnps} can be used to check, among other things, that a
rule-based policy or sketch is feature-acyclic.
Indeed, if $R$ is a terminating set of rules over the features in $\Phi$, as
determined by \sieve, $s_0,s_1,s_2,\ldots$ is a state sequence compatible with $R$,
and $s_i$ and $s_j$, with $i<j$, are two states with identical valuation over
the Boolean conditions defined by $\Phi$ (see below), then
there is a numerical feature $n$ such that its values satisfy $n(s_j)<n(s_i)$ and
$n(s_k)\leq n(s_i)$ for $i\leq k\leq j$.
This condition ensures that $R$ is feature-acyclic, and thus that it is
acyclic over any class $\Q$.

%%More generally, there exists a test for checking whether a set of rules
%%$R$ is \textbf{terminating} by virtue of the form of the rules, thus
%%implying that $R$ is feature acyclic for any class $\Q$ of problems.
%%The procedure called \textsc{sieve} was introduced by \citeay{sid:aaai2011}
%%in the context of Qualitative Numerical Planning (QNP) problems
%%\cite{sid:aaai2011,bonet:qnps,ivan:fond+}.
%%\sieve can be applied to any set of rules, either defining a policy
%%or sketch, and runs in time that is exponential in the number of
%%features in $\Phi$.

The \sieve algorithm, shown in Fig.~\ref{fig:sieve}, receives as input
a directed and edge-labeled graph $G=\tup{V,E,\ell}$, where the edge labels
$\ell(e)$ contain effects over Boolean and numerical features; i.e.,
expression of the form $p$, $\neg p$ and $\UNK{p}$ for Boolean features
$p$, and expressions of the form $\DEC{n}$, $\INC{n}$, and $\UNK{n}$
over numerical features $n$.
\sieve iteratively computes the strongly connected components (SCCs)
of a graph $G'$, initially set to the input graph $G$, and removes
edges from the graph until it becomes acyclic, or no more edges can
be removed. The graph is \textbf{accepted} iff it becomes acyclic,
otherwise is \textbf{rejected}.
An edge $e$ in a component $C$ of $G'$ can be removed if some feature $n$
is decreased in $e$ (i.e.\ $\DEC{n} \in\ell(e)$), and is not increased
in any other edge $e'$ in the same component (i.e.\ $\INC{n}\not\in\ell(e')$
and $\UNK{n}\not\in\ell(e')$).

The graph $G(R)=\tup{V,E,\ell}$ that is passed to \sieve as input is
constructed from a set $R$ of rules over a set $\Phi$ of features
as follows.
The vertices in $V$ correspond to the $2^{|\Phi|}$ valuations $v$
for the conditions $p$ and $\EQ{n}$ for the Boolean and numerical
features $p$ and $n$ in $\Phi$, and there is an edge $(v,v')$ in $E$
if the pair of valuations $v$ and $v'$ is compatible with some
rule $C\mapsto E$ in $R$.
A set of rules $R$ is \textbf{terminating} iff \sieve accepts the
graph $G(R)$.
% The theoretical properties of \sieve have been already established:
In our setting, this can be means the following:

\begin{theorem}[\citeay{sid:aaai2011}, \citeay{bonet:qnps}]
  \label{thm:sieve}
  Let $\Phi$ be a set of features, and let $R$ be a set of rules over $\Phi$ for a class $\Q$ of problems.
  If \sieve accepts $G(R)$, then  the binary relation $\prec_R$ is   feature-acyclic over $\Q$, and therefore,
  $R$ is a sketch that defines a serialization   $\prec_R$ for  $\Q$.
\end{theorem}
\Proof{%
  If \sieve accepts $G(R)$, $\tau=s_0,s_1,s_2,\ldots$ is a state sequence
  that is compatible with $R$, and the states $s_i$ and $s_j$, $i<j$,
  have identical valuation over the Boolean conditions for $\Phi$,
  then there is a numerical feature $n$ that is decremented in the
  path $\tau_{i,j}=s_i,s_{i+1},\ldots,s_j$ and not incremented in $\tau_{i,j}$
  \cite{sid:aaai2011,bonet:qnps}.
  Therefore, $f(s_i)\neq f(s_j)$. This implies that there is no such
  state sequence $\tau$ that contains two different states $s$ and $s'$
  such that $f(s)=f(s')$.
}

\begin{figure}
  \begin{tcolorbox}[title=\textbf{Algorithm~\ref{alg:sieve}:} \sieve]
    \refstepcounter{myalgorithm}
    \label{alg:sieve}
    \begin{algorithmic}[1]\small
      %\State \textbf{Input:} Graph $G=G(R)$ where $R$ is set of rules
      \State \AlgINPUT  Directed edge-labeled graph $G=\tup{V,E,\ell}$, where the labels contain feature effects
      \State \AlgOUTPUT Either accept or reject $G$
      \smallskip
      \State Initialize the graph $G' \gets G$
      \smallskip
      \State Repeat
      \State \qquad Compute the SCCs of graph $G'$
      \vskip .1em
      \State \qquad Choose SCC $T$ and numerical feature $n$ that is decreased but not increased in $T$; i.e.,
      \Statex\qquad\quad -- $T$ contains some edge $e$ such that $\DEC{n}\in \ell(e)$, and
      \Statex\qquad\quad -- $T$ contains no edge $e'$ such that $\INC{n}\in \ell(e')$ or $\UNK{n}\in \ell(e')$
      \vskip .1em
      \State \qquad Remove the edges in $T$ where $n$ is decreased
      \State until $G'$ is acyclic or no such SCC exist
      \smallskip
      \State \AlgACCEPT if $G'$ is acyclic, \AlgREJECT otherwise
    \end{algorithmic}
  \end{tcolorbox}
  \caption{The \sieve algorithm takes as input a directed edge-labeled graph $G$ that is either accepted or rejected.
    The graph $G$ is the graph $G(R)$ constructed using the feature-based rules in a set $R$.
    If $G$ is accepted, the binary relation on feature valuations induced by $R$ is deemed as \textbf{terminating},
    and thus $R$ is feature-acyclic (cf.\ Theorem~\ref{thm:sieve}).
    %Hence, $R$ defines a simple sketch on any class $\Q$ of problems for which the features used in $R$ are
    %well defined.
  }
  \label{fig:sieve}
\end{figure}

\Example{
  \medskip
  \begin{boxed-example}[\sieve for different sketches for the Delivery domain]
    \label{ex:sieve}
    Let us consider the Delivery domain with the set of features $\Phi=\{H,p,t,u\}$
    discussed in Example~\ref{ex:pi:delivery}. We consider different sets of rules
    that define different input graphs for \sieve:
    \begin{enumerate}[$\bullet$]
      \item First, let us consider the set $R_1$ of rules comprised of
        $\prule{H}{\neg H,\UNK{p},\UNK{t}}$ and $\prule{\neg H}{H,\UNK{p},\UNK{t}}$.
        The graph $G(R_1)$ contains many cycles as the first rule connects all
        nodes in which $H$ holds to all nodes in which $\neg H$ holds, and
        the second rule does the opposite. As no edge is labeled with a decrement,
        \sieve cannot remove any edge from the graph, and thus \textbf{rejects.}
      \item Consider now the set $R_2$ with $\prule{\neg H}{H,\UNK{p},\UNK{t}}$
        and $\prule{\GT{u}}{\DEC{u},\UNK{H},\UNK{p},\UNK{t}}$.
        In the graph $G(R_2)$, the numerical feature $u$ is decreased in some edges,
        but no edge contains $\INC{u}$ or $\UNK{u}$ in its label.
        Hence, \sieve removes all edges
        that contain $\DEC{u}$ in its label. This renders the graph acyclic as
        the only edges left connect nodes where $\neg H$ holds to nodes where
        $H$ holds.
        \sieve \textbf{accepts} the input graph $G(R_2)$, and thus $R_2$ can be
        used as an sketch for \QDelivery by Theorem~\ref{thm:sieve}.
      \item Finally, let us consider the rule-based policy $\pi$ given in
        Example~\ref{ex:pi:delivery}.
        Figure~\ref{fig:sieve:graph} shows the (part of the) graph $G(\pi)$ that
        is given to \sieve.
        The graph contains three strongly connected components $C_1$, $C_2$, and
        $C_3$, with the first one being the only non-singleton component.
        In such a component, the numerical feature $u$ is decreased but not increased.
        \sieve then removes all the edges whose label contains \DEC{u}.
        The removal of such edges renders the graph acyclic which leads
        \sieve to \textbf{accept} $G(\pi)$.
        By Theorem~\ref{thm:sieve}, $\pi$ defines a sketch for \QDelivery.
    \end{enumerate}
  \end{boxed-example}
}

\begin{figure}[t]
  \centering
  \resizebox{\linewidth}{!}{
  \begin{tikzpicture}[thick,>={Stealth[inset=2pt,length=8pt,angle'=33,round]},
                        font={\footnotesize},node distance=2cm,
                        qs/.style={draw=black,fill=gray!20!white},
                        init/.style={qs,fill=yellow!50!white},
                        goal/.style={qs,fill=green!50!white},
                        qa/.style={qs,fill=red!50!white}]
    %% Graph for delivery with unique target cell
    \node[init]                    (n0) at (0,0) { $n_0: \overline{H}, \GT{p}, \EQ{t}, \GT{u}$ };
    \node[qa,  right = of n0]      (n2)          { $n_2: \overline{H}, \EQ{p}, \EQ{t}, \GT{u}$ };
    \node[qs,  right = of n2]      (n4)          { $n_4: H,            \EQ{p}, \EQ{t}, \GT{u}$ };
    \node[goal,right = of n4]      (n6)          { $n_6: \overline{H}, \EQ{p}, \EQ{t}, \EQ{u}$ };

    \node[init,below = 1.5 of n0]  (n1)          { $n_1: \overline{H}, \GT{p}, \GT{t}, \GT{u}$ };
    \node[qs,  right = of n1]      (n3)          { $n_3: \overline{H}, \EQ{p}, \GT{t}, \GT{u}$ };
    \node[qs,  right = of n3]      (n5)          { $n_5: H,            \EQ{p}, \GT{t}, \GT{u}$ };
    \node[qa,  right = of n5]      (n7)          { $n_7: \overline{H}, \GT{p}, \EQ{t}, \EQ{u}$ };
    %\node[goal,right = of n5, diagonal fill={red!70!white}{green!50!white}] (n7) { $n_7: \overline{H}, \GT{p}, \EQ{t}, \EQ{u}$ };

    \path[->] (n0) edge[out=140,in=40,looseness=4] node[above,yshift=0] { $\{ \DEC{p}, \UNK{t} \}$ } (n0);
    \path[->] (n0) edge[transform canvas={xshift=-30}] node[left,yshift=0] { $\{ \DEC{p}, \UNK{t} \}$ } (n1);
    \path[->] (n0) edge[] node[sloped,xshift=32,yshift=7] { $\{ \DEC{p}, \UNK{t} \}$ } (n3);
    \path[->] (n0) edge[red!70!white] node[above,yshift=0,red!70!white] { $\{ \DEC{p}, \UNK{t} \}$ } (n2);
    \path[->] (n2) edge[transform canvas={yshift=4},red!70!white] node[above,yshift=0,red!70!white] { $\{ H \}$ } (n4);
    \path[->] (n4) edge[] node[above,yshift=0] { $\{ \neg H, \DEC{u}, \UNK{p} \}$ } (n6);
    \path[->] (n4) edge[transform canvas={yshift=-4},red!70!white] node[below,yshift=0,red!70!white] { $\{ \neg H, \DEC{u}, \UNK{p} \}$ } (n2);
    \path[->] (n4) edge[red!70!white] node[sloped,xshift=10,yshift=7,red!70!white] { $\{ \neg H, \DEC{u}, \UNK{p} \}$ } (n7);
    \path[->] (n4) edge[transform canvas={xshift=0},bend right=20] node[above,sloped,xshift=0,yshift=0] { $\{ \neg H, \DEC{u}, \UNK{p} \}$ } (n0);

    \path[->] (n1) edge[out=220,in=320,looseness=4] node[below,yshift=0] { $\{ \DEC{p}, \UNK{t} \}$ } (n1);
    \path[->] (n1) edge[transform canvas={xshift=-10}] node[right,yshift=10] { $\{ \DEC{p}, \UNK{t} \}$ } (n0);
    \path[->] (n1) edge[red!70!white] node[sloped,xshift=-32,yshift=7,red!70!white] { $\{ \DEC{p}, \UNK{t} \}$ } (n2);
    \path[->] (n1) edge[] node[above,yshift=0] { $\{ \DEC{p}, \UNK{t} \}$ } (n3);
    \path[->] (n3) edge[] node[above,yshift=0] { $\{ H \}$ } (n5);
    \path[->] (n5) edge[] node[right,yshift=0] { $\{ \DEC{t} \}$ } (n4);
    \path[->] (n5) edge[out=220,in=320,looseness=4] node[below,xshift=0,yshift=0] { $\{ \DEC{t} \}$ } (n5);
  \end{tikzpicture}}
  \caption{Relevant part of the graph $G(R)$ passed to \sieve for the set $R$ of rules
    for the Delivery domain that contains $\prule{\neg H,\GT{p}}{\DEC{p},\UNK{t}}$,
    $\prule{\neg H, \EQ{p}}{H}$, $\prule{H,\GT{t}}{\DEC{t}}$, and $\prule{H,\GT{u},\EQ{t}}{\neg H, \DEC{u}, \UNK{p}}$.
    Yellow and green nodes identify the initial and goal states respectively.
    Red nodes and edges stand for nodes and transitions in the policy graph
    that do not arise in instances.
    $G(R)$ has the strongly connected components $C_1=\{n_0,n_1,\ldots,n_5\}$, $C_2=\{n_6\}$, and $C_3=\{n_7\}$.
    In the component $C_1$, the numerical feature $u$ is decreased but not increased.
    Hence, \sieve removes all the edges whose label contains $\DEC{u}$.
    After this removal, the graph becomes acyclic because all the edges
    that go from right to left in the figure contain $\DEC{u}$ in their
    labels.
  }
  \label{fig:sieve:graph}
\end{figure}

\section{Summary of Results and Meaning}

In this section, we summarize the main ideas and results of the paper.
For simplicity, we do not restate the conditions of the theorems in full
and focus instead on their meaning and the story that they reveal.
While the work builds on an earlier paper \cite{bonet:aaai2021}, most of the
results are new and convey a simpler, more meaningful narrative.
Three key changes are: an slightly more general and convenient definition of
admissibility based on \emph{sets} of tuples and not sequences, the new
notion of envelopes, and the definition of both policies and serializations
as binary relations on states, expressed syntactically by means of rules.
In addition, serializations are no longer assumed to be transitive relations. 
As usual, $P$ is a planning instance from a class $\Q$, $\pi$ is a
general policy, $T$ is a set atom tuples over $P$, and $T^k$ denotes the
set of conjunctions of up to $k$ atoms.
A summary of the main theorems above and their meaning follows:

\begin{enumerate}[$\bullet$]
  \item Theorems~\ref{thm:iw(T)}--\ref{thm:iw}.
    If $T$ is admissible, \iw{T} finds an optimal plan and $w(P)\leq\size(T)$.
    If $w(P)\leq k$, \iw{k} finds an optimal plan, and IW finds a (not necessarily optimal) plan.
  \item Theorem~\ref{thm:cost-envelope:admissible}: $T$ admissible iff $OPT(T)$ is a cost-envelope.
  \item Theorem~\ref{thm:cost-envelope:iw(T)}: \iw{T} is optimal if $T$ contains $T'$ such that
    $\OPT(T')$ is a cost-envelope, and \iw{k} is optimal if such $T'$ is contained in $T^k$.
\end{enumerate}

\noindent \textbf{Meaning.}
If $T$ is admissible and hence a cost envelope, $P$ is solved optimally by
\iw{T} which expands up to $|T|$ nodes and also by
\iw{T'} if $T' \subseteq T$. The \textbf{width} of $P$, $w(P)$,
is the minimum $size(T)$ of an admissible $T$, and \iw{k} solves $P$ optimally if $w(P) \leq k$.
\iw{k} is equivalent to \iw{T^k}.

\begin{enumerate}[$\bullet$]
  \item Theorem~\ref{thm:cost-envelope:optimal-pi}: $\OPT(T)$ is a cost-envelope in $P$ if it is a closed $\pi$-envelope of an optimal policy $\pi$ for $\Q$, $P \in \Q$.
  \item Theorem~\ref{thm:iwt:pi} and corollaries: If  $\OPT(T)$ is a closed $\pi$-envelope of an  optimal policy $\pi$, and $T \subseteq T^k$,
    \iw{k} reaches goal of $P$ through an optimal $\pi$-trajectory.
  \item Theorem~\ref{thm:width:necessity}:   $w(P)>k$ if  every optimal plan for $P$ contains a state outside $\OPT(T^k)$.
    
\end{enumerate}
\noindent\textbf{Meaning.}
These results and the ones above explain why many standard planning
domains have bounded width when goal atoms are considered.
The reason is that such classes of problems admit general optimal
policies $\pi$ that can be ``applied'' in an instance $P$ by just
considering tuples of atoms of bounded size.
If this bound is $k$, \iw{k}  finds the goal of $P$ in
polynomial time through a $\pi$-trajectory, without having to know
$\pi$ at all. The width of $P$ is greater than $k$ if no optimal plan ``goes through''
states that are all  in  $\OPT(T^k)$.

% \item \iwf{\Phi} complete if $\OPT(F)$ is a cost-envelope for some set $F$ of feature valuations
\begin{enumerate}[$\bullet$]
  \item Theorem~\ref{thm:iw(Phi):optimal}: \iwf{\Phi} is optimal if $\OPT(F)$ is a closed $\pi$-envelope for some set $F$ of feature valuations,
    and $\pi$ is optimal.
  \item Theorem~\ref{thm:iw(Phi):simple}: $\OPT(F)$ is a closed $\pi$-envelope if $\pi$ is optimal and reaches all and only the states in $\OPT(F)$.
\end{enumerate}
\noindent\textbf{Meaning.}
The \iwf{\Phi} procedure is like \iw{T} but with the feature valuations
over $\Phi$ playing the role of the atom tuples in $T$.
The procedure is meaningful because in many tasks the number of possible
feature valuations for a given $\Phi$ is exponentially smaller than the
number of states (e.g., $\QClear$ above).
% , or because a problem $P$ may
% have unbounded width, due to its encoding, but be solvable by \iwf{\Phi} (e.g., the Marbles domain).
Two relevant questions are what sets of features $\Phi$ ensure that \iwf{\Phi}
solves a problem (optimally) and whether the features used by a policy $\pi$
that solves the problem do.
The general answer to this last question is no: as shown in the example
for Towers of Hanoi, one can define general policies in terms of a bounded
and small set of Boolean features $\Phi$,
\Omit{
  \footnote{\alert{(** CHECK: this is not
  a policy in our setting because alternation issue ** Also, don't need
  to spell out policy here as it appears in a boxed example ** Last, if
  we want Hanoi, probably must refer to recent NeurIPS paper **)}
  A policy for Towers of Hanoi over
  3 pegs and any number of disks can be expressed in terms of rules over 3
  Boolean features $b_{ij}$, $1 \leq i < j \leq 3$ that are true when the
  disk on top of peg $i$ is larger than the disk on top of peg $j$ \cite{hanoi:aaai2023}.
  The policy alternates between two types of moves: n even moves, it moves
  the smallest disk to the peg that is on its ``left'', where the rightmost
  peg is assumed to be to the ``left'' of the leftmost peg, and in odd moves,
  it moves the non-smallest disk, to its only possible destination.
  This policy can be expressed as a simple general rule-based policy provided
  that an extra Boolean feature that tracks if the last disk moved was the
  smallest one (this requires extending the domain with a Boolean variable
  that tracks the same condition).
}}
yet the length of  any plan for Hanoi will grow exponentially 
with  the number of disks. This simple combinatorial argument rules out $\Phi$
as a good set of features for \iwf{\Phi} in Hanoi, even when $\Phi$ supports
a solution policy. The results above provide a more general argument that cuts in both ways.
Namely, if no subset $F$ of feature valuations results in $\OPT(F)$ being
a closed cost-envelope (as in Hanoi), then \iwf{\Phi} will not solve $P$
optimally in general, and if $F$ is one such set of feature
valuations, then \iwf{\Phi} will solve  $P$ optimally.

\medskip

The definition of policies and serialization as binary relation on states,
expressed in compact form by the same type of rules, makes the relation
between policies and serializations direct:
\Omit{
  \footnote{\alert{Policy: binary
    relations on states that is false on non-transitions? See interaction
    tight/loose serialization vs.\ need to close $\Q$, e.g., in Hanoi, etc.}}
}
\Omit{\footnote{\alert{1. For this,
  policies must be binary relation on states; i.e., state pairs, not just
  transitions, and the relation must be false for non transitions. This is
  important and we need to be consistent.
  2. Also check that classes $\Q_{P,\prec}$ of subproblems can be defined
  in different ways, and sometimes we are making the assumption that the
  subproblems are solved optimally, sometimes not. Be consistent.
  3. Remember that in \siwR, IW can't be assumed to be solve subproblems
  optimally, and this affects how family of subproblems $\Q_{P,\prec}$
  are defined.}
}
}
\begin{enumerate}[$\bullet$]
  \item Theorem~\ref{thm:serialization:policies}: Policy $\pi$ solves $\Q$ iff $\pi$ is a serialization over $\Q$ of  width zero  (semantics).
  \item Theorem~\ref{thm:sketch:policies}: Rule-based policy $R$ solves $\Q$ iff $R$ is  a rule-based serialization (sketch) over $\Q$ of width zero (syntax).
  \item Theorem~\ref{thm:siwp}: A serialization of width $k$ over $\Q$ implies that problems $P$ in $\Q$
    can be solved by solving subproblems of width bounded by $k$, greedily,
    with the \siwR procedure. The number of subproblems to be solved, and hence the running time of \siwR,
    however, is not necessarily polynomial (e.g., Hanoi).
    \Omit{\footnote{\alert{Important: \siwR may solve subproblems
      non-optimally via IW, so for the class of problems $\Q_{P,\prec}$ have to be defined without assuming that
      subproblems are solved optimally. Check if we are being consistent with this.}
    }}
  \item Theorem~\ref{thm:siwR} and \ref{thm:sieve}: A {terminating} rule-based serialization (sketch) of bounded
    width over $\Q$, implies that problems $P$ in $\Q$ are solved in polynomial time by \siwR. 
  %\item \hector{Glitch: Policy for Hanoi with 3 booleans is acyclic and solves Hanoi, hence it is a rule-based serialization (sketch), but number of subproblems is \textbf{exponential}. Indeed, it's not enough for serialization to be rule-based, it has to be \textbf{terminating}. In other words, termination gives us acyclicity (needed in def of serializations) and also a bound $N^{|\Phi|}$ on number of subproblems. This bound is not true in general, but rule-based policies.}
  \item Theorem~\ref{thm:sieve}: Termination of rule-based serializations and policies can be checked in time exponential in the number of features by \sieve.
\end{enumerate}

\noindent\textbf{Meaning.}
A general policy is a general serialization
of zero width; namely, a particular, type of serialization in which the
subproblems can be solved greedily in a single step. Serializations of
bounded width result in subproblems that can be solved greedily in polynomial
time, while terminating rule-based serializations always result in a
polynomial bounded of number of subproblems.
The direct correspondence between policies and serializations is new and important,
although it has not been recognized before. The reasons have been the lack
of general and flexible formal accounts of serializations, compact languages
for describing them, and width-like measures for bounding the complexity of
subproblems.  Policies have been formulated as binary relations on states
and not as mapping from states to actions because actions do not generalize across
instances. This choice has also helped to make the relation between policies
and serializations more explicit. Finally, termination is a property of the set of rules and has nothing do
with the class of problems. Introduced by \citeay{sid:aaai2011}, termination
gives us \emph{state acyclicity}, needed in the definition of serializations,
and \emph{feature-value acyclicity}, needed for the polynomial bound $N^{\ell}$ on the number of subproblems
(provided that features are linear), where $N$ is the number of atoms in the problem
and $\ell$ is the number of numerical features used in the rules. If a terminating sketch has bounded width over a class $\Q$ of
problems, the problems $P$ in $\Q$ are solved greedily by \siwR in polynomial time.

\section{Extensions, Variations, and Limitations}

We have presented a framework that accommodates policies and serializations,
and have established relations between width and policies, on the one hand,
and policies and serializations, on the other.
Extensions, variations, and limitations of this framework are briefly
discussed next.

% \hector{To keep in mind in this section and elsewhere: in \siwR, IW can't be assumed to be solving subproblems optimally, even if width is bounded.}

%%%%% AQUI 

\medskip
\noindent\textbf{Optimality and width.}
In the definition of an admissible set of tuples $T$ and, hence, in the definition
of width that follows, it is said that if an optimal plan $\sigma$ for a tuple $t$
in $T$ {is not an optimal plan} for $P$, then an optimal plan for another tuple
$t'$ in $T$ can be obtained by appending a single action to $\sigma$.
A similar condition appears in the original definition of admissibility and width
for sequences of tuples \cite{nir:ecai2012}.
If the condition that ``$\sigma$ \emph{is not an optimal plan} for $P$'' is replaced
by ``$\sigma$ \emph{is not a plan} for $P$,'' the resulting definition of width
(size of a min-size admissible set $T$) still guarantees that $P$ is solved by \iw{k}
if the width of $P$ is bounded by $k$, but not that $P$ is solved optimally by \iw{k}.
In the serialized setting, where optimal solutions of subproblems do not translate into
optimal solution of problems, this relaxation of the definitions of admissibility
(and width) makes sense, and it has been used for learning sketches of bounded
width more effectively \cite{drexler:icaps2022}.

\medskip

\noindent\textbf{Syntax of policy and sketch rules}. The features and rules provide a convenient, 
compact,  and general language for expressing policies and serializations, while the choice of  features,
Boolean and numerical, follow  the type of variables used in  qualitative numerical planning problems (QNPs)
for defining bounds and termination conditions \cite{sid:aaai2011,bonet:qnps}.
In QNPs, it is critical that numerical variables change via  ``qualitative'' increments and
decrements, as reasoning with arbitrary numerical is  undecidable \cite{helmert:numeric}.
Still, the restriction that numerical features $n$ can \emph{only} appear in effect expressions
of the form $\INC{n}$, $\DEC{n}$ or $\UNK{n}$ is somewhat arbitrary, and  other 
effect expressions like $\neg\INC{n}$, $\neg\DEC{n}$, $\EQ{n}$, or $\GT{n}$ can  be accommodated
with minor changes. %\hector{** Added this paragraph **}

\medskip
\noindent\textbf{Non-deterministic sketches and policies.}
Policies and sketches are non-deterministic in the sense that many state
transitions and pairs $(s,s')$ can be compatible with a policy or sketch
rule. If a policy $\pi$ solves a problem $P$, it is because \emph{all}
$\pi$-transitions lead to the goal, and in the case of a bounded-width
sketch, it is because the achievement of \emph{any} such subgoal $s'$
leads to the goal.
This means that in a state $s$, one can pick any (policy or sketch) rule
$C \mapsto E$ such that $C$ is true in $s$, and move from $s$ to any state
$s'$ that satisfies the rule.
If the sketch has bounded width (a policy is a sketch of zero width),
then at least one such rule exists.
In certain cases, however, it is convenient to guarantee that there 
is  one such state $s'$ for \emph{any} rule $C \mapsto E$ whose antecedent
$C$ is true at $s$, so that one can choose the rule to ``apply'' in a state
without having to look ahead for the existence of such states $s'$.
Sketches that have this additional property have been called \emph{modular},
as the sketch rules are considered independently of each other.
%\footnote{\alert{(** Before, we called this soundness **)}}
One can then talk about the width of a sketch rule $C \mapsto E$ as the
maximum width of the problems with initial state $s$ and goal states $s'$
such that the state pair $(s,s')$ satisfies the rule, or $s'$ is a goal state
of the problem. Modular sketches are useful for learning \emph{hierarchical policies},
where a sketch rule representing a class of problems of width greater than
$k$ is decomposed into sketch rules of width bounded by $k$, and so on
iteratively, until sketch rules are obtained with   width zero \cite{drexler:kr2023}.

\medskip
\noindent\textbf{Non-deterministic domains.}
The notion of width and the type of general policies considered are for
deterministic planning domains. It is not yet clear how to extend the width  notion
to non-deterministic domains while preserving certain key properties like
that bounded width problems can be solved in polynomial time, and that large
classes of benchmark domains fall into such a class for suitable types of goals.
The extension of general policies for non-deterministic domains appears to be simpler
although it has not been explored. In principle, a policy $\pi$ for non-deterministic
domains would still classify  state transitions $(s,s')$ as good or bad (in $\pi$ or not),
and non-deterministic actions in $s$ are compatible with $\pi$ if they give
rise to good state transitions from $s$ only.
% Such a policy $\pi$ solves a non-deterministic problem if all the fair
% $\pi$-trajectories reach the goal, with the notion of fairness being
% determined by the type of action non-determinism \cite{ivan:fond+}.

\section{Related Work}

We review a number of related research threads.

\medskip

\noindent\textbf{Width, general policies, and sketches.}
This paper builds on prior  works that introduced sketches \cite{bonet:aaai2021},
the language for expressing general policies in terms of features and rules \cite{bonet:ijcai2018},
and the  notion of width and the  IW search procedures \cite{nir:ecai2012,nir:review}.
%Width-based algorithms have been used in classical planning \cite{nir:aaai2017,nir:icaps2017}
%and in other settings where the model of the actions is not known \cite{nir:ijcai2015,tomas:gvg}.
Methods for learning general policies and sketches of bounded width have also been developed \cite{frances:aaai2021,drexler:icaps2022},
leading more recently to methods for learning hierarchical policies  \cite{drexler:kr2023}.

\medskip
\noindent\textbf{General policies.}
The problem of learning general policies has a long history \cite{khardon:action,martin:generalized,fern:generalized},
and general policies have   been formulated in terms of first-order logic
\cite{srivastava:generalized,sheila:generalized2019}, and  first-order regression \cite{boutilier2001symbolic,wang2008first,van2012solving,sanner:practicalMDPs}.
More recently, general policies for classical planning have been represented
by neural nets and learned using {deep learning} methods
\cite{sid:sokoban,trevizan:dl,sanner:dl,erez:generalized,mausam:dl2,simon:kr2023}.

\medskip
\noindent\textbf{Hierarchical  planning.}
While the subgoal structure of a domain is important for handcrafting effective
hierarchical task networks, HTNs do not actually encode subgoal structures
 but general top-down strategies where (non-primitive)  methods decompose
into other methods  \cite{htn:planning,nau:shop,aiello:htns}.
Techniques for learning HTNs  usually appeal to annotated traces that convey the
intended decompositions \cite{learning:htn,learning:htn2}, and 
other methods   for deriving hierarchical decompositions in planning include
precondition relaxations \cite{sacerdoti:precs} and causal graphs \cite{knoblock:hierarchy}.

\medskip
\noindent\textbf{Hierarchical RL and intrinsic rewards.}
Hierarchical structures have also been used in reinforcement learning in the
form of options \cite{sutton:options}, hierarchies of machines
\cite{parr:ham} and MaxQ hierarchies \cite{dietterich:maxq},
among others. While this ``control knowledge'' is often provided by hand, a vast literature
has explored techniques for learning them by considering ``bottleneck states''
\cite{mcgovern-barto-icml2001}, ``eigenpurposes'' of the matrix dynamics \cite{machado:eigenpurposes},
and informal width-based considerations \cite{junyent:width}. Intrinsic rewards have also been
introduced for improving exploration leading to  exogenous rewards \cite{singh:intrinsic},
and some authors have addressed the problem of learning intrinsic rewards. Interestingly, the
title of one of the papers in the area is the question ``What can learned intrinsic rewards capture?''
\cite{singh:intrinsic2}. The answer that follows from  our setting is clean and simple: intrinsic
rewards should capture the general, low-width subgoal structure of the domain. Lacking a language to talk
about families of problems and about subgoal structure, however, the answer to the question found
in the RL literature is  purely experimental and less crisp: learned intrinsic rewards are
just supposed to speed up the convergence of (deep) RL.

\medskip
\noindent\textbf{Reward machines and sketches.}
A recent language for encoding subgoal structure in RL is based on reward
machines \cite{reward-machines} and the closely related proposal of
restraining bolts \cite{restraining-bolts}. 
In these cases, the temporal structure of the (sub)goals to be achieved
results in an automata which is  combined with the system MDP to produce
the so-called cross-product MDP. A number of RL algorithms for exploiting the known structure of the subgoal
automata have been developed \cite{reward-machines-algorithms} as well as
algorithms for learning them \cite{learning-reward-machines}.
There is indeed a close relation between reward machines and sketches, as
both convey subgoal structure, but there some important  differences too.
First, reward machines encode the structure of explicit temporal goals (e.g., do $X$, then $Y$, and finally $Z$),
while sketches encode structure that is implicit in the problem goal given the domain. 
Second, reward machines are defined in terms of additional propositional variables; sketches,
in terms of state features that  do not require cross-products.
Third, sketches  come with a theory of width that tells us where 
to  split problems  into subproblems and why. And, finally, sketches come
with a notion of termination that ensures that subgoaling does not result in cycles. 

\section{Conclusions}

We have established results that explain why many standard planning domains
have bounded width, and have introduced a number of notions, like policy
and cost envelopes that shed light on this relation and on the optimality
and completeness of old and new IW-algorithms like \iw{T} and \iwf{\Phi}.
We have also redefined the semantic and syntactic notions of general policies
and serializations, making their relation direct and clean: a policy is a
serialization that gives rise to subproblems of zero width which  can be solved
greedily. This relation between policies and problem decompositions has not
been recognized before. The paper is revised version of an earlier paper \cite{bonet:aaai2021} that
touched similar themes and introduced the notion of sketches. The goal has
been to make the results more transparent, useful, and meaningful.

\section*{Acknowledgements}

% The research of H. Geffner has been supported by the Alexander von Humboldt Foundation with funds from the Federal Ministry for Education and Research.
% This research was partially supported by the European Research Council (ERC), Grant No.\ 885107, and by project TAILOR, Grant No.\ 952215, both funded by the EU Horizon 2020 research and innovation programme, the Excellence Strategy of the Federal Government and the NRW L\"{a}nder.
% This work was partially supported by the Wallenberg AI, Autonomous Systems and Software Program (WASP) funded by the Knut and Alice Wallenberg Foundation.

The research of H. Geffner has been supported by the Alexander von Humboldt Foundation
with funds from the Federal Ministry for Education and Research. The research has also
received funding from the European Research Council (ERC), Grant agreement No.\  885107,
and Project TAILOR, Grant agreement No.\ 952215, under EU Horizon 2020 research and innovation
programme, the Excellence Strategy of the Federal Government and the NRW L\"{a}nder, and
the Knut and Alice Wallenberg (KAW) Foundation under the WASP program.

\Omit{ %% Sections in original paper

\section{Features}

\hector{From here on OLD: See what to preserved from this. Do need alternative representations at all}

\hector{Changed title from Representations to Features; Representations not that important now actually. Also: definitions of problem, domain, etc;
should be moved up, before Width; perhaps new section background. Here just Features, and in principle, Width over problems extended with features}

\Omit{
The width of a planning problem is tied to the representation language
and the encoding of the problem in the language.
For example, the problem of moving a number of packages $N$ from one room to the next,
one by one, has a width that grows with $N$ in standard encodings where each
package has a name, but width $2$ when the packages are indistinguishable
from each other and the number of packages is encoded in unary
(with one atom per counter value).\footnote{The problem would still not have
  bounded width if the counter is represented in binary using a logarithmic number
  of atoms.
}

In order to deal with a variety of possible languages and encodings, and since
width-based methods rely on the \emph{structure of states} but not on the \emph{structure of actions} (i.e.,
action preconditions and effects), we consider \emph{first-order languages} for describing states
in terms of atoms that represent objects and relations, leaving out from the language the representation of action
preconditions and effects. It is assumed that {the possible state transitions} $(s,s')$
from a state $s$ are a \emph{function of the state} but no particular representation of this
function is assumed. In addition, the state language is extended with \emph{features} $f$ whose
values $f(s)$ in a state $s$ are determined by the state. The features provide additional expressive power
and ways for bridging different state representation languages, although they are logically redundant
as their value is determined by the truth value of the atoms in the state. The features extend the notion of
\emph{derived predicates} as defined in PDDL, as they do not have to be boolean,
and they do not have to be defined in the language of first-order logic or logic programs
\cite{axioms:pddl}, but can be defined via procedures. Domains, instances, and states
are defined as follows:
}

\hector{Define features; say numerical, Boolean; moved def of domains up. Define feature valuations (don't need feature tuples) E.g.,}
A feature $r$  over a class of problems $\Q$ denotes  well-defined  function $f_r(s)$ over the  reachable states $s$ of the instances $P \in \Q$.
In our setting, the features will be   numerical or Boolean. Numerical features  take values in the non-negative integers and are denoted by symbols like $m$ and $n$,
while  Boolean features take values in the Boolean domain and  are denoted by symbols like $p$ and $q$.  For  a set of features $\Phi$, a feature valuation $f$ is a vector
that assigns to each feature in $\phi$ a possible feature value. The feature valuation $f$ at a state $s$ is given by the value $f_r(s)$  of the features $r \in \Phi$, and is denoted as $f(s)$.

\Omit{%BLAI Oct-2022
This is all standard except for the two details mentioned before: there are no action schemas,
and there are state features. For the former, it is implicitly assumed that in each problem
$P$, there is a function that maps states $s$ into the set of possible transitions $(s,s')$.
This implies, for example, that the states may contain \emph{static atoms,} like adjacency
relations, whose truth value are not affected by any action.
For the features, we make the assumption that they are \textbf{linear},
in the sense that they can be computed efficiently and only span a linear
number of values. More precisely:
}

\smallskip

\hector{Define width of a problem $P=...$ extended with set of features $F$; i.e., width of $P+F$
is the width of $P'$: a problem that extends $P$ with atoms $p$ and $n=v$ for Boolean and numerical features $\phi_p$
and $\phi_n$ respectively, where $v=0, \ldots, v_{max}$ is a non-negative integer bounded by the maximum value
that feature $\phi_n$ can take in a (reachable) state of  $P$.}

% \begin{linearassumption}

% The features $f$ in $F$ are either boolean or numerical, ranging in the latter case
%   over the non-negative integers.

\hector{Moved from below to here; need editing} The \textbf{boolean feature valuations} determined by state $s$
refer to the truth valuations of the expressions $p$ and $n=0$ for the
boolean and numerical features $p$ and $n$ in $\Phi$, respectively.
%over the expressions $p(s)=true$ and $n(s)=0$
%that result from a feature valuation over the features $p$ and $n$ in $\Phi$.
%Notice that
While the number of feature valuations is not bounded
as the size of the instances in $\Q$ is not bounded in general,
%as when there is no bound on the size of the instances in $\Q$,
the number of boolean feature valuations is always $2^{|\Phi|}$.

\noindent \textbf{Linear features:} We say that a feature is linear when it can be computed
in linear time in the number $N$ of problem atoms, and it can take at most a linear number of values.
In some theorems below, we will make the assumption, common in practice, that all features
considered are linear in this sense. \hector{Adjusted this; btw, why you were saying time $O(bN)$
where $b$ is branching factor? Branching factor shouldn't get for characterizing state features?}
%   We assume a computational model where the arithmetic operations and comparisons over integers take constant time.
%   \blai{** Generalize assumption to poly-time features **}
% \end{linearassumption}

\smallskip

\Omit{%BLAI Oct-2022
This assumption rules out  features like $V^*(s)$ that stands for the optimal cost (distance) from $s$ to a goal
which may take a number of values that is not linear in the number of problem atoms, and whose computation may take
exponential time. In many cases, the features can be defined in the language of first-order logic
but this is not a requirement.
}

\Omit{
  For instance, in the Blocks world, $clear(x)$ can be defined as a boolean feature determined
  by the location of the blocks; i.e., $clear(x)$ iff $\neg \exists y. \, on(y,x)$, while counters $n(x)$
  can be defined as $|\phi(x)|$ for a formula $\phi(x)$ with $x$ as the only free variable,
  that counts the number of objects $c$ such that $\phi(c)$ is true in the state.
} % omit

\smallskip

\hector{Perhaps too many examples for something so simple? See eventually with full text, but emphasis should no longer be in conveying intuitions for the definition of width. }

\begin{example-no-eob}
  Three {state languages} for Blocksworld are:
  \begin{enumerate}[1.] \denselist
    \item $\L_{BW}^1$ with the binary predicate (symbol) $on^2$ and the unary $ontable^1$ (superindex indicates arity),
    \item $\L_{BW}^2$ with predicates $on^2$, $ontable^1$, $hold^1$, and $clear^1$,
    \item $\L_{BW}^3$ with predicates $on^2$ and $hold^1$, and boolean features $ontable^1$ and $clear^1$. \eob
  \end{enumerate}
\end{example-no-eob}

\begin{example-no-eob}
  Four languages for a domain {Boxes}, where boxes $b$ containing marbles
  $r$ must be removed from a table, and  for this,  marble must be removed one by one  first:
  \begin{enumerate}[1.] \denselist
    \item $\L_{B}^1$ with predicates $ontable^1(b)$ and $in^2(r,b)$,
    \item $\L_{B}^2$ with predicates $ontable^1$, $in^2$, and $empty^1(b)$,
    \item $\L_{B}^3$ with predicates $ontable^1$ and $in^2$, and features $n(b)$ that count the number of marbles in $b$,
    \item $\L_{B}^4$ with predicates $ontable^1$ and $in^2$, and features $m$ and $n$ counting the number of marbles in a
      box with the least number of marbles, and the number of boxes left. \eob
  \end{enumerate}
\end{example-no-eob}

\Omit{%BLAI Oct-2022
By abstracting away the details of the domain dynamics and the ability
to introduce features, it is simple to move from one state
representation to another.

The notion of width and the IW algorithms generalize to state
languages containing features in a direct fashion.
In both cases, the set of atoms considered is extended to contain
the possible feature values $f{=}v$ where $f$ is a
feature and $v$ is one of its possible values.
Features are logically redundant but
can have drastic effect on the problem width.

The width for class $\Q$ of problems $P$ over some domain $D$
is $k$, written as $w({\Q})=k$, if $w(P)=k$ for some $P \in \Q$
and $w(P') \leq k$ for every other problem $P'$ in $\Q$.
}

\smallskip

\begin{example}
  The width for the class of problems $\Q_{clear}$ where block $x$
  has to be cleared has width $1$ in the state languages
  $\L_{BW}^i$, $i=1,2,3$, while the class $\Q_{on}$ has
  width $2$ for the three languages.
  On the other hand, for the class $\Q_{B_1}$ of instances from
  Boxes with a single box, the width is not bounded as it grows
  with the number of marbles when encoded in the languages
  $\L_{B}^1$ and $\L_B^2$, but it is $1$ when encoded in the
  languages $\L_B^3$ or $\L_B^4$.
  Likewise, for the class $\Q_{B}$ of instances from Boxes with
  arbitrary number of boxes, the encoding in the language $\L_B^3$
  has width that is not bounded as it grows with the number of boxes, but remains
  bounded and equal to $2$ in $\L_B^4$. %\footnote{See the supplemental material for the proofs.}
\end{example}

\Omit{
  \begin{proof}[Proof sketch]
    The width of 1 for $\Q_{clear}$ over the languages $\L_{BW}^i$, $i=1,2,3$,
    is established before, as well as the width equal to 2 for $\Q_{on}$
    on the same three languages.

    Let us study the width of $\Q_{B_1}$, the instances from Boxes with a single box,
    and $\Q_B$, the instances from Boxes with arbitrary number of boxes, over the
    languages $\L_{B}^i$ for $i=1,2,3,4$.
    Observe that for $\Q_{B_1}$, the languages $\L_{B}^i$ for $i=3,4$ are equivalent
    since the feature $n(b)$ in $\L_B^3$ is exactly the feature $m$ in $\L_B^4$.
    For $\L_{B}^4$, the chain
    \[ \{ontable(b)\},\, \{m{=}k\},\, \{m{=}k{-}1\},\, \ldots,\, \{m{=}0\},\, \{\neg ontable(b)\} \]
    is an admissible chain of 1-tuples when the box has initially $k$ marbles.
    Therefore, the width of $\Q_{B_1}$ is 1 when using $\L_B^3$ or $\L_B^4$.
    On the other hand, the width when using $\L_B^i$, $i=1,2$, is not bounded
    since there is no way to make up an admissible chain with tuples that express
    transitions where the box goes from containing $j+1$ to $j$ marbles using
    less than $k$ atoms, where $k$ is the initial number of marbles in the box.
    Indeed, such a transition must be specified with
    a pair of tuples of $in^2$ atoms, the first that makes true exactly $j+1$
    atoms of the form $in(m_i,b)$, and the second tuple that makes true exactly
    $j$ such atoms.
    Hence, the width for $\Q_{B_1}$ is unbounded when using the languages $\L_B^1$ or $\L_B^2$.

    For the class, $\Q_B$, the chain
    \[ \{n{=}\ell,m{=}k_\ell\},\, \{n{=}\ell,m{=}k_\ell{-}1\},\,\ldots,\,\{n{=}\ell,m{=}0\},\ \{n{=}\ell{-}1,m{=}k_{\ell-1}\},\,\ldots,\{n{=}1,m{=}0\},\ \{n{=}0\} \]
    where $k_\ell$ is the number of marbles in the first box with smallest number of marbles, etc,
    is an admissible chain of size 2 over the language $\L_B^4$.
    On the other hand, it can be shown that there is no admissible chain of size 1.
    Hence, the width of $\Q_B^4$ is 2 over such language.
    For the language $\L^3_B$, $\Q_B$ has unbounded width since there is no way to specify an admissible chain of tuples where the number of boxes
    in the table decrease one by one from $\ell$ to zero.
  \end{proof}
} % omit

  \input{sections/generalized_policies}
  \input{sections/generalized_policies_and_width}
  \input{sections/serialized_Width}
  \input{sections/general_policies_serializations}
  \input{sections/sketches}

}

\color{black}
\bibliography{control}
\bibliographystyle{theapa}

\end{document}